\documentclass[3p,times]{elsarticle}

\usepackage{ecrc}

\volume{00}

\firstpage{1}

\journalname{Procedia Computer Science}

\runauth{}

\jid{procs}

\jnltitlelogo{Procedia Computer Science}

\CopyrightLine{2011}{Published by Elsevier Ltd.}

\usepackage{amssymb}

\usepackage[figuresright]{rotating}

\usepackage{booktabs} 
\usepackage{epigraph}
\usepackage{algorithm}
\usepackage{algpseudocode}
\usepackage{graphicx}
\usepackage{multirow}
\usepackage{amsfonts,amssymb,amsmath,color}
\usepackage{url}
\usepackage{subfigure}
\usepackage{todonotes}
\usepackage{caption}
\usepackage{ulem}
\usepackage{color, colortbl}
\newcommand{\tabincell}[2]{\begin{tabular}{@{}#1@{}}#2\end{tabular}}

\def\E{{\bf E}}

\def\x{{\bf x}}

\def\0{{\bf 0}}
\def\1{{\bf 1}}

\def\CM{{\mathcal C}}
\def\DM{{\mathcal D}}
\def\EM{{\mathcal E}}

\def\HM{{\mathcal H}}

\def\KM{{\mathcal K}}
\def\LM{{\mathcal L}}

\def\PM{{\mathcal P}}

\def\RM{{\mathcal R}}
\def\SM{{\mathcal S}}
\def\TM{{\mathcal T}}

\def\VM{{\mathcal V}}
\def\WM{{\mathcal W}}

\newcommand{\nop}[1]{}

\newtheorem{definition}{Definition}

\definecolor{Gray}{gray}{0.9}

\newcommand{\Blue}[1]{\textcolor[rgb]{0.00,0.00,1.00}{#1}}

\begin{document}

\begin{frontmatter}



\dochead{}

\title{ASER: Towards Large-scale Commonsense Knowledge Acquisition via Higher-order Selectional Preference over Eventualities}



\author[1]{Hongming Zhang}
  \ead{hzhangal@cse.ust.hk} 
  
  \author[1]{Xin Liu}
  \ead{xliucr@cse.ust.hk} 
  
  \author[1]{Haojie Pan}
  \ead{hpanad@cse.ust.hk} 
  
  \author[]{Haowen Ke}
  \ead{hkeaa@cse.ust.hk} 
  
  \author[]{Jiefu Ou}
  \ead{jouaa@connect.ust.hk} 
  
  \author[]{Tianqing Fang}
  \ead{tfangaa@cse.ust.hk} 
  
  \author[]{Yangqiu Song}
  \ead{yqsong@cse.ust.hk} 
  
\fntext[fn1]{The first three authors make equally important contributions to this work. Detailed contributions are in the end of this paper.}











\address{CSE, HKUST, HKSAR, China}

\begin{abstract}
Commonsense knowledge acquisition and reasoning have long been a core artificial intelligence problem. However, in the past, there has been a lack of scalable methods to collect commonsense knowledge. 
In this paper, we propose to develop principles for collecting commonsense knowledge based on selectional preference, which is a common phenomenon in human languages that has been shown to be related to semantics. We generalize the definition of selectional preference from one-hop linguistic syntactic relations to higher-order relations over linguistic graphs.
Unlike previous commonsense knowledge definition (e.g., ConceptNet), the selectional preference (SP) knowledge only relies on statistical distribution over linguistic graphs, which can be efficiently and accurately acquired from the unlabeled corpus with modern tools, rather than human-defined relations.
As a result, acquiring SP knowledge is a much more scalable way of acquiring commonsense knowledge.
Following this principle, we develop a large-scale eventuality (a linguistic term covering activity, state, and event)-based knowledge graph ASER, where each eventuality is represented as a dependency graph, and the relation between them is a discourse relation defined in shallow discourse parsing.
The higher-order selectional preference over collected linguistic graphs reflects various kinds of commonsense knowledge. 
For example, dogs are more likely to bark than cats as the eventuality ``dog barks'' appears 14,998 times in ASER while ``cat barks'' only appears 6 times. ``Be hungry'' is more likely to be the reason rather than result of ``eat food'' as the edge $\langle$``be hungry,'' \texttt{Cause}, ``eat food''$\rangle$ appears in ASER while $\langle$``eat food,'' \texttt{Cause}, ``be hungry''$\rangle$ does not.
Moreover, motivated by the observation that humans understand events by abstracting the observed events to a higher level and can thus transferring their knowledge to new events, we propose a conceptualization module on top of the collected knowledge to significantly boost the coverage of ASER.
In total, ASER contains 648 million edges between 438 million eventualities. 
After conceptualization with Probase, a selectional preference based concept-instance relational knowledge base, our concept graph contains 15 million conceptualized eventualities and 224 million edges between them.
Detailed analysis is provided to demonstrate its quality.
All the collected data, APIs, and tools that can help convert collected SP knowledge into the format of ConceptNet are available at \url{https://github.com/HKUST-KnowComp/ASER}.
\end{abstract}

\begin{keyword}
Commonsense Acquisition \sep Selectional Preference \sep Eventualities


\end{keyword}

\end{frontmatter}


\section{Introduction}\label{sec:introduction}



Knowledge is crucial to understanding natural language.
When reading, in addition to linguistic knowledge of the vocabulary and grammar of a language, readers need to have knowledge about the structure of texts, knowledge about the subject, and background or commonsense knowledge about the world in order to comprehend the text.
For example, when a user says ``I am hungry'' to a chatbot at 1:00 pm, the chatbot should be able to understand that the user may want to have lunch rather than breakfast and recommend some nearby restaurants.
This requires the chatbot to understand the complex commonsense knowledge about user's states and potential consequent activities (i.e., being hungry can motivate the user to eat) and the implications of location and time (i.e., compared with breakfast, lunch is more likely to appear at 1:00 pm. Thus the chatbot should recommend some real food rather than just a cup of coffee).

Commonsense reasoning has long been a challenging problem in the artificial intelligence field.
As discussed in \cite{liu2004conceptnet}, commonsense knowledge refers to ``millions of basic facts and understanding possessed by most people.''
Unlike factual knowledge like ``London is the capital of UK,'' which is always true, commonsense knowledge is often not inevitably true and only reflects a kind of contextual preference. 
For example, in most cases, rocks are not used for eating, but some birds do eat rocks to digest.
Such kind of knowledge is also called factoids~\cite{GordonDS10a,GordonS10b}.
To effectively represent such preference-like commonsense knowledge, selectional preference~\cite{resnik1997selectional} was proposed, which was traditionally defined on top of single dependency connections (e.g., \texttt{nsubj}, \texttt{dobj}, and \texttt{amod}).
Given a word and a dependency relation, humans have preferences for which words are likely to be connected.
For instance, when seeing the verb ``sing,'' it is highly plausible that its object is ``a song,'' and when seeing the noun ``air,'' it is highly plausible that its modifier is ``fresh.''
However, such selectional preference can only represent commonsense inside an event or state (e.g., which event/state is more likely to happen) and cannot represent commonsense between events/states.
One such example is discussed by~\cite{Wilks1975IAU} and similar examples are frequently observed in the Winograd Schema Challenge~\cite{levesque2011winograd}: 
\begin{itemize}
  \item The \texttt{soldiers} fired at the \texttt{women}, and we saw several of \texttt{them} fall.
\end{itemize}
To resolve the pronoun ``\texttt{them}'' in the above example, Wilks argued that machines need to access the {\it partial information} 
``{\it hurt things tending to fall down},'' which can be translated into the following form: (hurt, \texttt{X}) $\xrightarrow{\rm ResultIn}$ (\texttt{X}, fall).

In history, many efforts have been devoted to acquiring commonsense knowledge in the form of multi-relational factoids.
For example, the Cyc project initiated in the 1980s~\cite{researchCyc} and ConceptNet (originated from Open Mind Common Sense, OMCS) initiated in 2002~\cite{liu2004conceptnet}, tried to use experts or ordinary people to annotate commonsense knowledge collectively.
However, as aforementioned, two properties of commonsense knowledge determine that we cannot acquire all commonsense knowledge with such approaches.
First, the scale of commonsense knowledge could be enormous and it is infeasible to perform crowd-sourcing for commonsense knowledge acquisition on such a huge scale. 
Second, commonsense knowledge is often a kind of preference rather than fixed fact, and thus it is not suitable to represent commonsense knowledge with fixed triplets (e.g., $\langle$``rock,'' \texttt{NotUsedFor}, ``eat''$\rangle$) as used in Cyc and ConceptNet.
Recently, pre-trained language representation models (e.g., BERT~\cite{DBLP:conf/naacl/DevlinCLT19} and RoBERTa~\cite{DBLP:journals/corr/abs-1907-11692}) have been developed to acquire rich human knowledge implicitly and have demonstrated promising results on many downstream tasks.
However, as shown in LAMA~\cite{DBLP:conf/emnlp/PetroniRRLBWM19} and TransOMCS~\cite{DBLP:conf/ijcai/ZhangKSR20}, even though these models are good at capturing factual knowledge about named entities, they still struggle at capturing commonsense knowledge, especially those complex commonsense knowledge between eventualities (a linguistic term covering activities, states, and events after~\cite{bach1986algebra}, e.g., ``I am hurt'').
One possible explanation is that compared with tokens or named entities, the distribution of eventualities is generally much more sparse.
More importantly, as discussed by ~\cite{liu2004conceptnet}, much commonsense knowledge, which is trivial for humans, is typically not discussed in our daily language at all.
As a result, even though these deep pre-trained language representation models are good at acquiring knowledge from textual data, they could not effectively acquire or reason commonsense knowledge they rarely or never see in the form of word sequences.

\begin{figure}
\centering
\includegraphics[width=0.8\linewidth]{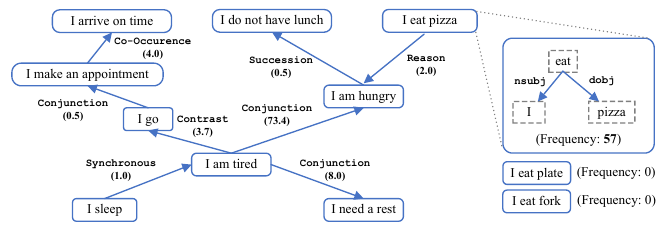}

\caption{ASER Demonstration. Eventualities are connected with weighted directed edges. Each eventuality is a dependency graph.}
\label{fig:ASER-demo}
\end{figure}


To explore a scalable way of acquiring commonsense knowledge, in this paper, we propose an approach to constructing a large-scale weighted eventuality knowledge graph, ASER (Activities, States, Events, and their Relations), by extending the traditional definition of selectional preference to higher-order selectional preference over eventualities.
The eventualities (i.e., nodes of ASER) are extracted using selected dependency patterns.
The edges are based on discourse relations (e.g., \texttt{Result}) in discourse analysis. 
As shown in Figure~\ref{fig:ASER-demo}, both nodes and edges are associated with frequency-based weights to reflect higher-order selectional preferences given a specific linguistic (either dependency or discourse) pattern.
As discussed by~\cite{resnik1997selectional,zhang2019sp-10k}, such frequency distribution can serve as a good fit for humans' selectional preference, which is indeed the commonsense knowledge.
An example is shown in Figure~\ref{fig:ASER-demo}. In ASER, ``I eat plate'' and ``I eat fork'' never appear in ASER while ``I eat pizza'' appears 57 times. We can infer that ``plate'' and ``fork'' are not subjects that can be eaten while ``pizza'' is.
Similarly, the frequencies of edges can be used to reflect higher-order selectional preference between eventualities.
For example, by observing that $\langle$``{\it Person} be hungry''-\texttt{Result}-``{\it Person} eat''$\rangle$ appear at least 12 times while $\langle$``{\it Person} be hungry''-\texttt{Reason}-``{\it Person} eat''$\rangle$ appears only once, we can know that ``{\it Person} be hungry'' is more likely to result in rather than be caused by ``{\it Person} eat.''
We argue that the higher-order selectional preference in ASER can be scalable and effective to represent previously defined commonsense knowledge types in ConceptNet~\cite{liu2004conceptnet} and potentially many other types of commonsense knowledge.

To build such a large-scale eventuality knowledge graph, we first leverage unsupervised algorithms and existing tools (e.g., dependency/discourse parsing) to extract eventualities and their relations from raw documents.
For the eventuality extraction, considering that the English language's syntax is relatively fixed and consistent across domains and topics, 
instead of defining complex triggers and role structures of events, we use syntactic patterns to extract all possible eventualities.
We do not distinguish between semantic senses or categories of particular triggers or arguments in eventualities but treat all extracted words with their dependency relations as hyperedge in a graph to define an eventuality as a primitive semantic unit in our knowledge graph.
For eventuality relation extraction, we adopt an end-to-end discourse parser~\cite{DBLP:conf/conll/WangL15} to determine the discourse relations between eventuality spans automatically and then create edges based on the predicted relation. 
Compared with previous commonsense knowledge acquisition methods, acquiring selectional preference knowledge with linguistic patterns and discourse relation prediction models is much cheaper and scalable. Thus, it can be used to extract large-scale selectional preference knowledge from the unlabeled corpus.
After that, to overcome the challenge that a large portion of the commonsense knowledge is rarely expressed in textual corpus and motivated by the observation~\cite{zacks2001event} that human beings often conceptualize the events to a more abstract level such that they can be applied to new events, we propose to leverage existing conceptualization techniques ~\cite{SongWWLC11,SongWW15} to automatically generalize the knowledge we observed and extracted to those unseen eventualities.

As a result, we create ASER, which contains 438,648,952 unique eventualities and 648,514,465 edges.
Table~\ref{tab:size_comparison} provides a size comparison between three variations of ASER\footnote{ASER (core) includes all extracted eventualities that appear more than once, ASER (full) includes all extracted eventualities, and ASER (concept) includes all conceptualized eventualities.} (i.e., core, full, and concept) and existing eventuality-related (or simply verb-centric) knowledge bases.
Essentially, they are not large enough as modern knowledge graphs and inadequate for capturing the richness and complexity of eventualities and their relations.
FrameNet~\cite{framenet} is considered the earliest knowledge base defining events and their relations. It provides annotations about relations among about 1,000 human-defined eventuality frames, which contain 27,691 eventualities. 
However, given the fine-grained definition of frames, the scale of the annotations is limited.
ACE~\cite{NIST05} (and its follow-up evaluation TAC-KBP~\cite{aguilar2014comparison}) reduces the number of event types and annotates more examples in each of the event types.
PropBank~\cite{palmer2005proposition} and NomBank~\cite{meyers2004nombank} build frames over syntactic parse trees, and focus on annotating popular verbs and nouns.
TimeBank focuses only on temporal relations between verbs~\cite{pustejovsky2003timebank}.
While the aforementioned knowledge bases are annotated by domain experts, 
OMCS/ConceptNet\footnote{Following the original definition, we only select the four relations (``HasPrerequisite,'' ``HasFirstSubevent,'' ``HasSubEvent,'' and ``HasLastSubEvent'') that involve eventualities.}~\cite{liu2004conceptnet}, Event2Mind~\cite{Event2Mind}, ProPora~\cite{proparNaacl2018}, ATOMIC~\cite{Maarten2019Atomic}, ATOMIC-2020~\cite{DBLP:conf/aaai/HwangBBDSBC21}, and GLUECOSE~\cite{DBLP:conf/emnlp/MostafazadehKMB20} leveraged crowdsourcing platforms or the general public to annotate commonsense knowledge about eventualities, in particular the relations among them.
Furthermore, KnowlyWood~\cite{TandonMDW15KnowlyWood} uses semantic parsing to extract activities (verb+object) from movie/TV scenes and novels to build four types of relations (parent, previous, next, similarity) between activities using inference rules.
Compared with all these eventuality-related KGs, ASER is larger by one or more orders of magnitude in terms of the numbers of eventualities and relations it contains.

\begin{table}[t]
    \centering
    {\footnotesize
    \begin{tabular}{l|c|c|c}
    \toprule
         & \# Eventuality & \# Relation & \# Relation Types \\
         \midrule
         FrameNet & 27,691 & 1,709 & 7 \\
         ACE & 3,290 & 0 & 0 \\
         PropBank &  112,917 & 0 & 0 \\ 
         NomBank  & 114,576 & 0 & 0 \\ 
         TimeBank & 7,571 & 8,242 & 1 \\ 
         OMCS (Only include edges about eventualities) & 74,989 & 116,097 & 4\\
         Event2Mind & 24,716 & 57,097 & 3\\
         ProPora & 2,406 & 16,269 & 1 \\
         ATOMIC & 309,515 & 877,108 & 9 \\
         ATOMIC-2020 & 638,128 & 1,331,113& 23 \\
         GLUECOSE & 286,753 & 304,099 & 10 \\
         Knowlywood & 964,758 & 2,644,415 & 4 \\ 
           \midrule
         ASER (core) & 52,940,258 & 52,296,498 & 14 \\
         ASER (full) & 438,648,952 & 648,514,465 & 14 \\
         \midrule
         ASER (concept) & 15,640,017 & 224,213,142 & 14 \\
         
    \bottomrule
    \end{tabular}
    }
    \caption{ Size comparison of ASER and existing eventuality-related resources. \# Eventuality, \# Relation, and \# Relation Types are the number of eventualities, relations between these eventualities, and relation types. For KGs containing knowledge about both entity and eventualities, we report the statistics about the eventualities subset. ASER (core) filters out eventualities that appear only once and thus has better accuracy while ASER (full) can cover more knowledge.  ASER (concept) runs conceptualization to aggregate diverse relations from a much cleaner ASER, resulting in a much denser commonsense knowledge graph.}
    \label{tab:size_comparison}
\end{table}

In summary, our contributions are as follows.
\begin{enumerate}
    \item \textbf{Representation of commonsense knowledge with higher-order selectional preference}: We extend the original definition of selectional preference to higher-order selectional preference between eventualities, and show that we can cheaply acquire selectional preference knowledge from the unlabeled corpus and convert such knowledge into commonsense knowledge in the format of other commonsense knowledge bases such as ConceptNet and ATOMIC.
    \item \textbf{Definition of ASER}: We define a brand new knowledge graph (KG) where the primitive units of semantics are eventualities. We organize our KG as a relational graph of hyperedges. 
Each eventuality instance is a hyperedge connecting several vertices, which are words. A relation between two eventualities in our KG represents one of the 14 relation types defined in PDTB~\cite{prasad2007penn} or a co-occurrence relation.
    \item \textbf{Scalable Extraction of ASER}: We perform eventuality extraction over large-scale corpora. We design several high-quality patterns based on dependency parsing results to extract all eventualities that match these patterns and then apply a discourse parsing system to extract the eventuality relations. In the end, we leverage a conceptualization module to generalize the extracted knowledge to unseen eventualities.
\item \textbf{Inference over ASER:} We also provide several ways of commonsense inference over ASER.
We show that both eventuality and relation retrieval over one-hop or multi-hop relations can be modeled as conditional probability inference problems.
\item \textbf{Evaluation and Applications of ASER:} We conduct an extensive evaluation to demonstrate the quality of extracted eventuality knowledge and the transferability from such linguistic-based knowledge to commonsense knowledge.
\end{enumerate}


\begin{figure}
    \centering
    \includegraphics[width=0.8\linewidth]{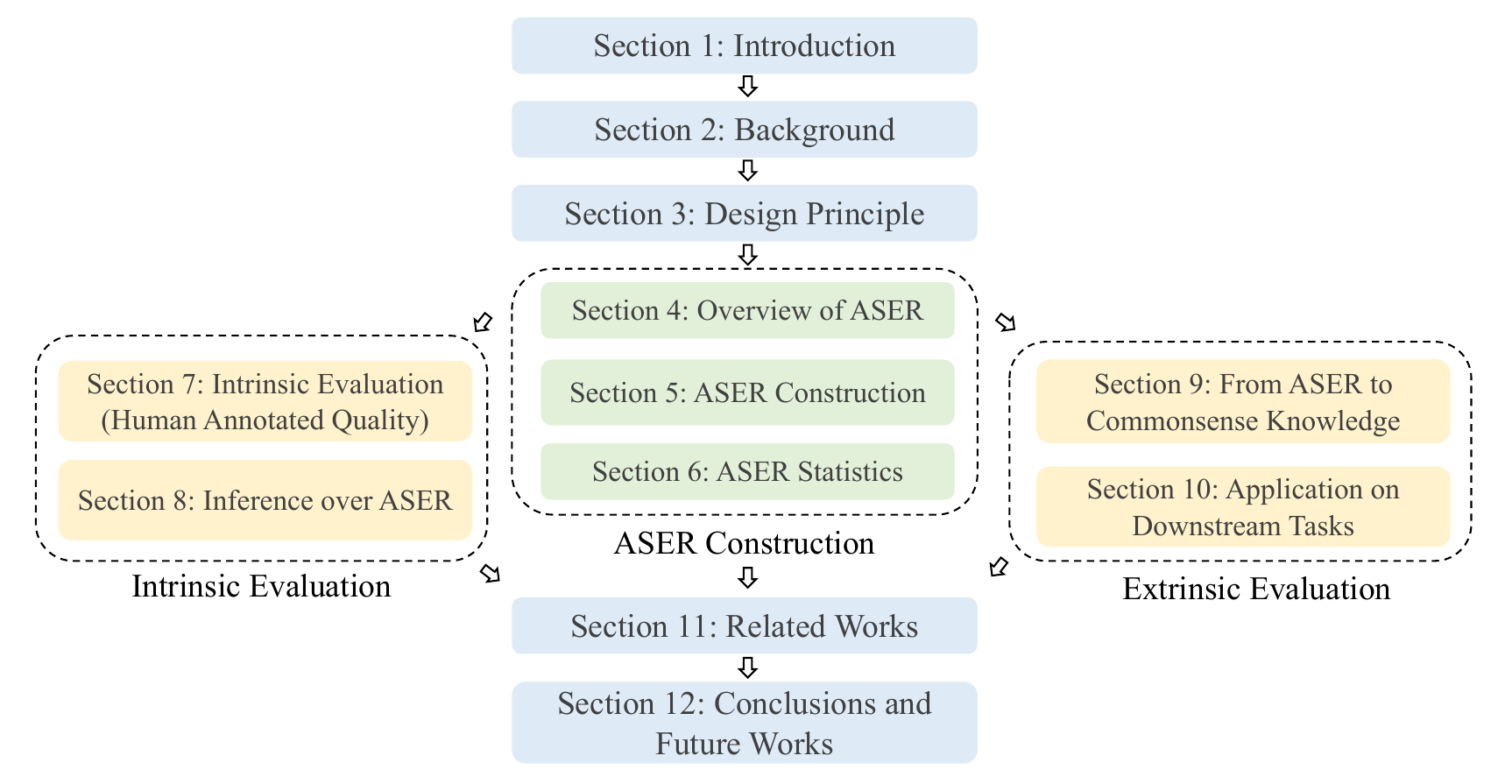}
    \caption{Paper Road Map.}
    \label{fig:road_map}
\end{figure}

The paper organization is presented in Figure~\ref{fig:road_map}.
After the introduction section, we introduce background knowledge about previous works on large-scale knowledge bases construction in Section~\ref{sec:background}. And then, in Section~\ref{sec:design_principles}, we introduce the design principles of ASER. Based on these principles, we present the construction details of ASER in Section \ref{sec:aser-concepts}, \ref{sec:aser-construction}, and \ref{sec:statistics}. Specifically, in Section \ref{sec:aser-concepts}, we show the overall framework and all used notations in ASER. After that, we discuss how to extract those eventualities, eventuality relations, and concept-level eventuality knowledge in Section \ref{sec:aser-construction}. All statistics including the number of unique eventualities and edges are presented in Section \ref{sec:statistics}.
After constructing ASER, we conduct both intrinsic and extrinsic evaluations and analyses to analyze the quality of our knowledge base.
In Section~\ref{sec:intrinsic-evaluation}, we randomly sample eventualities and edges from ASER and invite crowdsourcing annotators from the Amazon Mechanical Turk to annotate the quality.
To better understand the ASER knowledge, we conduct an in-depth inspection of potential inference on ASER knowledge in Section \ref{sec:inference}.
To prove that ASER can indeed cover rich commonsense knowledge, we conduct experiments in Section~\ref{sec:transferability} to demonstrate the transferability from SP knowledge to commonsense knowledge defined in other human-defined commonsense knowledge bases such as ConceptNet~\cite{liu2004conceptnet} and ATOMIC~\cite{Maarten2019Atomic}.
After that, in Section \ref{sec:extrinsic-evaluation}, we show that the knowledge in ASER could be helpful for downstream tasks such as commonsense reading comprehension and daily dialogue generation.
In Section~\ref{sec:related_work}, we introduce the related works about commonsense knowledge acquisition, traditional syntactic-based information extraction, and conceptualization.
In the end, we conclude this paper and introduce all the released resources with Section~\ref{sec:conclusion}.

\section{Background}\label{sec:background}

In his conceptual semantics theory, Ray Jackendoff, a Rumelhart Prize\footnote{The David E. Rumelhart Prize is funded for contributions to the theoretical foundations of human cognition.} winner, describes semantic meaning as ``a finite set of mental primitives and a finite set of principles of mental combination~\cite{Jackendoff}.'' The primitive units of semantic meanings include {\it Thing} (or {\it Object}), {\it Activity},\footnote{In his original book, he called it {\it Action}. But given the other definitions and terminologies we adopted~\cite{ALEXANDER1978,bach1986algebra}, it means {\it Activity}.} {\it State}, {\it Event}, {\it Place}, {\it Path}, {\it Property}, {\it Amount}, etc.
Understanding the semantics related to the world requires the understanding of these units and their relations.
Traditionally, linguists and domain experts built knowledge graphs (KGs)\footnote{Traditionally, people used the term ``knowledge base'' to describe the database containing human knowledge. In 2012, Google released its knowledge graph where vertices and edges in a knowledge base are emphasized. We discuss in the context of the knowledge graph, as our knowledge is also constructed as a complex graph. For more information about terminologies, please refer to~\cite{DBLP:conf/i-semantics/EhrlingerW16}.} to formalize these units and enumerate categories (or senses) and relations of them.
Typical KGs include WordNet~\cite{WordNet} for words, FrameNet~\cite{framenet} for events, and Cyc~\cite{researchCyc} and ConceptNet~\cite{liu2004conceptnet} for commonsense knowledge. 
However, their small scales restricted their usage in real-world applications.

Nowadays, with the growth of web contents, computational power, and the availability of crowdsourcing platforms, many modern and large-scale KGs, such as Freebase~\cite{freebase}, KnowItAll~\cite{knowitall}, TextRunner~\cite{BankoCSBE07}, YAGO~\cite{YAGO,HoffartSBW13}, BabelNet~\cite{NavigliP12},  DBpedia~\cite{auer2007dbpedia}, NELL~\cite{NELL}, Probase~\cite{wu2011taxonomy}, and Google Knowledge Vault~\cite{dong2014knowledge}, have been built based on semi-automatic mechanisms.
Most of these KGs are designed and constructed based on facts about {\it Things} or {\it Objects}, such as instances and their concepts, named entities and their categories, as well as their properties and relations.
On top of them, a lot of semantic understanding problems such as question answering~\cite{BerantCFL13} can be supported by grounding natural language texts on knowledge graphs, e.g., asking a bot for {the nearest restaurants for lunch}.
Nevertheless, these KGs may fall short in circumstances that require not only fact knowledge about {\it Things} or {\it Objects}, but also the commonsense knowledge about {\it Activities}, {\it States}, and {\it Events}.
Consider the aforementioned utterance that a human would talk to the bot at 1 PM: ``I am hungry,'' which may also imply one's need for a restaurant recommendation. This, however, will not be possible unless the bot can identify that the consequence of being hungry would be {``having lunch''} at noon.

In this paper, we propose to leverage higher-order selectional preference to discover and store commonsense knowledge about {\it Activities} (or process, e.g., ``I sleep''), {\it States} (e.g., ``I am hungry''), {\it Events} (e.g., ``I make a call''), and their {\it Relations} (e.g., ``I am hungry'' may result in ``I have lunch''), for which we call ASER. 
In fact,  {\it Activities}, {\it States}, and {\it Events}, which are expressed by verb-related clauses, are all eventualities following the commonly adopted terminology and categorization proposed by Mourelatos~\cite{ALEXANDER1978} and Bach~\cite{bach1986algebra}.
Previous literature on eventualities mostly focuses on extracting eventualities from text with pre-defined event schemas, which enumerates triggers with senses and arguments with roles, defined in FrameNet~\cite{framenet} or ACE~\cite{NIST05}.
However, as the pre-defined event ontology is often domain-specific and small (e.g., ACE contains 33 event types), the extracted events cannot cover all commonsense.
Different from them, instead of using a small event ontology, we use patterns over the dependency graphs, which could contain multiple words and dependency edges, to extract eventualities.
Any events that satisfy the pre-defined patterns will be extracted, and thus we achieve much broader coverage.
Besides the eventuality extraction, extracting relations between eventualities is another vital research problem in the NLP community.
For example, HieVe~\cite{DBLP:conf/lrec/GlavasSMK14} focuses on extracting super-sub event relations, and TimeBank~\cite{pustejovsky2003timebank} focuses on the temporal relations.
These works typically focus on identifying implicit relations, which is a very challenging task, and the state-of-the-art models can only achieve 59.5 F1~\cite{DBLP:conf/emnlp/WangCZR20} and 75.5 F1~\cite{DBLP:conf/emnlp/HanNP19} on the HieVe and TimeBank datasets, respectively.
As a result, current models are still not ready to be used to extract high-quality relations between events.
At the same time, the current state-of-the-art model on implicit discourse relations can only achieve the accuracy of 57.26~\cite{DBLP:conf/ijcai/LiuOSJ20} on CoNLL-2015 dataset~\cite{xue2015conll}.
As an alternative, we discard the implicit relations and only focus on explicit discourse relations between events. By doing so, we sacrifice the recall but make sure the high-quality of the collected knowledge. 
For example, the used discourse parser proposed by 
\cite{DBLP:conf/conll/WangL15}
can guarantee 90.14\% F1 on explicit discourse relation classification.
Simultaneously, we try to scan a huge corpus to guarantee the resulting knowledge graph's overall coverage.

\section{Design Principles}\label{sec:design_principles}

As aforementioned, ASER is a large-scale eventuality-based knowledge graph.
Here by eventuality, we mean {\it Activities}, {\it States}, and {\it Events}, which are defined based on the commonly adopted terminology and categorization proposed by Mourelatos~\cite{ALEXANDER1978} and Bach~\cite{bach1986algebra}:
\begin{itemize}
    \item {\bf Activity: } An activity is also called a process~\cite{ALEXANDER1978,bach1986algebra}. Both activity and event are occurrences (actions) described by active verbs. An example is ``The coffee machine is brewing coffee.'' 
    \item {\bf State:} A state is usually described by a stative verb and cannot be qualified as actions. 
    A typical state expression is ``The coffee machine is ready for brewing coffee.''
    \item {\bf Event:} An event is defined as an occurrence that is inherently countable~\cite{ALEXANDER1978}. For example, ``The coffee machine brews a cup of coffee once more'' is an event 
  because it admits a countable noun ``a cup'' and cardinal count adverbials ``once,'' while ``The coffee machine brews coffee'' is not an event with an imperfective aspect which is not countable.
 \end{itemize}

Unlike the previous works~\cite{DBLP:journals/coling/SiegelM00,DBLP:conf/emnlp/ZhangCWSR20}, we do not distinguish activities (or processes), states, and events. Instead, we use dependency patterns to represent all the eventualities that can be activities, states, and events and also discourse relations~\cite{prasad2007penn} such as \textit{COMPARISON.Contrast} and \textit{CONTINGENCY.Cause} as the relation types between eventualities based on the following two design principles.

\begin{figure}
    \centering
    \includegraphics[width = 0.7\linewidth]{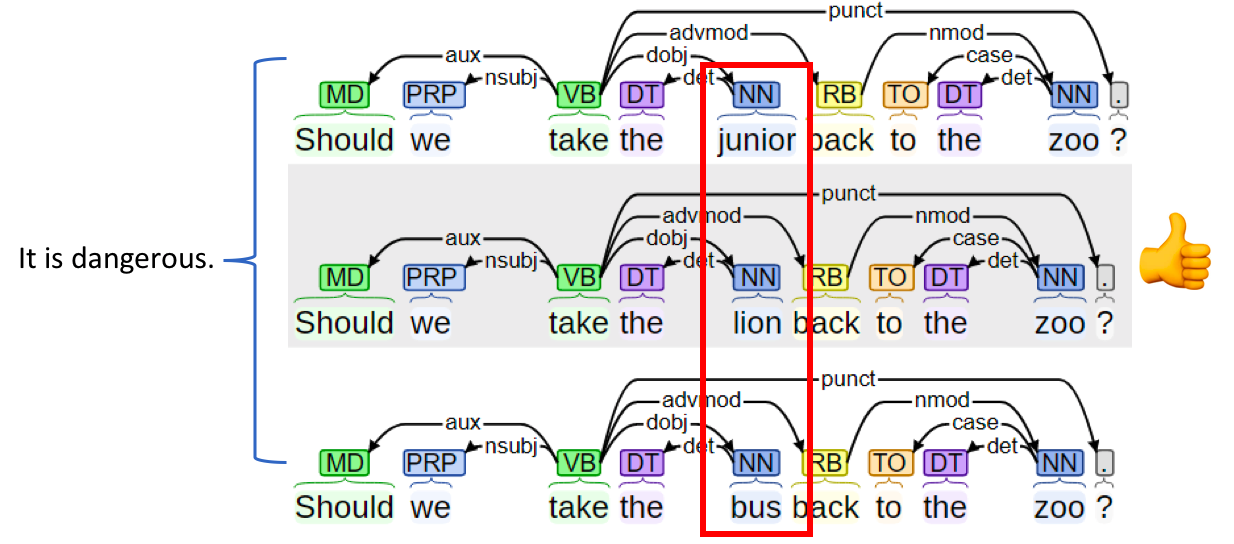}
    \caption{Principle Demonstration. When we fix the grammar, humans' preferences over the linguistic description are reflecting the commonsense. The examples are based on ones used in~\cite{katz1963structure}.}
    \label{fig:principle_demo}
\end{figure}

\subsection{The Lower Bound of a Semantic Theory}


As discussed by the lower bound of a semantic theory~\cite{katz1963structure}, understanding human language requires both knowledge about the language (i.e., grammar) and knowledge about the world.
As a result, if we fix the grammar structure of the linguistic descriptions, their difference will be mostly captured by the semantics.
An example is shown in Figure~\ref{fig:principle_demo}. There are three sentences that share the same grammar structure but describe different events, which may have different reasons, effects, and sub-events.
Given that the previous context is ``It is dangerous,'' humans normally will prefer the second sentence to appear in this context because a lion is a dangerous animal. 
And such preference can reflect the commonsense knowledge we are looking for.
Motivated by this, we propose to use dependency patterns to categorize eventualities and discourse relations as the relations between eventualities.
As a result, the frequency of eventualities and edges can be naturally used to represent humans' preferences when the grammar structure is fixed.

\begin{table*}[t]
\small
	\centering
    \subtable[\texttt{dobj}]{
      \begin{tabular}{c|c}
            \toprule 
               SP  Pair & Plausibility \\
            \midrule
              (eat, meal)  & 10.00 \\
              (close, door) & 8.50 \\
              (convince, people) & 7.75 \\
              (touch, food) & 5.50 \\
              (hate, investment) & 4.00 \\
              (confront, impulse) & 2.78 \\

              (eat, mail) & 0.00 \\
                \bottomrule
            \end{tabular}
    }
    \subtable[\texttt{nsubj}]{
      \begin{tabular}{c|c}
            \toprule 
               SP  Pair & Plausibility \\
            \midrule
              (singer, sing)  & 10.00 \\
              (law, permit) & 7.78 \\
              (women, pray) & 5.83 \\
              (realm, remain) & 3.06 \\
              (victim, contain) & 2.22 \\
              (bar, act) & 1.39 \\
              (textbook, eat) & 0.00 \\

                \bottomrule

            \end{tabular}
    }
    \subtable[\texttt{amod}]{
      \begin{tabular}{c|c}
            \toprule 
              SP   Pair & Plausibility \\
            \midrule
              (fresh, air)  & 9.77 \\
              (new, method) & 8.89 \\
              (young, people) & 6.82 \\
              (medium, number) & 4.09\\
              (immediate, food) & 2.50\\
              (eager, price) & 1.36 \\
              (secret, wind) & 0.75 \\
                \bottomrule
            \end{tabular}
    }
    \subtable[\texttt{dobj\_amod}]{
      \begin{tabular}{c|c}
            \toprule 
              SP   Pair & Plausibility \\
            \midrule
              (lift, heavy \textit{object})  & 9.17 \\
              (design, new \textit{object}) & 8.00 \\
              (recall, previous \textit{object})  & 7.05 \\
              (attack, small \textit{object}) & 5.23 \\
              (drag, drunk \textit{object}) & 4.25 \\
              (inform, weird \textit{object}) & 3.64 \\
              (earn, rubber \textit{object})  & 0.63 \\
                \bottomrule
            \end{tabular}
    }
    \subtable[\texttt{nsubj\_amod}]{
      \begin{tabular}{c|c}
            \toprule 
              SP   Pair & Plausibility \\
            \midrule
              (friendly \textit{subject}, smile)  & 10.00 \\
              (evil \textit{subject}, attack) & 9.00 \\
              (recent \textit{subject}, demonstrate) & 6.00\\
              (random \textit{subject}, bear) & 4.00\\
              (happy \textit{subject}, steal)  & 2.25 \\
              (stable \textit{subject}, understand) & 1.75 \\
              (sunny \textit{subject}, make)  & 0.56 \\
                \bottomrule
            \end{tabular}
    }
	\caption{Examples of first-order and second-order selectional preference and their plausibility ratings provided by human annotators~\cite{zhang2019sp-10k}. \textit{Object} and \textit{subject} are place holders to help understand the second-hop selectional preference relations.} \label{tab:SP10Kdemo}
\end{table*}

Historically, such grammar-based semantics is called selectional preference~\cite{resnik1993selection}, which is a relaxation of selectional restrictions~\cite{katz1963structure}.
Initially, the research on selectional preference focuses on the IsA hierarchy in WordNet~\cite{WordNet} and verb-object dependency relations.
Later on, the idea of selectional preference was extended to verb-subject dependency relations.
Several first-order selectional preference examples are as follows.
\begin{itemize}
  \item \textsf{SP}(Cat, \texttt{IsA}, Animal) $>$ \textsf{SP}(Cat, \texttt{IsA}, Plant)
  \item \textsf{SP}(Eat, \texttt{dobj}, Food) $>$ \textsf{SP}(Eat, \texttt{dobj}, Rock)
  \item \textsf{SP}(Sing, \texttt{nsubj}, Singer) $>$ \textsf{SP}(Sing, \texttt{nsubj}, House)
\end{itemize}

Recently, to represent more complex commonsense knowledge, the principle of selectional preference was extended to the second-order~\cite{zhang2019sp-10k}.
The motivation is that humans tend to have a strong preference over the property of certain verbs' subjects and objects.
For example, we can formalize the commonsense that the subject of eat is more likely to be ``hungry'' rather than ``tasty'' with the following second-order selectional preference:
\begin{itemize}
  \item \textsf{SP}(Eat, \texttt{Nsubj-amod}, Hungry) $>$ \textsf{SP}(Eat, \texttt{Nsubj-amod}, Tasty)
\end{itemize}

More examples are shown in Table~\ref{tab:SP10Kdemo}. Higher plausibility scores indicate that the annotators have a stronger preference for the combination.
For the first-order selectional preference, people are most likely to select ``meal'' rather than ``mail'' as the object of ``eat.'' 
Similarly, we can see that ``heavy'' is a common property of the \textit{object} of ``lift.''
As shown by the experiments in ~\cite{zhang2019sp-10k}, such selectional preference knowledge is crucial for solving commonsense reasoning tasks such as Winograd Schema Challenge~\cite{levesque2011winograd}.

In this work, we further extend the idea of selectional preference to discourse relations between eventualities, which is denoted as higher-order selectional preference over eventualities.


\subsection{The Need of Aggregating ``Partial Information'' in Commonsense Reasoning}

As discussed by~\cite{Wilks1975IAU}, to effectively represent the selectional preference over linguistic relations and use that knowledge for language inference, we need to do aggregation over the ``partial information,'' which may ``not be invariably true'' but ``tends to be of a very high degree of generality indeed''~\cite{liu2004conceptnet}.
For example, Wilks used the following sentence as an example.
\begin{itemize}
  \item The \texttt{soldiers} fired at the \texttt{women}, and we saw several of \texttt{them} fall.
\end{itemize}
We know that \textit{them} should refer to \textit{women} rather than \textit{soldiers} because we have the partial information that ``hurt things tending to fall down.''
Formally, it can be translated into the following form: 
\begin{itemize}
  \item (hurt, \texttt{X}) $\xrightarrow{\rm ResultIn}$ (\texttt{X}, fall).
\end{itemize}

There are many ways to find such representations of knowledge, e.g., first or even second-order logic.
However, existing logic-based or semantic frame-based methods such as combinatory categorial  grammar~\cite{steedman2011combinatory} or semantic role labeling~\cite{framenet,kingsbury2002treebank} require large amounts of annotation.
Moreover, the semantic roles defined in labeled frames~\cite{framenet,kingsbury2002treebank} are too coarse-grained to support fine-grained conceptual reasoning.
An efficient way of acquiring such partial information is to do the aggregation over collected selectional preference about the instance-level eventualities and their conceptualizations.
We have shown that using such higher-order selectional preference, we can solve a subset of Winograd Schema Challenge (WSC)~\cite{levesque2011winograd} with 70\% accuracy~\cite{zhang2019sp-10k}.
For example, to solve the WSC example:
\begin{itemize}
  \item The \texttt{fish} ate the \texttt{worm}. \texttt{It} was tasty.
  \item The \texttt{fish} ate the \texttt{worm}. \texttt{It} was hungry.
\end{itemize}

\noindent we can merge all subjects and object, and get the following frequency information:
\begin{itemize}
  \item \textsf{Frequency}(‘\texttt{X} eats \texttt{Y}’, \texttt{co-occur}, ‘\texttt{X} is hungry’) = 18 and \\ \textsf{Frequency}(‘\texttt{X} eats \texttt{Y}’, \texttt{co-occur}, ‘\texttt{Y} is hungry’) = 1;
  \item \textsf{Frequency}(‘\texttt{X} eats \texttt{Y}’, \texttt{co-occur}, ‘\texttt{X} is tasty’) = 0 and \\ \textsf{Frequency}(‘\texttt{X} eats \texttt{Y}’, \texttt{co-occur}, ‘\texttt{Y} is tasty’) = 7.
\end{itemize}
These numbers reflect the aforementioned second-order selectional preferences based on which we can solve the questions.
Although such aggregation has been shown to be useful for the Winograd Schema Challenge, the collected partial information can be too coarse.
It only aggregates all information to be \texttt{X} or \texttt{Y}.
However, in real-world applications, we also need to know the following question for fine-grained concepts other than humans 
\begin{itemize}
  \item \textsf{Frequency}(‘\texttt{Company} acquires \texttt{Startups}’, \texttt{ResultIn}, ‘\texttt{Stock} increases’)=? 
\end{itemize}
Therefore, a principled way of performing the conceptualization of instances and partial concept information aggregation is needed.
Thus, we propose to leverage another existing knowledge base, Probase~\cite{wu2011taxonomy}, to perform conceptualization~\cite{SongWWLC11,SongWW15} over entities that Probase can recognize. 
For example, after observing that both ``having a cat'' and ``having a dog'' can cause ``being happy,'' and with the help of Probase, we can conceptualize and aggregate ``having a cat'' and ``having a dog'' to be ``having a pet,'' we can then conclude that ``having a pet'' can cause ``being happy.''

One thing worth mentioning is that the main methodology of ASER is that after the aggregation, the more heavily weighted (i.e., frequent) eventualities or edges make more sense than the less heavily weighted ones.
As a result, when we conduct the conceptualization, we do not need to consider the context because other eventualities exist, and after the aggregation, the more heavily weighted ones will still make more sense. For example, given the eventuality ``I eat apple,'' we do not need to worry about which one of ``fruit'' and ``company'' we should conceptualize ``apple'' to because we will see many other eventualities related to fruit such as ``I eat banana'' and ``I eat orange.'' In the end, the overall weight of ``I eat fruit'' will still be much higher than ``I eat company.''

\section{Overview of ASER}\label{sec:aser-concepts}




ASER is a hybrid graph combining a hypergraph $\{\VM, \EM\}$ where each hyperedge is constructed over vertices, and a traditional graph $\{\EM, \RM\}$ where each edge is built among eventualities.
For example, $E_h$=\texttt{(I, am, hungry)} and $E_t$=\texttt{(I, eat, anything)} are eventualities, where we omit the internal dependency structures for brevity.
They have a relation $\langle E_h, \texttt{Result}, E_t \rangle$, where \texttt{Result} is the relation type.
We devise the formal definition of ASER as below.

\begin{definition}
{\bf ASER KG} is a hybrid graph $\HM$ of eventualities $E$'s. Each {\bf eventuality} $E$ is a hyperedge linking to a set of vertices $v$'s.
Each {vertex} $v$ is a {\bf word} in the vocabulary.
We define $v\in \VM$ in the vertex set and $E \in \EM$ in the hyperedge set.
$\EM \subseteq \PM(\VM)\setminus \{\emptyset\}$ is a subset of the power set of $\VM$.
We also define a {\bf relation} $R_{i,j}\in \RM$ between two eventualities $E_i$ and $E_j$, where $\RM$ is the relation set.
Each relation has a {\bf type} $T\in \TM$ where $\TM$ is the type set.
Overall, we have ASER KG $\HM=\{\VM, \EM, \RM, \TM\}$.
\end{definition}


        
\subsection{Eventuality}\label{sec:eventuality-pattern}

Unlike named entities or concepts, which are noun phrases, eventualities are usually expressed as verb phrases, which are more complicated in structure. 
Our definition of eventualities is built upon the following two assumptions:
(1) Syntactic English patterns are relatively fixed and consistent; (2) The eventuality's semantic meaning is determined by the words it contains.
To avoid the extracted eventualities being too sparse, we use words fitting specific patterns rather than a whole sentence to represent an eventuality.
Also, to make sure the extracted eventualities have complete semantics, 
we retain all necessary words extracted by patterns rather than those simple verbs or verb-object pairs in sentences.
The selected patterns are shown in Table~\ref{tab:eventuality-pattern}.
For example, for the eventuality \texttt{(dog, bark)}, we have a relation \texttt{nsubj} between the two words to indicate that there is a subject-of-a-verb relation in between.
We now formally define an eventuality as follows.
\begin{definition}
An eventuality $E$ is a hyperedge linking multiple words $\{v_1, \ldots, v_N \}$, where $N$ is the number of words in eventuality $E$. Here, $v_1, \ldots, v_N\in \VM$ are all in the vocabulary. A pair of words in $E$ $(v_i,v_j)$ may follow a syntactic relation $e_{i,j}$. 
The weight of $E$, denoted as $w_{E}^{(e)}$, is defined by the frequencies of appearance in the whole corpora.
\end{definition}

\begin{table}[t]
\small
	\centering
	{
	\begin{tabular}{p{0.44\textwidth}|c|p{0.38\textwidth}}
		\toprule 
		Pattern & Code & Example \\
		\midrule
		$n_1$-\texttt{nsubj}-$v_1$ &s-v& ``The dog barks'' \\
		$n_1$-\texttt{nsubj}-$v_1$-\texttt{dobj}-$n_2$ &s-v-o& ``I love you'' \\
        $n_1$-\texttt{nsubj}-$v_1$-\texttt{xcomp}-$a$ &s-v-a& ``He felt ill'' \\

        $n_1$-\texttt{nsubj}-$v_1$-\texttt{xcomp}-$v_2$ &s-v-v&``I want to go'' \\
        $n_1$-\texttt{nsubj}-($v_1$-\texttt{iobj}-$n_2$)-\texttt{dobj}-$n_3$ &s-v-o-o& ``You give me the book''\\
        $n_1$-\texttt{nsubj}-$v_1$-\texttt{xcomp}-$v_2$-\texttt{dobj}-$n_2$ &s-v-v-o&``I want to eat the apple'' \\
        $n_1$-\texttt{nsubj}-($v_1$-\texttt{dobj}-$n_2$)-\texttt{xcomp}-$v_2$-\texttt{dobj}-$n_3$& s-v-o-v-o & ``I ask you to help us''\\
        $n_1$-\texttt{nsubj}-($v_1$-\texttt{dobj}-$n_2$)-\texttt{xcomp}-($v_2$-\texttt{iobj}-$n_3$)-\texttt{dobj}-$n_4$& s-v-o-v-o-o & ``president urges the congress to make her citizen''\\
        $n_1$-\texttt{nsubj}-$a_1$-\texttt{cop}-$be$ &s-be-a& ``The dog is cute'' \\
        $n_1$-\texttt{nsubj}-$n_2$-\texttt{cop}-$be$ &s-be-o& ``He is a boy'' \\
        $n_1$-\texttt{nsubj}-$v_1$-\texttt{xcomp}-$n_2$-\texttt{cop}-$be$ &s-v-be-o& ``I want to be a hero''\\
        $n_1$-\texttt{nsubj}-$v_1$-\texttt{xcomp}-$a_1$-\texttt{cop}-$be$ &s-v-be-a& ``I want to be slim''\\
		$n_1$-\texttt{nsubj}-($v_1$-\texttt{iobj}-$n_2$)-\texttt{xcomp}-$n_3$-\texttt{cop}-$be$& s-v-o-be-o & ``I want her to be hero''\\
        $n_1$-\texttt{nsubj}-($v_1$-\texttt{iobj}-$n_2$)-\texttt{xcomp}-$a_1$-\texttt{cop}-$be$& s-v-o-be-a & ``I want her to be happy''\\
        $there$-\texttt{expl}-$be$-\texttt{nsubj}-$n_1$& there-be-o & ``There is an apple''\\
        $n_1$-\texttt{nsubjpass}-$v_1$ &spass-v& ``The bill is paid''\\
        $n_1$-\texttt{nsubjpass}-$v_1$-\texttt{dobj}-$n_2$ & spass-v-o & ``He is served water''\\
        $n_1$-\texttt{nsubjpass}-$v_1$-\texttt{xcomp}-$v_2$-\texttt{dobj}-$n_2$& spass-v-v-o & ``He is asked to help us''\\

		\bottomrule
	\end{tabular}
	}
	\caption{Selected eventuality patterns (``v'' stands for normal verbs other than ``be,'' ``be'' stands for ``be'' verbs, ``n'' stands for nouns, ``a'' stands for adjectives, and ``p'' stands for prepositions.), Code (to save space, we create a unique code for each pattern and will use that in the rest of this paper), and the corresponding examples. 
	} \label{tab:eventuality-pattern}
\end{table}


We use patterns from dependency parsing to extract eventualities $E$'s from unstructured large-scale corpora.
Here $e_{i,j}$ is one of the relations that dependency parsing may return.
Although the recall is sacrificed in this way, our patterns are of high precision, and we use massive corpora to extract as many eventualities as possible. 
This strategy is also shared with many other modern KGs~\cite{knowitall,BankoCSBE07,NELL,wu2011taxonomy}.

\begin{table}[t]
    \small
	\centering
	{\
	\begin{tabular}{p{0.24\textwidth}|p{0.68\textwidth}}
		\toprule 
		Relation & Explanation  \\
		\midrule
		$\langle E_h, \texttt{Precedence}, E_t \rangle$ & $E_h$ happens before $E_t$. \\
        $\langle E_h, \texttt{Succession}, E_t \rangle$ & $E_h$ happens after $E_t$. \\
        $\langle E_h, \texttt{Synchronous}, E_t \rangle$ & $E_h$ happens at the same time as $E_t$. \\ \midrule
        $\langle E_h, \texttt{Reason}, E_t \rangle$ & $E_h$ happens because $E_t$ happens. \\
        $\langle E_h, \texttt{Result}, E_t \rangle$ & If $E_h$ happens, it will result in the happening of $E_t$. \\
        $\langle E_h, \texttt{Condition}, E_t \rangle$ & Only when $E_t$ happens, $E_h$ can happen. \\ \midrule
        $\langle E_h, \texttt{Contrast}, E_t \rangle$ & $E_h$ and $E_t$ share a predicate or property and have significant difference on that property. \\
        $\langle E_h, \texttt{Concession}, E_t \rangle$ & $E_h$ should result in the happening of $E^\prime$, but $E_t$ indicates the opposite of $E^\prime$ happens. \\ \midrule
        $\langle E_h, \texttt{Conjunction}, E_t \rangle$ & $E_h$ and $E_t$ both happen. \\
        $\langle E_h, \texttt{Instantiation}, E_t \rangle$ & $E_t$ is a more detailed description of $E_h$. \\
        $\langle E_h, \texttt{Restatement}, E_t \rangle$ & $E_t$ restates the semantics meaning of $E_h$.\\ 
        $\langle E_h, \texttt{Alternative}, E_t \rangle$ & $E_h$ and $E_t$ are alternative situations of each other. \\
        $\langle E_h, \texttt{ChosenAlternative}, E_t \rangle$ & $E_h$ and $E_t$ are alternative situations of each other, but the subject prefers $E_h$. \\
        $\langle E_h, \texttt{Exception}, E_t \rangle$ & $E_t$ is an exception of $E_h$. \\ \midrule
        $\langle E_h, \texttt{Co-Occurrence}, E_t \rangle$ & $E_h$ and $E_t$ appear in the same sentence. \\
		\bottomrule
	\end{tabular}
	}
	\caption{Eventuality relation types between two eventualities $E_h$ and $E_t$ and explanations.}	\label{tab:relation-def}
\end{table}
        
\subsection{Eventuality Relation}

For relations among eventualities, as introduced in Section~\ref{sec:introduction}, we follow PDTB's~\cite{prasad2007penn} definition of relations between sentences or clauses but simplify them to eventualities.
Following the CoNLL 2015 discourse parsing shared task~\cite{xue2015conll}, we select 14 discourse relation types and an additional co-occurrence relation to build our knowledge graph.

\begin{definition}\label{def:relations}
A relation $R$ between a pair of eventualities $E_h$ and $E_t$ has one of the following types $T \in \TM$ and all types can be grouped into five categories: 
{\bf Temporal} (including {Precedence}, {Succession}, and {Synchronous}),
{\bf Contingency} (including {Reason}, {Result}, and {Condition}),
{\bf Comparison} (including {Contrast} and {Concession}),
{\bf Expansion} (including {Conjunction}, {Instantiation}, {Restatement}, {Alternative}, {ChosenAlternative}, and {Exception}), and
{\bf Co-Occurrence}.
The detailed definitions of these relation types are shown in Table~\ref{tab:relation-def}.
The weight of $R = \langle E_h, T, E_t \rangle$, which is denoted as $w_{R}^{(r)}$, is defined by the sum of weights of $\langle E_h, T, E_t \rangle$ that appear in the whole corpora.
\end{definition}






\subsection{ASER Conceptualization}

As aforementioned, to overcome the challenge that trivial commonsense is often omitted in humans' communication, we propose to leverage the conceptualization to generalize the knowledge about observed eventualities to unseen ones.
For each eventuality $E \in \EM$, whose weight is $w_{E}^{(e)}$, and we can conceptualize $E$ to $E^\prime$ with confidence $w_{E, E^\prime}^{(c)}$, we will get a new conceptualized eventuality $\E^\prime$ with the weight $w_{E^\prime}^{(c)} = w_E^{(e)} \cdot w_{E, E^\prime}^{(c)}$.
Similarly, assume that an edge $R \in \RM$ is $ \langle E_h, T, E_t \rangle$ and its weight is $w_{R}^{(r)}$, and $E_h$ and $E_t$ can be conceptualized to $E_h^\prime$ and $E_t^\prime$ with the confidence $w_{E_h, E_h^\prime}^{(c)}$ and $w_{E_t, E_t^\prime}^{(c)}$, respectively.
We can then get a new conceptualized edge $\langle E_h^\prime, T, E_t^\prime \rangle$ with the weight $w_{R}^{(r)} \cdot w_{E_h, E_h^\prime}^{(c)} \cdot w_{E_t, E_t^\prime}^{(c)}$.
Details about how to leverage an external hypernym knowledge base to get the conceptualized eventualities and determine the confidence scores are presented in Section~\ref{sec:aser-construction}.

\subsection{KG Storage}

In total, we use the following three tables of the SQLite database to store ASER. 
\begin{itemize}
    \item \textit{Eventuality}: As aforementioned, all eventualities in ASER are dependency graphs, where vertices are the words and edges are dependency relations.
We generate unique ``eids'' for eventualities by hashing their words, pos-tags, and dependencies and store eventualities in an \textit{Eventuality} table with SQLite database where ``eids'' is the key, and patterns, verb(s), skeleton words, words, pos-tags, dependencies, and frequencies are the other attribute columns.
\item \textit{Concept}: To effectively distinguish the eventualities before and after the conceptualization, we store eventualities created by the conceptualization step in another \textit{Concept} table and denote the id as ``cid.'' 
As the dependency edges are inherited from the original eventualities, we only hash the conceptualized words to generate the ``cids.'' For each conceptualized eventuality, we store its ``cid,'' pattern, frequency, and ``eids'' of the original eventualities.  
\item \textit{Relations}: We store the relations between eventualities in the \textit{Relations} table. For each pair of eventualities (i.e., $E_h$ and $E_t$), if there is at least an edge between them, we will create an instance and generate a ``rid'' for them by hashing the concatenation of their ``eids.'' For the storage efficiency and retrieval feasibility, we store all edges and the associated weights between $E_h$ and $E_t$ as well as the eventuality ids of $E_h$ and $E_t$ in that instance.
\end{itemize}



\section{ASER Construction}\label{sec:aser-construction}
In this section, we introduce the ASER construction details.


\begin{figure}[!t]
\centering
\includegraphics[width=0.75\linewidth]{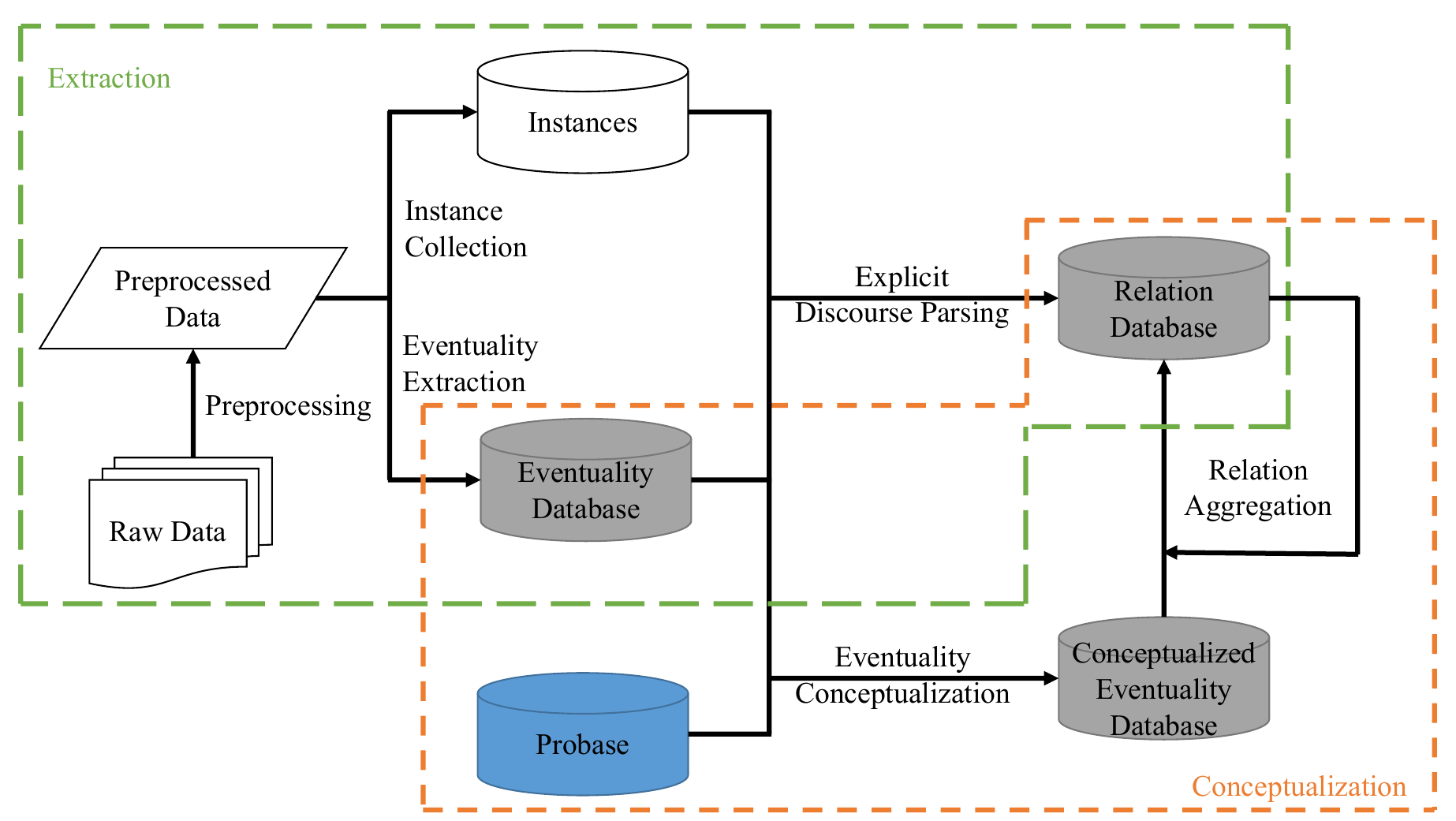}
\caption{ASER construction framework. The extraction and the conceptualization process are shown in the orange dash-dotted and green dashed boxes, respectively. The blue database is Probase and three gray databases are the resulted ASER.}
\label{fig:framework}
\end{figure}

\subsection{System Overview}\label{sec:system-overview}

The overall framework of our extraction system is shown in Figure~\ref{fig:framework}.
After collecting the raw corpora, we first preprocess the texts with the dependency parser.
Then we perform eventuality extraction with pattern matching.
We collect sentences and adjacent sentence pairs that contain more than two eventualities into an instance collection.
After that, we extract discourse relations from these candidate instances with the help of an explicit discourse parser~\cite{DBLP:conf/conll/WangL15}.
Considering that the discourse parser's discourse argument span might not be identical to the extracted eventualities, we apply token-based Simpson's similarity between the arguments spans and eventualities to determine whether the discourse arguments are enough to represent the meaning of the extracted eventualities. We only keep the extraction results with the Simpson's similarity larger than 0.8.
After the initial ASER construction, we leverage the \textit{IsA} relations between nouns and named entities from Probase~\cite{wu2011taxonomy} to conduct the conceptualization.
In the end, we aggregate relations between conceptualized eventualities by retrieving head and tail eventualities from the conceptualized eventuality database and the eventuality relation database.
In the following sub-sections, we will introduce each part of the system separately.

\subsection{Corpora}

To ensure the broad coverage of ASER, we select corpora from different resources (reviews, news, forums, social media, movie subtitles, e-books) as the raw data. The details of these datasets are as follows.

$\bullet$ Yelp: Yelp is a social media platform where users can write reviews for businesses (e.g., restaurants). The latest release of the Yelp dataset\footnote{\url{https://www.yelp.com/dataset/challenge}} contains over five million reviews. 

$\bullet$ New York Times (NYT): The NYT~\cite{nyt} corpus contains over 1.8 million news articles from the NYT throughout 20 years (1987 - 2007).

$\bullet$ Wiki: Wikipedia is one of the largest free knowledge datasets. To build ASER, we select the English version of Wikipedia.\footnote{\url{https://dumps.wikimedia.org/enwiki/}}

$\bullet$ Reddit: Reddit is one of the largest online forums. In this work, we select the anonymized post records\footnote{\url{https://www.reddit.com/r/datasets/comments/3bxlg7}} over one period month.  

$\bullet$ Movie Subtitles: The movie subtitles corpus was collected by~\cite{lison2016opensubtitles2016}, and we select the English subset, which contains subtitles for more than 310K movies.

$\bullet$ E-books: The last resource we include is the free English electronic books from Project Gutenberg.\footnote{\url{https://www.gutenberg.org/}}

We merge these resources as a whole to perform the knowledge extraction. The detailed statistics are presented in Table~\ref{tab:corpora-statistics}.

\begin{table}[!t]
    \small
	\centering
		
	{
	\begin{tabular}{c|c|c|c|c}
		\toprule 
		Name   & \# Sentences & \# Tokens    & Corpus Size & Category \\
		\midrule
		YELP & 54.5 M & 838.8 M  & 2.5G  & Reviews \\ %
		NYT & 49.8 M & 1,179.4 M   & 3G  & News \\  %
        Wiki & 110.6 M & 2,435.4 M  & 13G & Knowledge  \\ %
        Reddit & 253.6 M & 3,371.3 M  & 21G  & Forum  \\ %
        Subtitles & 444.6 M & 3,229.4 M   & 13G  & Movie Scripts \\ %
        E-books & 210.6 M & 3,610.0 M    & 21G  & Stories \\ %
        \midrule
        Overall & 1,123.7 M & 14,664.2 M  & 73.5G & -\\
		\bottomrule
	\end{tabular}
	}
	\caption{Statistics of used corpora. (M means millions and G means Gigabytes.)} \label{tab:corpora-statistics}	
\end{table}

\subsection{Preprocessing}\label{sec:preprocessing}

For each document, we aim to extract eventualities, relations between eventualities, conceptualized eventualities, and relations between conceptualized eventualities.
Based on the consideration of the text parsing complexity and quality, we parse each paragraph\footnote{As the discourse parser extracts discourse relations by the constituency tree of a sentence or trees of adjacent sentences, parsing sentences one by one would miss or misclassify some discourse relations.} instead of a whole document with the CoreNLP tool\footnote{\url{https://stanfordnlp.github.io/CoreNLP}} to acquire the lemmatized tokens, pos-tags, named entities, the dependency graph, and the constituency tree.
Before parsing, we replace URLs with a special token $\langle$\textit{URL}$\rangle$ and drop tables in Reddit data.

\subsection{Eventuality Extraction}\label{sec:eventuality-extraction}

\begin{algorithm}[t]\caption{Eventuality Extraction with One Pattern $p$}\label{algorithm:eventuality-extraction}
	\textbf{INPUT:} Parsed dependency graph $\DM$, center verb $v$, positive dependency edges $\PM_p^{(pos)}$, optional edges $\PM_p^{(opt)}$, and negative edges $\PM_p^{(neg)}$.\\
    \textbf{OUTPUT:} extracted eventuality $E$.
    \begin{algorithmic}[1]
    \State Initialize eventuality edge list $\DM^\prime$.
    \State Set the center verb $v$ as the $v_1$ in the pattern $p$
    \For{each connection $d$ (a relation and the associated word) in positive dependency edges $\PM_p^{(pos)}$}  
        \If{find $d$ in $\DM$}
            \State Append $d$ in $\DM^\prime$.
        \Else
        	\State Return \textit{null}.
        \EndIf 
    \EndFor
    \For{each connection $d$ in optional dependency edges $\PM_p^{(opt)}$}  
        \If{find $d$ in $\DM$}
              \State Append $d$ in $\DM^\prime$.
        \EndIf 
    \EndFor
    \For{each connection $d$ in negative dependency edges $\PM_p^{(neg)}$}  
      \If{find $d$ in $\DM$}
          \State Return \textit{null}.
      \EndIf 
    \EndFor
    \State Build eventuality instance $E$ from $\DM^\prime$.
    \State Return $E$.
    \end{algorithmic}
\end{algorithm}

To ensure that all the extracted eventualities are semantically complete without being too complicated, we design 18 patterns to extract the eventualities via pattern matching.
Each of the patterns contains three kinds of dependency edges: positive dependency edges, optional dependency edges, and negative dependency edges.
All the positive edges are shown in Table~\ref{tab:eventuality-pattern}.
Six more dependency relations (\texttt{advmod}, \texttt{amod}, \texttt{nummod}, \texttt{aux}, \texttt{compound}, and \texttt{neg}) are optional dependency edges that can associate with any of the selected patterns.
We omit all optional edges in the table because they are the same for all patterns.
All other dependency edges are considered negative dependency edges, designed to ensure all the extracted eventualities are semantically complete and all the patterns are exclusive with each other.
Take sentence ``I have a book'' as an example, we will only select $\langle$``I,'' ``have,'' ``book''$\rangle$ rather than $\langle$``I,'' ``have''$\rangle$ as the valid eventuality, because ``have''-\texttt{dobj}-``book'' is a negative dependency edge for pattern ``s-v.'' 


To extract eventualities from sentence $s$, considering that $s$ may contain multiple eventualities, we first split it into simple clauses based on the constituency tree. To do so, besides the commonly used \textit{SBAR} node, we also follow previous discourse parsing systems~\cite{DBLP:conf/conll/WangL15} to use a connective classifier to detect possible separators.
As a result, we split sentences based on both the subordinate conjunctions and connectives.
After that, for each verb $v$ in sentence $s$, we find the dependency graph $\DM$ of the simple clause that contains $v$.
We then try to match $\DM$ with all patterns one by one.
For each pattern, we put the verb $v$ as the starting point (i.e., $v_1$ in the pattern) and then try to find all the positive dependency edges. If we can find all the positive dependency edges around the center verb, these matched edges and words linked by these edges are considered as potential edges and words of a valid eventuality.
Next, other edges and words are added via optional dependency edges.
In the end, we will check if any negative dependency edge can be found in the dependency graph. If not, we will keep current edges and words as a valid eventuality. Otherwise, we will disqualify it. 
The pseudo-code of the eventuality extraction algorithm is in Algorithm~\ref{algorithm:eventuality-extraction}.
The time complexity of eventuality extraction is $\mathcal{O}(|\SM| \cdot \overline{|\DM|} \cdot \overline{|\VM^{(v)}|})$ where $|\SM|$ is the number of sentences, $\overline{|\DM|}$ is the average number of dependency edges in a dependency parse tree, and $\overline{|\VM^{(v)}|}$ is the average number of verbs in a sentence.

\subsection{Eventuality Relation Extraction}\label{sec:relation_extraction}

\begin{algorithm}[t]\caption{Eventuality Relation Extraction}\label{algorithm:relation-extraction}

	\textbf{INPUT:} Parsed constituency trees $\KM_1$ and $\KM_2$ from adjacent sentences.\\
    \textbf{OUTPUT:} Extracted relations $\RM$.
    \begin{algorithmic}[1]
    \State Initialize relation list $\RM$ as empty.
    \State Extract possible connectives $\CM$ by a connective extractor given $\KM_1$ and $\KM_2$.
    \For {each possible connective $c \in \CM$}
        \If {two arguments of $c$ in the same sentence}
            \State Extract $A_1$ and $A_2$ by a SS arguments extractor given $c$ and the sentence.
        \Else
            \State Extract $A_1$ by a PS argument1 extractor given $c$ and $\KM_1$.
            \State Extract $A_2$ by a PS argument2 extractor given $c$ and $\KM_2$.
        \EndIf
        \If {$A_1$ is not \textit{null} and $A_2$ is not \textit{null}}
            \State Classify the relation $y$ by a explicit relation classifier given $c$, $\KM_1$, and $\KM_2$
            \State Find eventualities $\EM_h$ that are extracted from $A_1$.
            \State Find eventualities $\EM_t$ that are extracted from $A_2$.
            \State Set weight $w$ as $1 / (|\EM_h| \cdot |\EM_t|)$.
            \For{each eventuality $E_h$ in $\EM_h$}
                \For{each eventuality $E_t$ in $\EM_t$}
                    \State Build relation instance $R = \langle E_h, y, E_t \rangle$ with a weight $w$.
                    \State Append $R$ in $\RM$.
                \EndFor
            \EndFor
        \EndIf
    \EndFor
    \State Return $\RM$.
    \end{algorithmic}
\end{algorithm}

We then introduce how to extract the relations between eventualities.
Specifically, we employ an end-to-end discourse parser to extract the discourse relations.
The discourse parser's job is to parse a piece of text into a set of discourse relations between two adjacent or non-adjacent discourse units.
Take the sentence ``I have a story book, but it is not interesting.'' as an example. Ideally, a good discourse parser extracts ``I have a story book'' as arg1, ``it is not interesting'' as arg2, ``but'' as the connective, and annotate the relation as ``Contrast.''
In our current pipeline, we use the state-of-the-art discourse parser~\cite{DBLP:conf/conll/WangL15}, which is pre-trained on the CoNLL 2015 Shared Task data (PDTB)~\cite{xue2015conll}. From CoNLL 2015 results,\footnote{\url{https://www.cs.brandeis.edu/~clp/conll15st/results.html}} we can find out that this discourse parser can achieve 90.00\% and 90.79\% F1 scores on the test data from PDTB and the blind test data from Wikinews respectively on the explicit relation classification, but performance drops to 42.72\% and 34.45\% on the implicit relation classification. Hence, to guarantee the extraction quality, we only consider the explicit discourse relations.
In explicit discourse parsing, there are two situations: both arguments are in the same sentence or not.
Statistics show that less than 0.1\% arguments are located in non-adjacent sentences in the explicit scenario, so we simply assume that the first argument is located in the same sentence (SS) or the previous sentence (PS).
Specifically, the explicit discourse parser is consist of five components: 
(1) connective extractor to identify whether a word is a possible connective,
(2) arg1 position classifier to decide whether the arg1 is located in the same sentence as the connective $c$ or the previous sentence of $c$;
(3) SS argument extractor to extract the spans of two arguments in the same sentence;
(4) PS argument extractor to extract the spans of two arguments in adjacent sentences;
(5) explicit relation classifier to classify the relation type of $c$.
Extractors in this system are essentially binary classifiers to identify whether a word is a connective or a part of any argument.
The pseudo-code of eventuality relation extraction algorithm is shown in Algorithm~\ref{algorithm:relation-extraction}.

As the extracted arguments might not be identical as the extracted eventualities, we use the Simpson's similarity to determine whether the discourse relations between arguments can be assigned to the extracted eventualities:
\begin{align}
    w_{A,E}^{(sim)} = \text{Simpson}(A, E) = \frac{| \WM_A \cup \WM_E |}{\text{min}\{|\WM_A|, |\WM_E|\}}, \label{eq:simpson}
\end{align}
where $A$ is an argument, $E$ is an eventuality, $\WM_A$ and $\WM_E$ are token sets of $A$ and $E$, $|\cdot|$ is the size of a token set.
If the similarity $\text{Simpson}(A, E) \geq 0.8$, we consider the argument-level relations relevant to $A$ can be assigned to the eventuality $E$ with a weight $w_{A,E}^{(sim)}$, which is inversely proportional to the size of all matched eventualities $|\EM|$.
It is worth noting that Eq.~(\ref{eq:simpson}) allows one argument $A$, which could include multiple eventualities as long as all tokens in eventualities can be covered by $A$.
In this situation, the weight of the relation between the eventuality $E_1 \in \EM_h$ from arg1 and the eventuality $E_2 \in \EM_t$ from arg2 is inversely proportional to the product of extracted eventuality sizes from two arguments $|\EM_h| \cdot |\EM_t|$.
Section~\ref{sec:aser_extract_example} provides detailed descriptions.

\subsection{Enriching ASER with Conceptualization}\label{sec:conceptualization}

We then introduce the conceptualization details. For each noun or pronoun in the extracted eventualities, we will try to conceptualized it to a higher level with the following steps. If it is a named entity, we will conceptualized it to the corresponding NER tags. Specifically, we include the 13 NER types: ``{\it Time},'' ``{\it Date},'' ``{\it Duration},'' ``{\it Money},'' ``{\it Percent},'' ``{\it Number},'' ``{\it Country},'' ``{\it State or Province},'' ``{\it City},'' ``{\it Nationality},'' ``{\it Person},'' ``{\it Religion},'' ``{\it URL}.'' 
If it is a personal pronoun (e.g., ``I,'' ``you,'' or ``they''),
we will conceptualize it to ``{\it PersonX}.''\footnote{If there are multiple people in the same edge, we will distinguish them with ``{\it PersonX}'' and ``{\it PersonY}'' etc.}
As all aforementioned conceptualization is designed by experts, we set the conceptualization probability to be 1.
If it is a regular noun, we will try to conceptualize it with Probase~\cite{wu2011taxonomy}.
Specifically, for each noun, we will retrieve its top-five hypernyms (i.e., concepts) and the associated probability from Probase.

Given an eventuality $E$ with $m$ tokens ${t_1, t_2, \cdots, t_m}$ to be mapped into concept tokens, we conceptualize it to a conceptualized eventuality $C$ with the probability:
\begin{equation}
    \text{Pr}(C|E) = \prod_{i=1}^m \text{Pr}(t_i^{(c)} | t_i^{(e)}).
\end{equation}
Here $t_i^{(c)}$ is the corresponding token-level concept for token $t_i^{(e)}$. And $\text{Pr}(t_i^{(c)} | t_i^{(e)})$ is the likelihood for $\langle t_i^{(e)}, \texttt{IsA}, t_i^{(c)} \rangle$ provided by Probase or 1.0 if $t_i^{(e)}$ can be conceptualized with rules.
For each conceptualized eventuality $C$, we would have a list of eventualities $\EM_{C}$ that can be conceptualized to it. We can then compute the overall weight of $C$ with Eq.~(\ref{eq:concept_weight}), where $w_{E}^{(e)}$ is the weight of $E$:
\begin{equation}
    w_{C}^{(c)} = \sum_{E \in \EM_{C}} \text{Pr}(C|E) \cdot w_{E}^{(e)}. \label{eq:concept_weight}
\end{equation}

\begin{figure}[!t]
\centering
\includegraphics[width=0.95\linewidth]{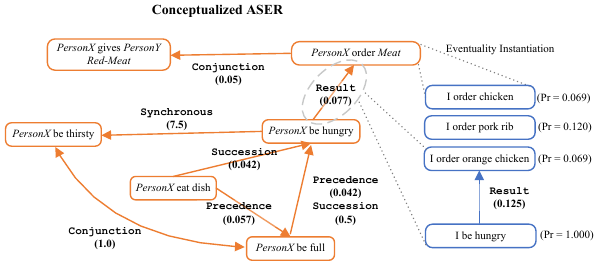}
\caption{Demonstration of the conceptualized ASER. The eventualities are conceptualized and connected with weighted edges. Each concept contains it's projections to specific eventualities.}
\label{fig:concept_demo}
\end{figure}

We then introduce how to construct the edges between a conceptualized eventuality $C$ and an original eventuality $E$.
For any $E^\prime \in \EM_{C}$, if there is an edge $\langle E^\prime, T, E \rangle$ or $\langle E, T, E^\prime \rangle$, we can then construct a new edge $\langle C, T, E \rangle$ or $\langle E, T, C \rangle$ with the weight based on Eq.~(\ref{eq:event_concept_edge1}) or Eq.~(\ref{eq:event_concept_edge2}), respectively, where $w_{R}^{(r)}$ means of weight of the relation $R$.

\begin{align}
    w_{\langle C, T, E \rangle}^{(r)} &= \sum_{E^\prime \in \EM_{C}} {\text{Pr}(C|E^\prime) \cdot w_{\langle E^\prime, T, E \rangle}^{(r)}}, \label{eq:event_concept_edge1} \\
    w_{\langle E, T, C \rangle}^{(r)} &= \sum_{E^\prime \in \EM_{C}} {w_{\langle E, T, E^\prime \rangle}^{(r)} \cdot \text{Pr}(C|E^\prime)}. \label{eq:event_concept_edge2}
\end{align}


As each conceptualized eventuality is correlated with a set of original eventualities, we need to aggregate the edges between the original eventualities to build the connections between the conceptualized ones.
Formally, given two conceptualized eventualities $C_h$ and $C_t$, we first retrieve all related original edges $\{\langle E_h, T, E_t \rangle | E_h \in \EM_{C_h}, E_t \in \EM_{C_t} \}$. Then we calculate the weight $\text{Pr}(C_h|E_h) \cdot w_{\langle E_h, T, E_t \rangle}^{(r)} \cdot \text{Pr}(C_t|E_t)$ for each related edge. Finally, we aggregate all weights to construct the weight as Eq.~(\ref{eq:concept_edge}) for the edge between $C_h$ and $C_t$ associated with the relation type $T$.

\begin{equation}
    w_{\langle C_h, T, C_t \rangle}^{(r)} = \sum_{E_h \in \EM_{C_h}} \sum_{E_t \in \EM_{C_t}} {\text{Pr}(C_h|E_h) \cdot w_{\langle E_h, T, E_t \rangle}^{(r)} \cdot \text{Pr}(C_t|E_t)}. \label{eq:concept_edge}
\end{equation}

An illustration of the conceptualized ASER is shown in Figure~\ref{fig:concept_demo}.
We can get the conceptualized eventuality ``\textit{PersonX} be hungry'' from ``I am hungry,'' ``they are hungry,'' and other extracted eventualities with $Pr(C|E)=1.000$ because their subjects (pronouns or names) are mapped to the token-level concept ``\textit{PersonX}.'' 
As a comparison, ``\textit{PersonX} order \textit{Meat}'' is not a deterministic eventuality: it can be conceptualized from ``I order chicken'' with $\text{Pr}(\textit{PersonX}\text{ order }\textit{Meat}|\text{I order chicken}) = \text{Pr}(\textit{PersonX}|\text{I}) \cdot \text{Pr}(\textit{Meat}|\text{chicken}) = 1.000 \cdot 0.069 = 0.069$, ``I order pork rib'' with $\text{Pr}(\textit{PersonX}\text{ order }\textit{Meat}|\text{I order pork rib}) = \text{Pr}(\textit{PersonX}|\text{I}) \cdot \text{Pr}(\textit{Meat}|\text{pork rib}) = 1.000 \cdot 0.120 = 0.120$, or other extracted ones. 
Based on Eq.~(\ref{eq:concept_weight}), after aggregating all weights together, we can get the concept weights for ``\textit{PersonX} be hungry'' and ``\textit{PersonX} order \textit{Meat}'' are 1389.000 and 27.705, respectively.
As for the relations between the two conceptualized eventualities,
we find $w^{(r)}_{\langle \text{I am hungry}, \texttt{Result}, \text{I order orange chicken} \rangle} = 0.125$ and $w^{(r)}_{\langle \text{I am too hungry}, \texttt{Result}, \text{I order the fried chicken} \rangle} = 1.000$, so the relation weight in the concept-level is calculated as follows:
\begin{align*}
    & w^{(r)}_{\langle \textit{PersonX} \text{ be hungry}, \texttt{Result}, \textit{PersonX}\text{ order }\textit{Meat} \rangle} \\
    &= \text{Pr}(\textit{PersonX} \text{ be hungry} | \text{I am hungry}) \cdot 0.125 \cdot \text{Pr}(\textit{PersonX} \text{ order } \textit{Meat} | \text{I order orange chicken}) \\
    & \quad + \text{Pr}(\textit{PersonX} \text{ be hungry} | \text{I am too hungry}) \cdot 1.000 \cdot \text{Pr}(\textit{PersonX} \text{ order } \textit{Meat} | \text{I order the fried chicken}) \\
    &= 1.000 \cdot 0.125 \cdot 0.069 + 1.000 \cdot 1.000 \cdot 0.069 \\
    &= 0.077.
\end{align*}
Similarly, we can calculate all weights among conceptualized eventualities associated with different relation types.
One thing worth mentioning is that the relation weights depend not only on the relation weights in the extracted knowledge bases but also on the conceptualization probabilities.




\subsection{ASER Building Example}
\label{sec:aser_extract_example}

\begin{figure}[!t]
\centering
\includegraphics[width=0.98\linewidth]{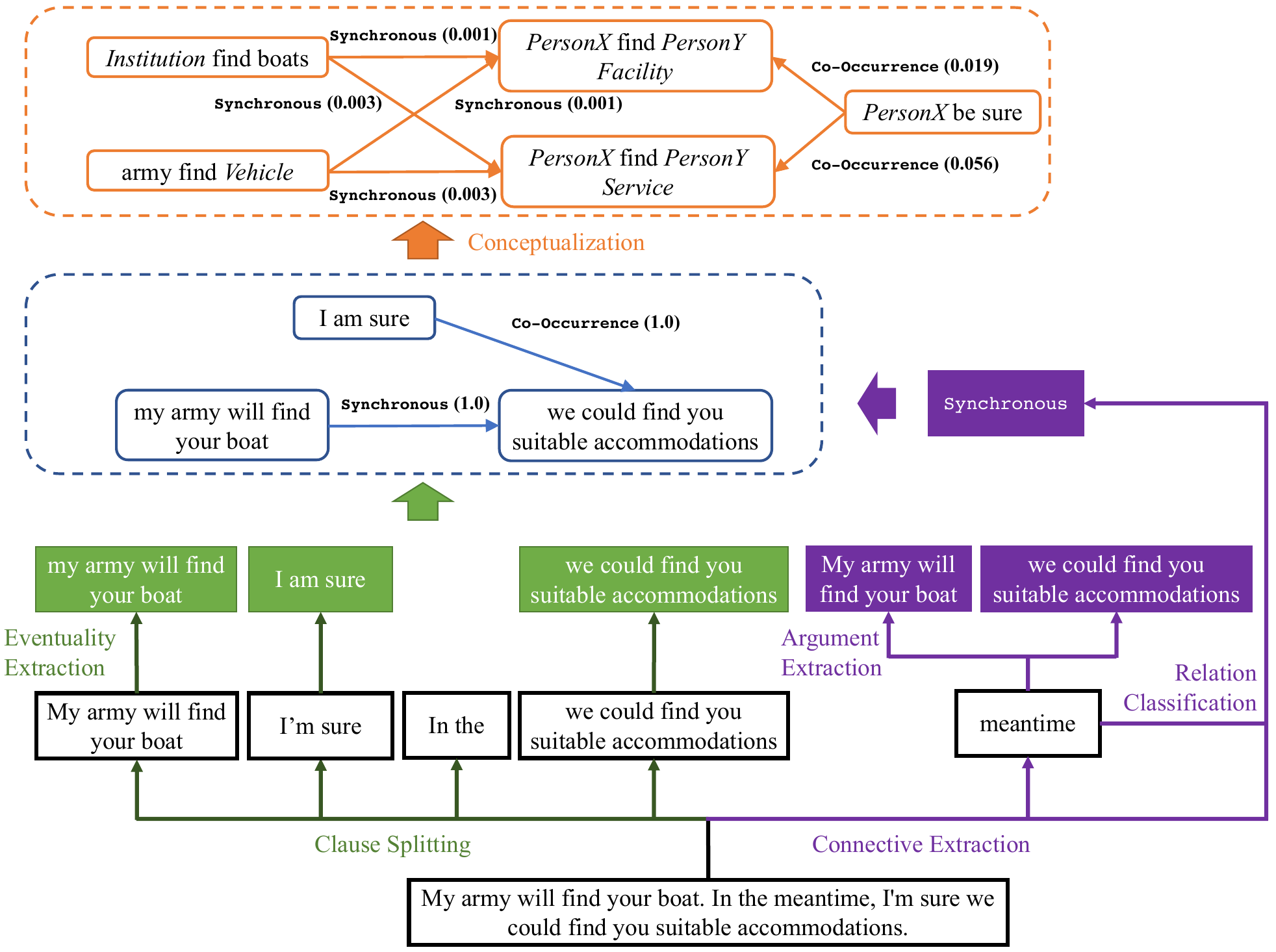}
\caption{ASER building example. The eventuality extraction, the relation extraction, and the conceptualization process are shown in green, violet, and orange colors, respectively.
For the clear representation, for each conceptualized eventuality, we only show the skeleton words and hide the optional ones.
We only show the two conceptualized eventualities with the highest probabilities for each extracted eventuality. As a demonstration, the weights of eventualities and relations are calculated from this example instead of the whole KG.
}
\label{fig:aser_building_example}
\end{figure}

At the end of this section, we use an example to demonstrate the whole extraction pipeline.
As shown in Figure~\ref{fig:aser_building_example}, given a text ``My army will find your boat. In the meantime, I'm sure we could find you suitable accommodations.,''\footnote{This case comes from Movie Subtitles.} our system will first detect the possible connective ``meantime'' and split this text into four simple clauses: ``My army will find your boat,'' ``In the,'' ``I'm sure,'' and ``we could find you suitable accommodations'' with the constituency parsing.
After that, our system will leverage the patterns designed in Table~\ref{tab:eventuality-pattern} to extract eventualities from the raw text by Algorithm~\ref{algorithm:eventuality-extraction}.
Simultaneously, two arguments ``My army will find your boat'' and ``we could find you suitable accommodations'' are extracted by argument extractors.
The discourse parsing system predicts the corresponding discourse relation as \texttt{Synchronous}.
As the first and last extracted eventualities can perfectly match the extracted arguments, we then create an edge $\langle$ ``my army will find your boat,'' \texttt{Synchronous}, ``we could find you suitable accommodations'' $\rangle$.
We also create an edge $\langle$ ``I am sure,'' \texttt{Co-Occurrence}, ``we could find you suitable accommodations'' $\rangle$ because the two eventualities appear in the same sentence.
After extracting the original eventualities and edges, we then try to expand it with the conceptualization.\footnote{In the real system, we first extract the original ASER, and then apply the conceptualization step over the whole KG. The presented single sentence example is just for the demonstration.} 
For example, ``I am sure'' can be directly conceptualized as ``\textit{PersonX} be sure'' directly because ``I'' is a personal pronoun.
As both of the other two eventualities contain regular nouns (i.e., ``army''), these eventualities can be conceptualized to multiple eventualities.
After checking Probase, we find out that ``army'' can be conceptualized to ``\textit{Institution}'' and ``\textit{Organization}'' with the weights 0.058 and 0.038, ``boat'' can be conceptualized to ``\textit{Vehicle}'' and ``\textit{Item}'' with the weights 0.059 and 0.049, ``accommodation'' can be conceptualized to ``\textit{Service}'' and ``\textit{Facility}'' with the weights 0.056 and 0.019, respectively.
We show the two most likely results for each original eventuality (if it has multiple possible conceptualization results) in Figure~\ref{fig:aser_building_example}.
In the end, we can construct edges between conceptualized eventualities, where the weights are the product of conceptualization probabilities, e.g., $\langle$ ``\textit{Institution} find boats,'' \texttt{Synchronous}, ``\textit{PersonX} find \textit{PersonY} \textit{Service}'' $\rangle$ with the weight $0.058 \times 0.056 = 0.003$,
$\langle$ ``\textit{PersonX} be sure,'' \texttt{Co-Occurrence}, ``\textit{PersonX} find \textit{PersonY} \textit{Facility}'' $\rangle$ with the weight $1.000 \times 0.019 = 0.019$.




\begin{table}[!ht]
    \small
	\centering
	{
		\begin{tabular}{c|c|c|c|c|c}
			\toprule 
			\multirow{2}{*}{Pattern} & \multicolumn{2}{c|}{ASER (full)} & \multicolumn{2}{c|}{ASER (core)} & {ASER (concept)} \\
			& \# Eventuality & \# Unique & \# Eventuality & \# Unique & \# Unique \\
			\midrule
	        s-v & 351,082,855 & 100,645,728 & 260,663,083 & 14,337,769 & 1,022,415 \\
            s-v-o & 284,103,317 & 159,948,356 & 139,031,585 & 18,100,360 & 8,252,653 \\
            s-v-a & 11,546,768 & 6,149,584 & 5,951,980 & 752,468 & 139,087 \\
            s-v-v & 24,549,946 & 11,129,566 & 14,624,526 & 1,591,424 & 216,413 \\
            s-v-o-o & 6,154,685 & 3,789,253 & 2,765,728 & 460,526 & 514,084 \\
            s-v-v-o & 29,445,708 & 18,659,717 & 12,720,497 & 2,187,577 & 1,482,783 \\
            s-v-o-v-o & 3,863,478 & 2,674,229 & 1,462,883 & 288,326 & 522,613 \\
            s-v-o-v-o-o & 91,532 & 59,290 & 40,428 & 8,499 & 18,461 \\
            s-be-a & 79,235,136 & 29,845,112 & 52,068,570 & 3,733,978 & 465,747 \\
            s-be-o & 98,411,474 & 53,503,410 & 49,979,659 & 6,337,042 & 2,312,209 \\
            s-v-be-a & 1,927,990 & 982,438 & 1,035,864 & 123,263 & 29,738 \\
            s-v-be-o & 2,322,890 & 1,574,896 & 909,250 & 184,298 & 139,239 \\
            s-v-o-be-a & 277,087 & 191,973 & 100,917 & 18,793 & 6,151 \\
            s-v-o-be-o & 307,031 & 231,289 & 95,815 & 22,411 & 32,796 \\
            there-be-o & 16,021,849 & 6,642,438 & 10,013,628 & 953,041 & 39,500 \\
            spass-v & 61,524,872 & 38,270,144 & 25,935,769 & 3,498,516 & 276,817 \\
            spass-v-o & 5,519,982 & 4,129,709 & 1,677,244 & 330,229 & 154,410 \\
            spass-v-v-o & 257,004 & 221,820 & 46,475 & 11,738 & 14,901 \\
            \midrule
            Overall & 976,643,604 & 438,648,952 & 579,123,901 & 52,940,258  & 15,640,017 \\
			\bottomrule
		\end{tabular}
	}
	\caption{Statistics of the eventuality extraction. \# Eventuality and \# Unique mean the total number and the unique number of extracted eventualities using corresponding patterns or conceptualized eventualities from them. }\label{tab:eventuality-statistics}
\vspace{-0.1in}
\end{table}

\begin{figure}[!ht]
    \centering
    \subfigure[Extracted eventualities]{\label{fig:dist_eventualities}
		\includegraphics[width=0.43\linewidth]{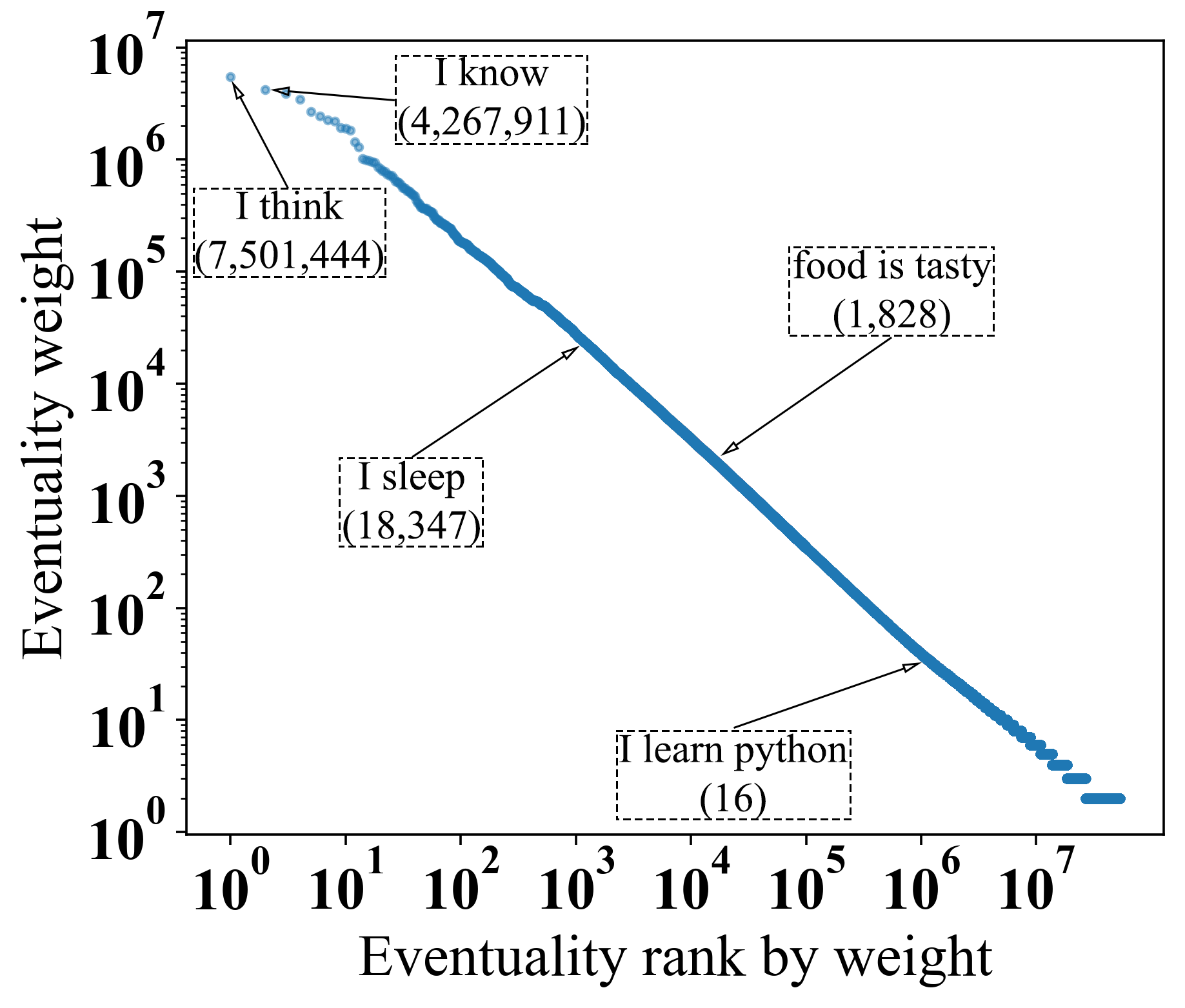}
	}
	\subfigure[Conceptualized eventualities]{\label{fig:dist_concepts}
		\includegraphics[width=0.43\linewidth]{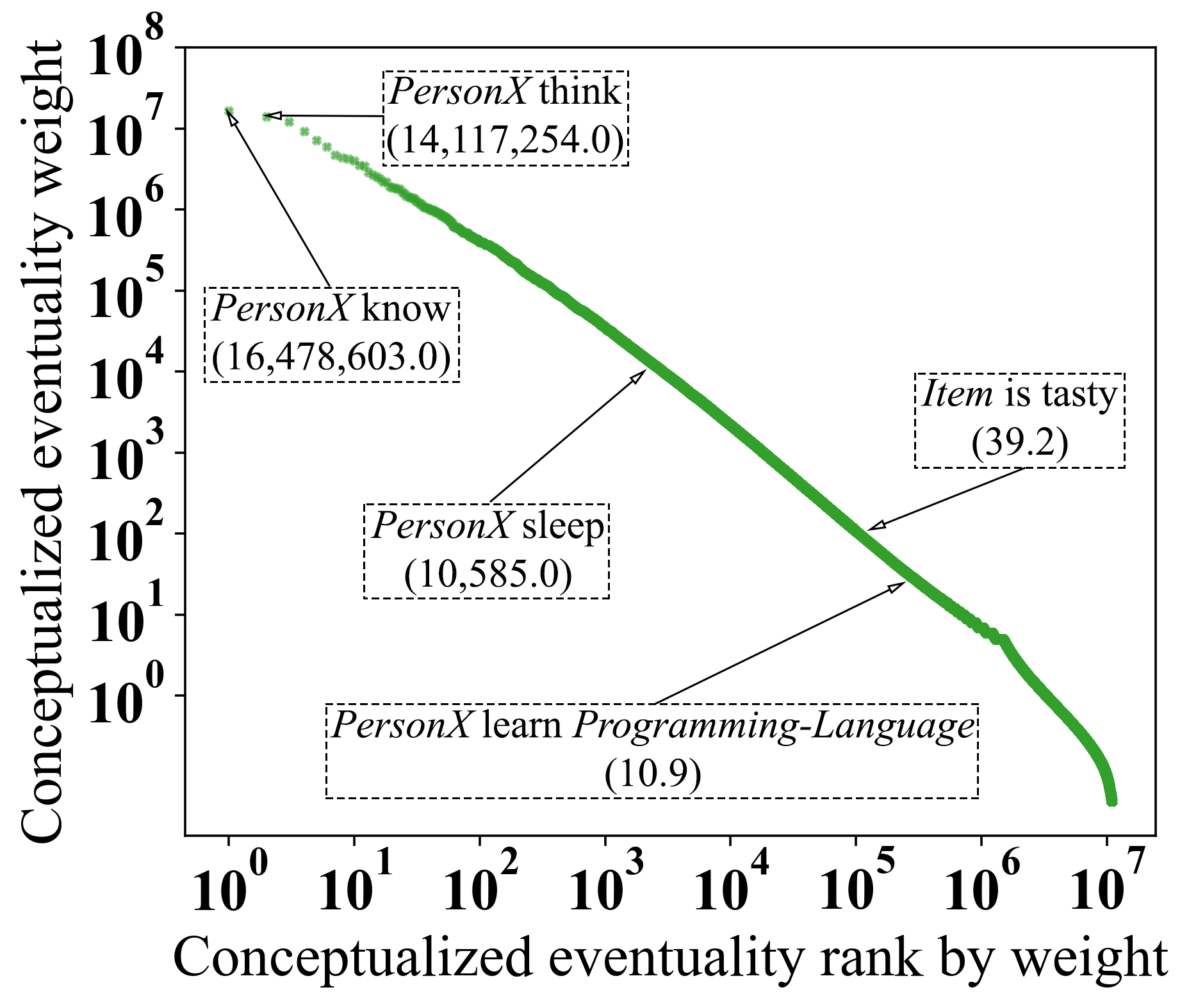}
	}
	\caption{Eventuality distributions.
	}
\end{figure}

\begin{table}[!t]
    \small
	\centering
		
	{
		\begin{tabular}{c|c|c|c}
			\toprule 
			{Relation} & {ASER (full)} & {ASER (core)} & {ASER (concept)}\\
			\midrule
	        Precedence & 14,058,213 & 1,790,016 & 4,798,015 \\ 
            Succession & 4,939,291 & 663,183 & 1,963,820 \\
            Synchronous & 19,464,898 & 3,123,042 & 8,013,943 \\
            Reason & 9,775,829 & 2,205,076 & 6,439,128 \\
            Result & 16,153,925 & 2,012,311 & 6,718,666 \\
            Condition & 18,052,484 & 3,160,271 & 8,063,967 \\
            Contrast & 59,333,901 & 8,655,661 & 24,978,311 \\
            Concession & 5,684,395 & 477,155 & 1,499,276 \\
            Conjunction & 82,121,343 & 13,978,907 & 45,597,200 \\
            Instantiation & 1,278,381 & 18,496 & 93,266 \\
            Restatement & 1,304,095 & 65,753 & 242,301 \\
            Alternative & 3,539,892 & 583,174 & 123,883 \\
            ChosenAlternative & 647,228 & 35,406 & 1,843,140 \\
            Exception & 106,000 & 20,155 & 93,412 \\ 
            Co-Occurrence & 412,054,590 & 49,232,161 & 113,744,814 \\
            \midrule
            Overall & 648,514,465 & 86,020,767 & 224,213,142 \\
			\bottomrule
		\end{tabular}
	}
	\caption{Statistics of the eventuality relation extraction. }\label{tab:relation-statistics}
\end{table}

\section{ASER Statistics}\label{sec:statistics}

In total, we collect 976,643,604 eventualities from the raw documents. We filter those low-frequency eventualities that only appear once and retain 52,940,258 unique eventualities in ASER (core). From Table~\ref{tab:eventuality-statistics}, we can find the ``s-v'' and ``s-v-o'' are the most frequent patterns.
On the other hand, even though those complex patterns appear relatively less frequently, thanks to the large scale of ASER, they still appear thousands to millions of times. 

The original eventuality distribution is presented in Figure~\ref{fig:dist_eventualities}.
In general, the distribution follows Zipf's law, where only a small number of eventualities appear many times while the majority of eventualities appear only a few times.
To better illustrate the distribution of eventualities, we also show several representative eventualities along with their weights, and we have two observations.
First, eventualities which can be used in general cases, like ``I think (7,501,444)'' and 
``I know'' (4,267,911) appear much more times than other eventualities.
Second, eventualities in ASER are more closely related to our daily life like ``I sleep (18,347)'' or ``food is tasty (1,828)'' rather than domain-specific ones such as ``I learn python (16).''

To achieve the balance between the quality and quantity of conceptualization results, we apply the conceptualization over eventualities whose frequencies are no less than five.
As shown in Table~\ref{tab:eventuality-statistics}, after the conceptualization, we get 15,640,017 more unique eventualities. It is obvious that patterns with more nouns (e.g., ``s-v-o,'' ``s-v-v-o,'' ``s-be-o'') dominate the conceptualized eventualities. The reason is that the conceptualization is only designed for nouns, and each noun phrase would be replaced with a general noun phase if such hypernym relation appears in Probase. 
For conceptualized eventualities, we can observe a similar distribution in Figure~\ref{fig:dist_concepts}. 
The top three conceptualized eventualities are ``{\it PersonX} know'' (16,478,603.0), ``{\it PersonX} think'' (14,117,254.0), and ``{\it PersonX} say'' (12,113,913.0).
Although ``I think'' (7,501,444) appears the most in the raw data (``I know'' appears 3,447,429 times in the raw data), but 
``you think'' (1,444,333), 
``he thinks'' (314,806), 
``they thinks'' (205,432), 
``we think'' (196,633), 
``it thinks" (174,729), 
``she thinks" (142,628), 
etc. appear much less than 
``you know'' (4,726,264), 
``he knows'' (396,013), 
``they know'' (260,656), 
``we know'' (457,115), 
``it knows'' (247,409), 
``she knows'' (190,803), 
etc., respectively.
Finally, the weight of ``{\it PersonX} know'' exceeds that of ``{\it PersonX} think.''

For relations, as shown in Table~\ref{tab:relation-statistics}, we collect 648,514,465 unique relations from six data resources across different categories.
To reduce noises in parsing and extraction, we also filter out relations that $\sum_{T^\prime \in \TM}{w_{\langle E_h, T^\prime, E_t \rangle}^{(r)}} <= 1$ where $E_h$ and $E_t$ are the head eventuality and the tail eventuality.
Furthermore, if the head or the tail is filtered out by eventuality filtering, the relation is also dropped.
Finally, we keep 86,020,767 unique relations in ASER (core), among which there are 36,788,606 relations belonging to 14 discourse relation types depending on the connectives and arguments, like \textit{Conjunction} (e.g., ``and''), \textit{Contrast} (e.g., ``but''), \textit{Condition}  (e.g., ``if''), \textit{Synchronous}  (e.g., ``meanwhile''), \textit{Reason} (e.g., ``because''), \textit{Result} (e.g., ``so'').
When we filter out more low-frequency eventualities, the number of relations decreases slightly.
For example, when we keep high-frequency eventualities whose frequencies are no less than five, 26.0\% of eventualities (13,766,746) and 61.5\% of relations (88,629,385) are preserved.
We apply the conceptualization over these preserved eventualities and relations based on quantity and quality considerations.
Finally, we obtain 15,640,017 unique conceptualized eventualities and 224,213,142 relations between these conceptualized eventualities.
In total, we have about 26 times more relations between conceptualized eventualities than original eventualities.

\begin{figure}[t]
    \centering
    \subfigure[Eventualities]{\label{fig:dist_nodes}
		\includegraphics[width=0.4\linewidth]{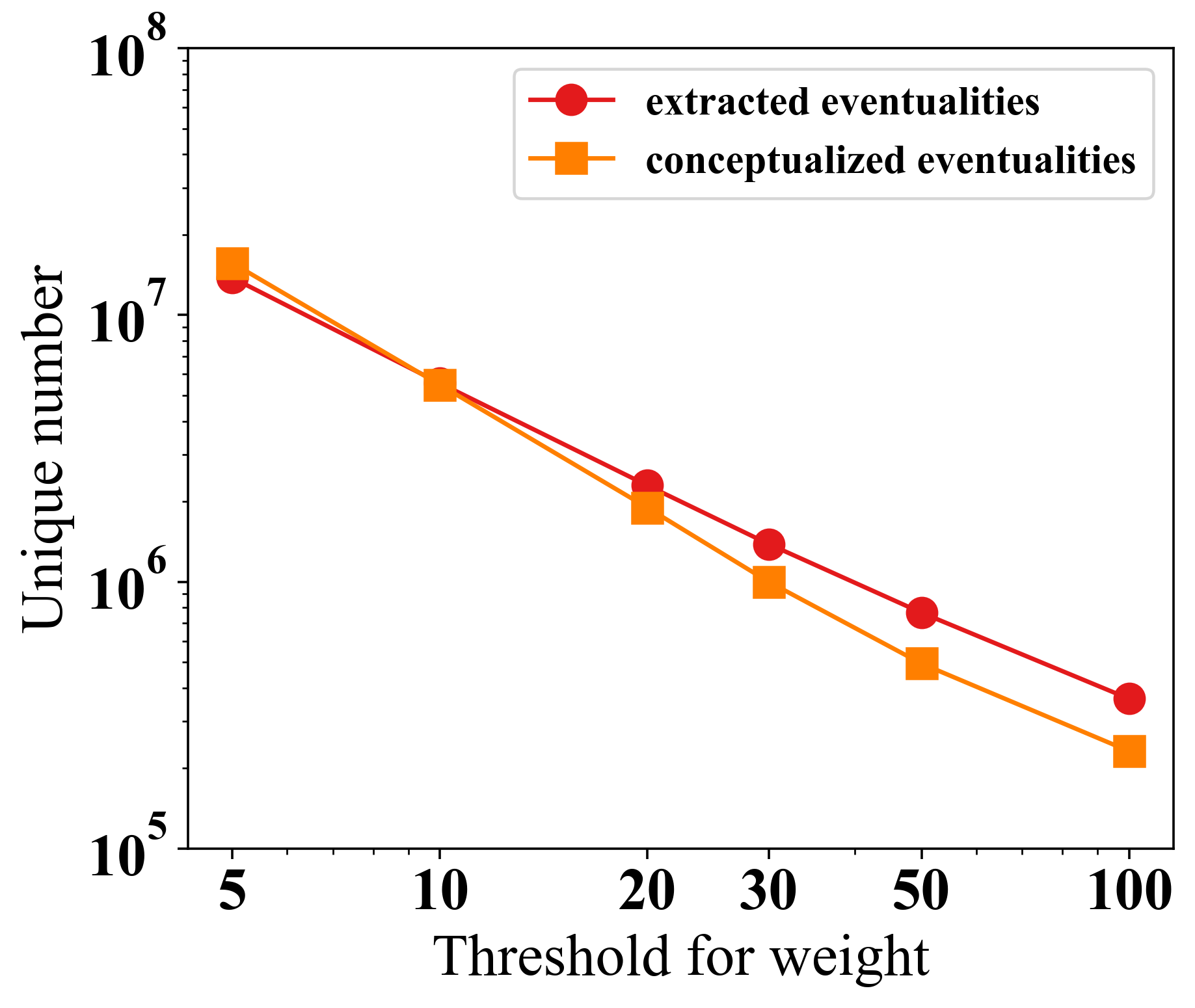}
	}
	\subfigure[Relations]{\label{fig:dist_links}
		\includegraphics[width=0.4\linewidth]{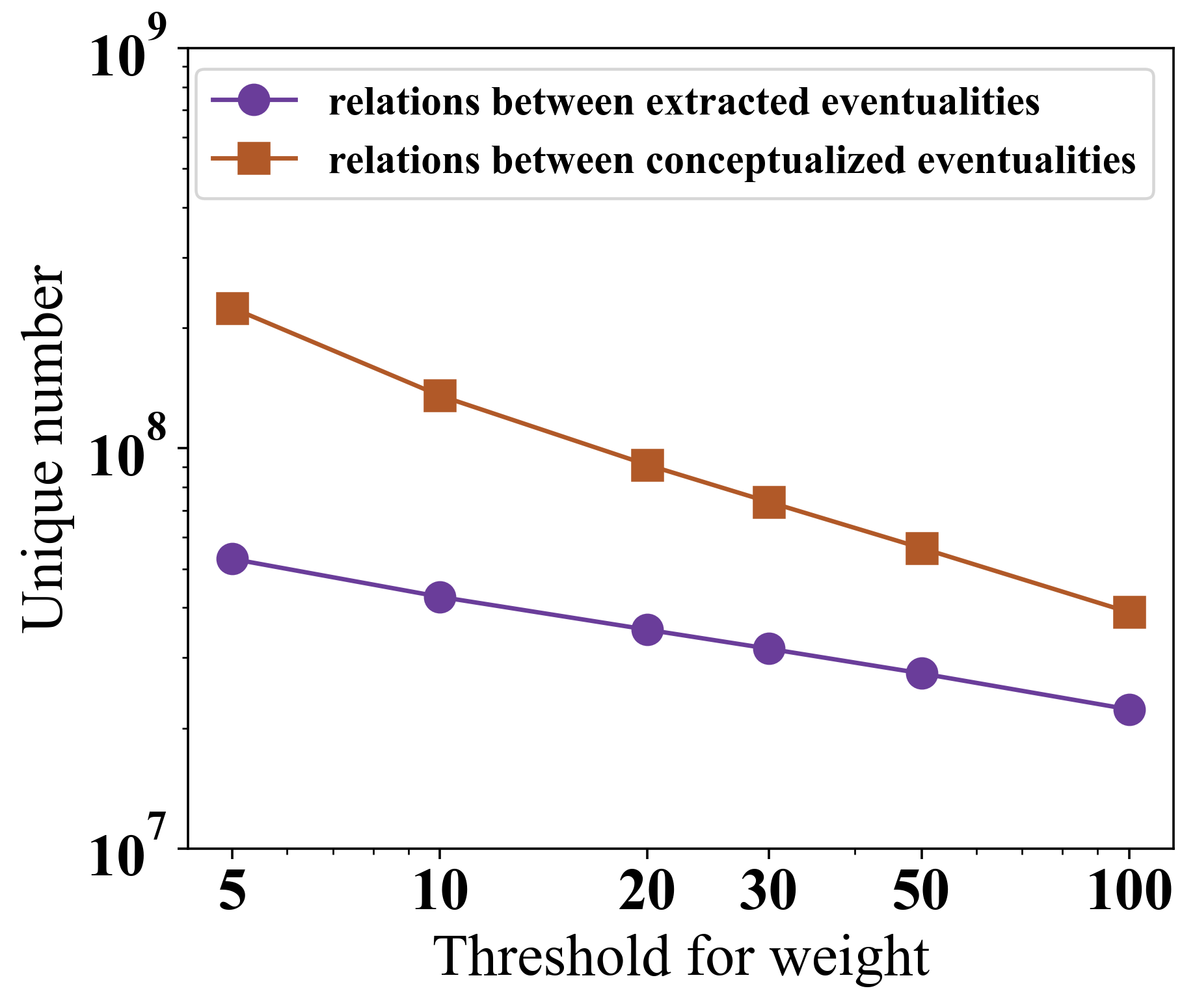}
	}
	\caption{Distributions of eventualities and relations.}
\end{figure}

To better understand the distributions of extracted and conceptualized knowledge, we show the number of eventualities and edges over different filtering thresholds in Figure~\ref{fig:dist_nodes} and Figure~\ref{fig:dist_links}, respectively.
For the extracted knowledge, the number of eventualities and relations decreases exponentially when the threshold ranges from 5 to 100.
For the conceptualized knowledge, the rate of diminishing is even larger.
When the threshold is less than 10,  the conceptualized eventuality size is greater than the original size.
But it is significantly less than the size of extracted eventualities as the threshold is larger than 10.
On the other hand, the number of conceptualized eventuality relations consistently exceeds the original relation size, which results in a denser conceptualized knowledge graph.


\section{Intrinsic Evaluation}\label{sec:intrinsic-evaluation}

In this section, we leverage human annotation to evaluate the quality of ASER from the following perspectives:

\begin{enumerate}
    \item \textbf{Eventuality Extraction}: We first evaluate how well the extracted eventualities can represent the original sentence's semantics. For example, if the original sentence is ``The kid goes to study,'' eventuality ``kid-go-to-study'' with pattern ``s-v-v'' can fully represent the semantics, but eventuality ``kid-go'' with the pattern ``s-v'' cannot. We show the percentage of all extracted eventualities that can fully represent the original sentences' core semantic based on different eventuality patterns.
    \item \textbf{Lower-order Selectional Preference}: Besides the extraction quality, we also care about how well the eventuality statistics in ASER can reflect human's selectional preference. For example, the frequency of ``I eat food'' should be higher than ``I eat house.'' As such preference appears inside eventualities, we denote them as the lower-order selectional preference. 
    \item \textbf{Discourse Extraction}: After evaluating the eventualities, we assess how well the extracted edges can correctly represent the discourse relations in the original sentence. For example, assume that the original sentence is ``he went to school while I was still preparing the breakfast.'' and we have successfully extracted two eventualities ``he went to school'' and ``I was preparing breakfast,'' the correct discourse relation between them should be ``Synchronous'' rather than ``Contrast.'' In this evaluation, we report the accuracy based on different discourse relations.
    \item \textbf{Higher-order Selectional Preference}: Last but not least, we annotate whether edge frequencies in ASER can reflect the higher-order selectional preference among eventualities or not. For example, the frequency of ``I am hungry''-\texttt{Result}-``I eat food'' should be larger than ``I am hungry''-\texttt{Reason}-``I eat food.''
\end{enumerate}

Evaluation details and result analysis are as follows.

\subsection{Eventuality Extraction}

\begin{figure}
    \centering
    \includegraphics[width=0.6\linewidth]{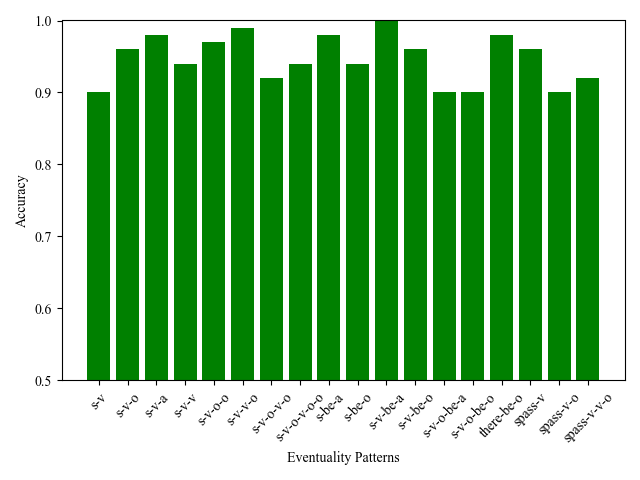}
    \caption{Human annotation of eventuality extraction quality.}
    \label{fig:eventuality_extraction}
\end{figure}

To evaluate the correctness of the selected eventuality patterns and the effectiveness of the extraction algorithm, we first employ the Amazon Mechanical Turk platform (MTurk)\footnote{\url{https://www.mturk.com/}} to evaluate the quality of eventuality extraction.
We randomly select 50 extracted eventualities for each eventuality pattern and then provide these extracted eventualities along with their original sentences to the annotators.
For each pair of eventuality and sentences, the annotators are asked to label whether the extracted eventuality phrase can fully and precisely represent the original sentences' semantic meaning. 
If so, they should label them with ``Valid.'' Otherwise, they should label it with ``Not Valid.''
For each eventuality, we invite six workers to label, and if at least four of them label it as ``Valid,'' we will consider it valid.
In total, we collected 5,400 annotations. To ensure high annotation quality, we require all the annotators to be the master annotator on the MTurk.

The annotation results are shown in Figure~\ref{fig:eventuality_extraction}.
From the result, we can see that all patterns achieve over 90\% accuracy, which demonstrates the high quality of the selected patterns and the association extraction algorithm.
As introduced in Algorithm~\ref{algorithm:eventuality-extraction}, to guarantee the quality of extracted eventualities, we require the extraction algorithm to be strict and selective.
Specifically, if there is an extra dependency edge not in the positive or possible relations of a corresponding pattern, we will discard the whole sentence.
By doing so, even though we sacrifice the overall recall, we guarantee high accuracy.
Luckily, as our approach is unsupervised, we can remedy the recall problem with a larger-scale corpus.
Among the 18 patterns, we notice that the more complex patterns tend to have relatively lower accuracy.
This makes sense because the more complex an extracted eventuality is, the more likely that some of the words are redundant to the eventuality semantics. Thus the annotator may think that the extracted eventuality is not elegant enough.

\subsection{Low-order Selectional Preference}

\begin{figure*}[tb]
    \centering
	\subfigure[High Frequency vs. Low Frequency  (before).]{\label{fig:low_sp_frequency}
		\includegraphics[width=0.47\linewidth]{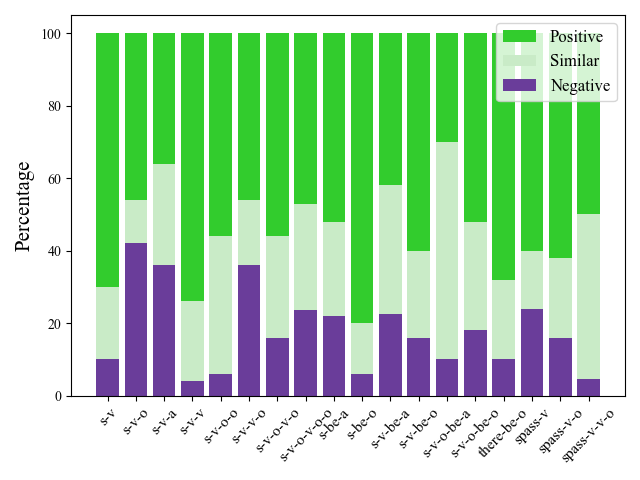}
	}        		
	\subfigure[Exist vs. Non-exist (before).]{\label{fig:low_sp_exist}
		\includegraphics[width=0.47\linewidth]{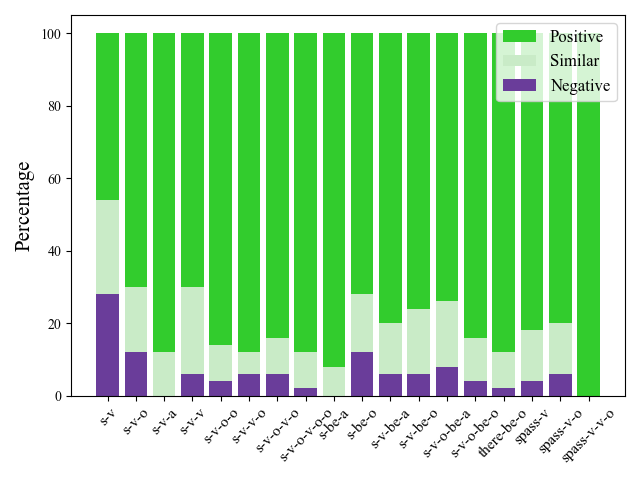}
	}
	\subfigure[High Frequency vs. Low Frequency (after). ]{\label{fig:low_sp_concept_frequency}
		\includegraphics[width=0.47\linewidth]{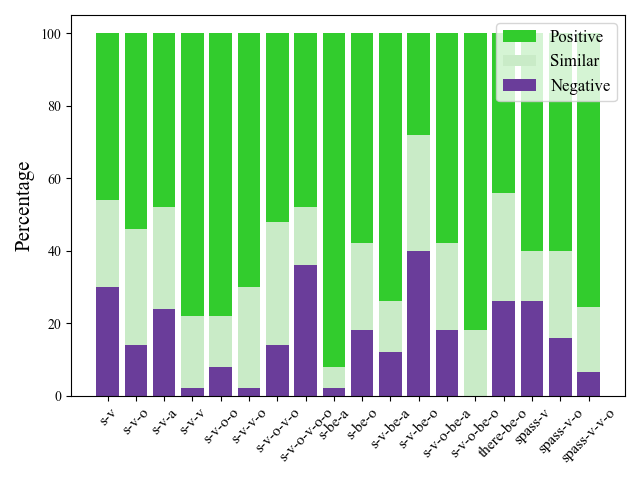}
	}        		
	\subfigure[Exist vs. Non-exist (after).]{\label{fig:low_sp_concept_exist}
		\includegraphics[width=0.47\linewidth]{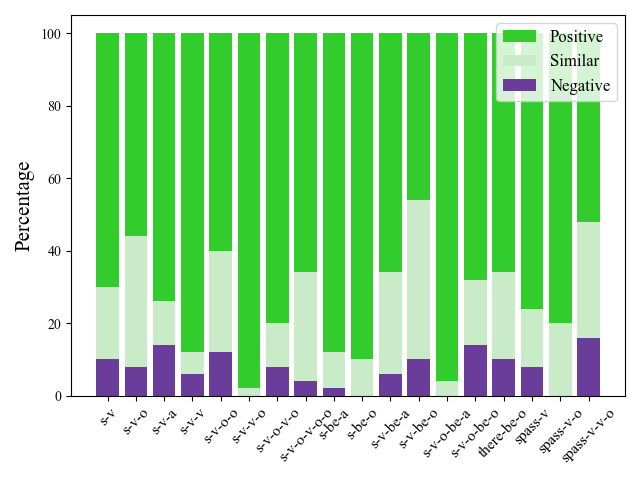}
	}   
	\caption{Human annotation of lower selectional preference in ASER. Experiments on the eventualities before and after the conceptualization are denoted with (before) and (after), respectively. The green color indicates the number of eventuality pairs that the more frequent eventuality makes more sense. The purple color indicates the number of eventualities pairs the less frequent eventuality makes more sense. } 
	\label{fig:low_sp}
\end{figure*}

To evaluate whether the eventuality frequencies in ASER can reflect human's low-order selectional preference, we first compare the plausibility of more frequent eventualities versus less frequent ones.
For each eventuality pattern, we randomly select 50 eventuality pairs such that they only have a one-word difference but with a significant frequency difference. Specifically, we require the frequency of the high-frequency one to be larger than five, which is the medium frequency of all eventualities, and the frequency of the high-frequency one must be at least five times larger than the frequency of the low-frequency one.
For each eventuality, we invite six annotators from MTruk to ask them which one of the eventualities seems more plausible to them. If more annotators agree that the more frequent one makes more sense, we will label that pair as a positive correlation. On the other hand, if more annotators agree that the less frequent makes more sense, we will label that pair as a negative correlation. If the voting draws, we will label it as similar.
As this evaluation fails to consider the eventualities with the frequency zero (i.e., they do not exist in ASER) and whether an eventuality exists or not is also a good preference indicator, we add another evaluation to prove that.
For each eventuality pattern, we randomly select 50 eventualities. Then for each of the eventualities, we randomly select a negative example by randomly changing a word inside the eventuality with another word of the same POS tag label such that their grammar structure is the same.
We also conduct filtering to guarantee the negative examples do not appear in ASER.
Last but not least, to show the influence of the conceptualization, we conduct the aforementioned two experiments on both the original ASER before the conceptualization and the final one after the conceptualization.\footnote{For the experiment on the ASER after the conceptualization, we only sample the conceptualized eventualities and ignore the original ones.}

We present the annotation results in Figure~\ref{fig:low_sp}\Blue{.}
The green color indicates the number of eventuality pairs that the more frequent eventuality makes more sense, and the purple color indicates the number of eventualities pairs the less frequent eventuality makes more sense.
From the result in Figure~\ref{fig:low_sp} (a), we can see that more than 70\% of the eventuality pairs as positively correlated, which is consistent with the previous study on the correlation between frequency and selectional preference~\cite{zhang2019sp-10k}.
At the same time, we also observe that about 30\% of the less frequent eventualities are also quite plausible, which is mainly because the frequency of an eventuality is also severely influenced by the rareness of the words inside the eventuality. For example, the eventuality ``I eat avocado'' appears much less than ``I eat apple'' because avocado is much rarer than apple rather than ``I eat apple'' makes more sense than ``eat avocado.'' 
The results in Figure~\ref{fig:low_sp} (b) help prove that the low-frequent eventualities still contain rich low-order selectional preference because, for more than 90\% of the pairs, the randomly extracted pairs in ASER makes more sense than those out of ASER.
Furthermore, the experimental results in Figure~\ref{fig:low_sp} (c) and (d) show that even though the conceptualization process significantly improves the coverage of ASER, it would not hurt the overall quality. This is mainly because, during the conceptualization step, we carefully design the new weights based on the original weight and the confidence scores provided by Probase~\cite{wu2011taxonomy}.

\subsection{Relation Extraction}

\begin{figure}
    \centering
    \includegraphics[width=0.6\linewidth]{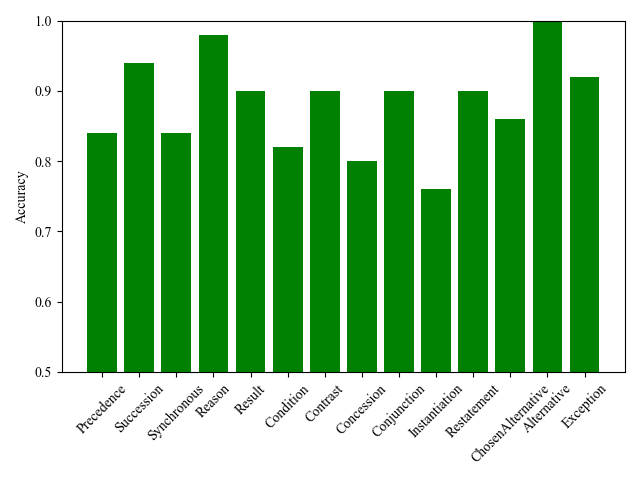}
    \caption{Human annotation of discourse relation extraction quality.}
    \label{fig:edge_extraction}
\end{figure}

Besides the eventuality extraction, we also care about the extraction quality of the discourse relations between eventualities.
For each relation type, we randomly select 50 edges and the corresponding sentences. We generate a question for each pair of them by asking the annotators if they think the extracted discourse relation can represent the correct relation in the original sentence.
If so, they should label it as ``Valid.'' Otherwise, they should label it as ``Not Valid.''
Similar to the eventuality extraction experiment, we invite six annotators for each edge. If more than four of them agree that the extracted relation is ``Valid,'' we will consider it to be ``Valid.''

From the results in Figure~\ref{fig:edge_extraction}\Blue{,} we can see that the overall accuracy is about 80\%, which is consistent with the reported performance of the used discourse relations extraction system~\cite{DBLP:conf/conll/WangL15}. Besides that, we also notice that the model performance varies on different relation types.
For example, the model tends to perform well on simple types such as ``Reason'' and ``Alternative'' because the popular connectives (i.e., ``because'' and ``or'') are less ambiguous.
As a comparison, when the connective is more ambiguous (e.g., ``while'' for ``Synchronous''), the overall performance will drop.

\subsection{Higher-order Selectional Preference}

\begin{figure*}[tb]
    \centering
	\subfigure[High Frequency vs. Low Frequency (before).]{\label{fig:high_sp_frequency}
		\includegraphics[width=0.47\linewidth]{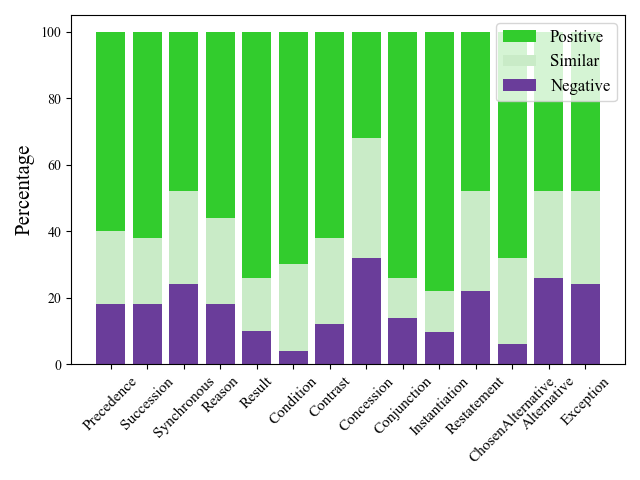}
	}        		
	\subfigure[Exist vs. Non-exist (before).]{\label{fig:high_sp_exist}
		\includegraphics[width=0.47\linewidth]{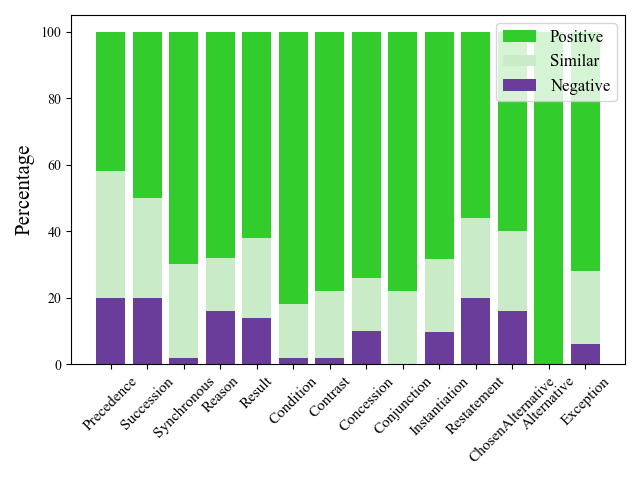}
	}
	\subfigure[High Frequency vs. Low Frequency (after).]{\label{fig:high_sp_concept_frequency}
		\includegraphics[width=0.47\linewidth]{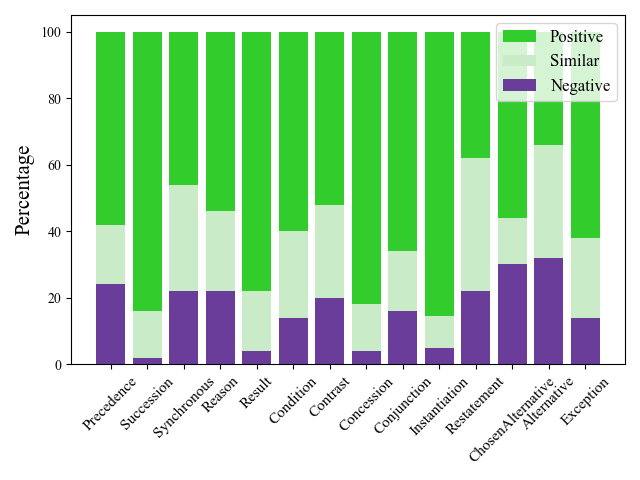}
	}        		
	\subfigure[Exist vs. Non-exist (after).]{\label{fig:high_sp_concept_exist}
		\includegraphics[width=0.47\linewidth]{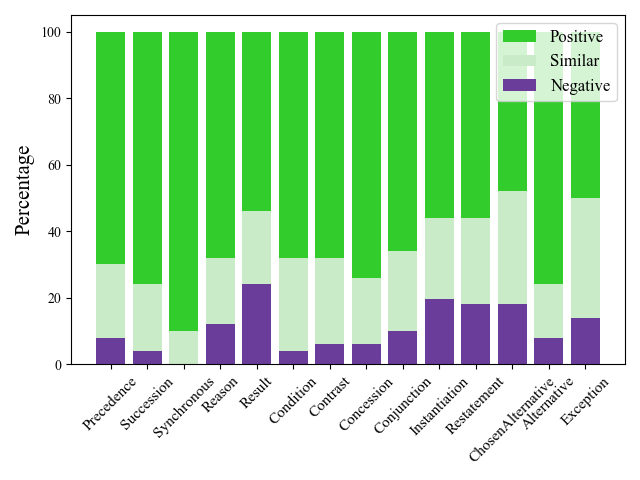}
	}   
	\caption{Human annotation of the higher selectional preference in ASER. Experiments on the eventualities before and after the conceptualization are denoted with (before) and (after), respectively. The green color indicates the number of edge pairs that the more frequent edge makes more sense, and the purple color indicates the number of edge pairs the less frequent edge makes more sense. } 
	\label{fig:high_sp}
\end{figure*}

Finally, we evaluate whether the edge frequency in ASER can be used to reflect human's high-order selectional preference about eventualities.
Similar to the evaluation on the lower-order selectional preference, we conduct two experiments (i.e., (1) High frequency vs. Low Frequency; (2) Exist vs. None-exist) on ASER before and after the conceptualization.\footnote{For the experiment on the ASER after the conceptualization, we only sample the conceptualized eventualities and ignore the original ones.}
For the ``High frequency vs. Low Frequency'' experiment, we randomly sample 50 edge pairs for each relation type such that the two edges in each pair share the same head eventuality, relation type, but different tail eventuality (e.g., $\langle$``I am hungry,'' \texttt{Result}, ``I eat food''$\rangle$ versus $\langle$``I am hungry,'' \texttt{Result}, ``I exercise''$\rangle$).
More importantly, the two sampled edges should have significantly different frequencies.
Specifically, we require the frequency of the high-frequency one to be larger than five, and the frequency of the high-frequency one must be at least five times larger than the frequency of the low frequency one.
For the ``Exist vs. None-exist'' experiment, for each relation type, we first randomly sample 50 edges. Then for each edge, we randomly replace a single word of the tail eventuality such that the new tail is very similar to the original one but the created edge does not exist in ASER.

The annotations results are presented in Figure~\ref{fig:high_sp}.
In general, we can make similar observations as the lower SP that the correlation is more significant when we compare the existing and non-existing edges.
Besides that, the experiments on the conceptualized ASER help demonstrate that the conceptualization module will not influence the overall quality of edge frequencies.

\section{Inference over ASER}~\label{sec:inference}

In this section, we first introduce two kinds of inferences (eventuality retrieval and relation retrieval) based on ASER. For each of them, inferences over both one-hop and multi-hop are provided. Complexities of these two retrieval algorithms are $\mathcal{O}(n^k)$, where $n$ is the number of average adjacent eventualities per eventuality and $k$ is the number of hops.
In this section, we show how to conduct these inferences over one-hop and two-hop as the demonstration. ASER is composed of eventualities and concepts. In line with the settings, we conduct case studies over extracted sub-graphs of eventuality and concept graphs. 
After that, we investigate the rule and meta-path-based inferences on ASER. For the rule-based inference, we leverage AMIE+ \cite{DBLP:journals/vldb/GalarragaTHS15}, a rule mining system on ontological knowledge bases (KBs), to discover closed and connected Horn rules on ASER. For the meta-path-based inference, we obtain the frequent meta-paths using statistical methods and perform case studies by instantiating the meta-paths in both eventuality and concept graphs.

\subsection{Eventuality Retrieval}
The eventuality retrieval inference is defined as follows. Given a head eventuality\footnote{ASER also supports the prediction of head eventualities given tail eventualities and relations. We omit it in this section for a clear presentation.} $E_h$ and a relation list $\LM$ = ($T_1, T_2, ..., T_k$), find related eventualities and their associated probabilities such that for each eventuality $E_t$ we can find a path, which contains all the relations in $\LM$ in order from $E_h$ to $E_t$.

\subsubsection{One-hop Inference}

\begin{table}[!ht]
\centering
\small
\begin{tabular}{l|c|l|c}
\toprule  
Head & Relation& Tail & Probability\\
\midrule  
You drink alcohol & $\texttt{Synchronous}$& You drown & 0.50 \\
I drink coffee & $\texttt{Result}$ & I calm down & 0.33 \\
You are an employee & $\texttt{Contrast}$ & You get fired & 0.50 \\
I am programmer & $\texttt{Result}$ & I have free time & 1.00 \\
You go to restaurant & $\texttt{Precedence}$ & You get sick & 0.50 \\
I am frightened & $\texttt{Reason}$ & Dog barks & 0.80 \\
I order chicken & $\texttt{Concession}$ & I am a vegan & 1.00 \\
It is my birthday &  $\texttt{Result}$ & We go to zoo & 0.20 \\
It is a cat & $\texttt{Condition}$ & It is a tiger & 0.67 \\
The surgery goes well & $\texttt{Result}$ & There is no complication & 0.50 \\
\bottomrule
\end{tabular}
\caption{Cases of one-hop eventuality inference in the eventuality graph. In the tables of the case study of eventualities, the words in eventualities are stored as lemmas in KBs. However, to clarify the examples, we manually correct the grammar mistakes.}
\label{tab:event_one_hop_node}
\end{table}

\begin{table}[!h]
\centering
\small
\begin{tabular}{l|c|l|c}
\toprule  
Head& Relation& Tail & Probability\\
\midrule  
\textit{Company} be \textit{Stakeholder-Group} & $\texttt{Condition}$ & \textit{PersonX} be successful & 0.53 \\
\textit{PersonX} have \textit{Issue} & $\texttt{Reason}$ & \textit{PersonX} be proud & 0.52 \\
\textit{PersonX} get \textit{Symptom} &	$\texttt{Synchronous}$	& \textit{PersonX} be \textit{Vulnerable-Group} & 0.50 \\
\textit{PersonX} be \textit{Emotion} & $\texttt{Succession}$ & \textit{PersonX} marry & 0.51 \\
\textit{AnimalX} bark & $\texttt{Result}$ & \textit{AnimalX} kill \textit{AnimalY} & 0.33 \\
\textit{PersonX} be \textit{Predator} & $\texttt{Result}$ & \textit{PersonX} tease \textit{PersonY} & 0.25 \\
\textit{PersonX} do \textit{Academic-Misconduct} & $\texttt{Contrast}$ & \textit{PersonX} tell \textit{Institute} & 0.52 \\
\textit{PersonX} play \textit{Sport} & $\texttt{Reason}$ & \textit{PersonX} love \textit{Activity} & 0.27 \\
\textit{PersonX} hurt \textit{Insect} & $\texttt{Condition}$ & \textit{PersonX} help \textit{Insect} & 0.83 \\
\textit{PersonX} have \textit{Social-Medium} & $\texttt{Result}$ & \textit{PersonX} post it & 0.72 \\

\bottomrule
\end{tabular}
\caption{Cases of one-hop eventuality inference in the concept graph. The concepts are marked as $\textit{italic}$ texts.}
\label{tab:concept_one_hop_node}
\end{table}

For the one-hop inference, we assume the target relation is $T_1$. We then define the probability of any potential tail node $E_t$ as:
\begin{equation}\label{eq:one-hop-eventuality-retrieval}
    \text{Pr}(E_t| E_h, T_1) = 
    \frac{w_{\langle E_h, T_1, E_t \rangle}^{(r)}}{\sum_{E_t^\prime, s.t., (E_h,T_1,E_t^\prime)\in \RM}{w_{\langle E_h, T_1, E_t^\prime \rangle}^{(r)}}},
\end{equation}
where $w_{\langle E_h, T_1, E_t \rangle}^{(r)}$ is the relation weight, which is defined in Definition 3. If no node is connected with $E_h$ via $T_1$, $\text{Pr}(E^\prime | E_h, T_1)$ will be 0 for any $E^\prime \in \EM$. 

Several interesting inference examples are observed. In Table \ref{tab:event_one_hop_node}, we list the reasonable examples of one-hop eventuality inference in the eventuality graph. We also list some of them as follows for discussion:
\begin{itemize}
  \item $\langle$``I drink coffee,'' $\texttt{Reason}$, ``I enjoy the flavor''$\rangle$
  \item $\langle$``You go to restaurant,'' $\texttt{Precedence}$, ``You got sick''$\rangle$
  \item $\langle$``It is a cat,'' $\texttt{Condition}$, ``It is a tiger''$\rangle$
\end{itemize}
It is observed that ``I enjoy the (coffee) flavor'' is likely to be the reason for ``I drink coffee.'' It is also common that if you eat in an unhygienic restaurant, you would probably get sick after you go to the restaurant. Given the fact that the tiger is the largest cat species, it is reasonable to say that if ``it is a tiger,'' ``it is a cat.''

The following examples in Table \ref{tab:concept_one_hop_node} show the results of one-hop eventuality inference in concept graph. 
\begin{itemize}
  \item $\langle$``\textit{Company} be \textit{Stakeholder-Group},'' $\texttt{Condition}$, ``\textit{PersonX} be successful''$\rangle$
  \item $\langle$``\textit{PersonX} hurt \textit{Insect},'' $\texttt{Condition}$, ``\textit{PersonX} help \textit{Insect}''$\rangle$
  \item $\langle$``\textit{PersonX} be \textit{Emotion},'' $\texttt{Succession}$, ``\textit{PersonX} marry''$\rangle$
\end{itemize}
For instance, if someone is successful, his/her company is likely a big corporation of stakeholders. The second one shows a situation that if an unprofessional person helps insects out of good wills, he/she probably hurts them in reverse. We could also infer from the last case that people tend to be emotional when they get married.

\subsubsection{Two-hop Inference}

\begin{table}[!h]
\centering
\resizebox{\textwidth}{!}{%
\begin{tabular}{l|c|l|c|l|c}
\toprule  
Head& Relation1 & Middle &Relation2& Tail & Probability\\
\midrule  
I go to school & $\texttt{Reason}$ & [I admire] & $\texttt{Synchronous}$& I am grown up & 0.50 \\
I go to bed & $\texttt{Conjunction}$ & [I sleep early] & $\texttt{Result}$ & I am healthy & 0.86 \\
We have dinner & $\texttt{Conjunction}$ & [Food is very good] & $\texttt{Contrast}$ & Service is not & 0.95 \\
You go to restaurant & $\texttt{Condition}$ & [They do something right] & $\texttt{Reason}$ & There is a line-up & 0.50 \\
We have lunch & $\texttt{Conjunction}$ & [We really hit it off] & $\texttt{Contrast}$ & She has a boyfriend at time & 0.50 \\
You drink alcohol & $\texttt{Contrast}$ & [You are fine] & $\texttt{Contrast}$ & You have no work  &0.75 \\
I am a vegan & $\texttt{Result}$ & [I do not eat fish] & $\texttt{Contrast}$ & We are hungry & 0.73 \\
I go to bar & $\texttt{Precedence}$ & [Our table is ready] & $\texttt{Result}$ & We take seats &  0.35\\
I go to restaurant & $\texttt{Reason}$ & [I have a coupon] & $\texttt{Contrast}$ & It is expired & 0.36 \\
I go to gym & $\texttt{Precedence}$ & [I go on a date] & $\texttt{Contrast}$ & We have nothing in common & 0.25 \\
\bottomrule
 \end{tabular} }
\caption{Cases of two-hop eventuality inference in the eventuality graph. In the table, we provide a typical example of middle nodes (embraced by brackets) to create a scenario for better understanding.} 
\label{tab:event_two_hop_node}
\end{table}

\begin{table}[!h]
\centering
\resizebox{\textwidth}{!}{%
\begin{tabular}{l|c|l|c|l|c}
\toprule  
Head& Relation1 & Middle &Relation2& Tail & Probability\\
\midrule  
\textit{PersonX} wait for \textit{PersonY} & $\texttt{Precedence}$ & [\textit{PersonX} be tired] & $\texttt{Result}$ & \textit{PersonX} go to sleep & 0.50 \\
\textit{PersonX} hate \textit{Animal} & $\texttt{Contrast}$ & [\textit{PersonX} be harmless] & $\texttt{Contrast}$ & \textit{PersonX} be \textit{Symptom} & 0.40\\
\textit{PersonX} be cranky & $\texttt{Synchronous}$ & [\textit{PersonX} be hungry] & $\texttt{Result}$ & \textit{PersonX} order \textit{Meat} & 0.23 \\
\textit{PersonX} be \textit{Artist} & $\texttt{Contrast}$ & [\textit{PersonX} play \textit{Sport}] & $\texttt{Reason}$ & \textit{PersonX} be strong & 0.33 \\
\textit{PersonX} regret & $\texttt{Condition}$ & [\textit{PersonX} despise \textit{PersonY}] & $\texttt{Reason}$ & \textit{PersonY} be \textit{Performer} & 0.20 \\
\textit{PersonX} pull gun & $\texttt{Reason}$ & [\textit{PersonX} startle] & $\texttt{Synchronous}$ & \textit{Domestic-Animal} bark & 0.50 \\
\textit{Predator} take down \textit{Animal} & $\texttt{Reason}$ & [It be \textit{Predator}] & $\texttt{Synchronous}$ & \textit{PersonX} shoot & 0.32 \\
\textit{PersonX} be \textit{Academic-Title} & $\texttt{Result}$ & [\textit{PersonX} be right] & $\texttt{Contrast}$ & \textit{PersonY} doubt it & 0.28 \\
\textit{PersonX} hear it & $\texttt{Synchronous}$ & [\textit{PersonY} play \textit{Musical-Instrument}] & $\texttt{Synchronous}$ & \textit{PersonY} be blue & 0.65 \\
\textit{PersonX} be \textit{Artist} & $\texttt{Condition}$ & [\textit{PersonX} strike \textit{PersonY}] & $\texttt{Synchronous}$ & \textit{PersonY} interview \textit{PersonX} & 0.40 \\

\bottomrule
\end{tabular}}
\caption{Cases of two-hop eventuality inference in the concept graph. In the table, we provide a typical example of middle nodes (embraced by brackets) to create a scenario for better understanding. The concepts are marked as $\textit{italic}$ texts.}
\label{tab:concept_two_hop_node}
\end{table}

On top of Eq.~(\ref{eq:one-hop-eventuality-retrieval}), it is easy for us to define the probability of $E_t$ on two-hop setting. Assume the two relations are $T_1$ and $T_2$ in order. We can define the probability as follows:
\begin{equation}\label{eq:two-hop-eventuality-retrieval}
    \text{Pr}(E_t| E_h, T_1, T_2) = 
    \sum_{E_m\in \EM_m}{\text{Pr}(E_m|E_h, T_1) \text{Pr}(E_t|E_m, T_2)},
\end{equation}
where $\EM_m$ is the set of intermediate node $E_m$ such that $(E_h,T_1,E_m)$ and $(E_m, T_2, E_t) \in \RM$.

We list the intuitive examples of two-hop eventuality inference in eventuality and concept graph in Table \ref{tab:event_two_hop_node} and Table \ref{tab:concept_two_hop_node}. To better understand the two relations between the head node and the tail node, a typical middle node embraced by brackets is provided. In the eventuality graph, three examples in Table \ref{tab:event_two_hop_node} are given for further explanation.
\begin{itemize}
  \item $\langle$``I go to bed,'' $\texttt{Conjunction}$, [``I sleep early''], $\texttt{Result}$, ``I am healthy''$\rangle$
  \item $\langle$``We have lunch,'' $\texttt{Conjunction}$, [``We really hit it off''], $\texttt{Contrast}$, ``She has a boyfriend at time''$\rangle$
  \item $\langle$``I go to restaurant,'' $\texttt{Reason}$, [``I have a coupon''], $\texttt{Contrast}$, ``It is expired''$\rangle$
\end{itemize}
The first example illustrates that ``I go bed'' and ``I sleep early'' tend to result in ``I am healthy.'' The second one describes a common social situation that I have lunch with a girl and we really hit it off. But she has a boyfriend at that time. Also, it is inferred that the reason why I go to that restaurant is that I have a coupon. However, I find out that the coupon is expired.

Leveraging the same method, we perform two-hop eventuality inference in the concept graph and the results are presented in Table \ref{tab:concept_two_hop_node}. 
\begin{itemize}
  \item $\langle$``\textit{PersonX} wait for \textit{PersonY},'' $\texttt{Precedence}$, [``\textit{PersonX} be tired''], $\texttt{Result}$, ``\textit{PersonX} go to sleep''$\rangle$
  \item $\langle$ ``\textit{PersonX} be cranky,'' $\texttt{Synchronous}$, [``\textit{PersonX} be hungry''], $\texttt{Result}$, ``\textit{PersonX} order \textit{Meat}'' $\rangle$
  \item $\langle$ ``\textit{PersonX} be \textit{Artist},'' $\texttt{Condition}$, [``\textit{PersonX} strike \textit{PersonY}''], $\texttt{Synchronous}$, ``\textit{PersonY} interview \textit{PersonX}'' $\rangle$
\end{itemize}
An interesting example shows that someone is waiting for his/her friend for such a long time that he/she is tired and decides to go to sleep. We also observe that the result of someone being cranky and hungry is most likely to be that he/she orders meats. The last one shows that the artifacts of an artist $\textit{PersonX}$ strikes $\textit{PersonY}$ and it happens at the same time as $\textit{PersonY}$ interviews with $\textit{PersonX}$.

\subsection{Relation Retrieval}

\begin{table}[]
\centering
\small
\begin{tabular}{c|l|l|c}
\toprule  
Relation & Head& Tail & Probability\\
\midrule  
$\texttt{Result}$ & You drink alcohol & You have to pee & 1.00 \\
$\texttt{Result}$ & I drink coffee & I order a cappuccino & 0.50 \\
$\texttt{Alternative}$ & You are a employee & You will be fired & 0.50 \\
$\texttt{Contrast}$ & I eat meat & I am not a steak lover & 1.00 \\
$\texttt{Precedence}$ & You go to sleep & you wake up & 1.00 \\
$\texttt{Contrast}$ & I go to school & I drop out & 0.50\\
$\texttt{Reason}$ & I am not picky & I go to restaurant & 0.43\\
$\texttt{Contrast}$ & I love to cook & I go to restaurant & 0.57\\
$\texttt{Precedence}$ & He waves his hat & The train stops & 1.00\\
$\texttt{Concession}$ & I go to gym & I am tired & 0.83\\

\bottomrule
\end{tabular}
\caption{Cases of one-hop relation inference in the eventuality graph. }
\label{tab:event_one_hop_relation}
\end{table}

\begin{table}[]
\centering
\small
\begin{tabular}{c|c|c|c}
\toprule  
Relation & Head& Tail & Probability\\
\midrule  
$\texttt{Condition}$ & \textit{Company} be \textit{Stakeholder-Group} & \textit{PersonX} do \textit{Local-Ad} & 0.10 \\
$\texttt{Contrast}$ & \textit{PersonX} call \textit{Agency} & It take \textit{Duration} & 0.18 \\
$\texttt{Reason}$ & \textit{PersonX} be \textit{Public-Figure} & \textit{PersonX} be professional & 0.30 \\
$\texttt{Synchronous}$& \textit{Animal} bite & \textit{Animal} be frightened & 0.23\\
$\texttt{Precedence}$ & \textit{PersonX} be \textit{Vulnerable-Group} & \textit{PersonX} quit \textit{Activity} & 0.88 \\
$\texttt{Synchronous}$ & It be \textit{Domestic-Animal} & It be \textit{Mammal} & 0.77 \\
$\texttt{Contrast}$ & \textit{Bird} catch \textit{Animal} & \textit{Animal} get cheese & 0.67 \\
$\texttt{Synchronous}$ & \textit{PersonX} whistle & \textit{Animal} bark & 1.00 \\
$\texttt{Condition}$ & \textit{PersonX} give lecture & \textit{PersonX} be \textit{Academic-Title} & 0.21 \\
$\texttt{Result}$ & \textit{PersonX} play \textit{Sport} & \textit{PersonX} be fit & 0.87 \\

\bottomrule
\end{tabular}
\caption{Cases of one-hop relation inference in the concept graph. The concepts are marked as $\textit{italic}$ texts.}
\label{tab:concept_one_hop_relation}
\end{table}

The relation retrieval inference is defined as follows. Given two nodes $E_h$ and $E_t$, find all relation lists and their probabilities such that for each relation list $\LM$ = ($T_1, T_2, ..., T_k$), we can find a path from $E_h$ to $E_t$, which contains all the relations in $\LM$ in order.

\subsubsection{One-hop Inference}

Assuming that the path length is one, we define the probability of one relation $R = \langle E_h, T, E_t \rangle$ given $E_h$ and $E_t$ as:
\begin{equation}\label{eq:one-hop-plausibility}
    \text{Pr}(R | E_h, E_t) = \text{Pr}(T| E_h, E_t) =
    \frac{w_{\langle E_h, T, E_t \rangle}^{(r)}}{\sum_{T^\prime \in \TM}{w_{\langle E_h, T^\prime, E_t \rangle}^{(r)}}},
\end{equation}
where $\TM$ is the relation type set.

In Table \ref{tab:event_one_hop_relation} and \ref{tab:concept_one_hop_relation}, we perform one-hop relation inference in eventuality and concept graph separately. The relations between head nodes and tail nodes are retrieved to present the commonsense in daily life. In Table \ref{tab:event_one_hop_relation}, the eventuality relations are mined to show frequent patterns in eventuality graph. 
\begin{itemize}
  \item $\langle$ $\texttt{Result}$, ``I drink coffee,'' ``I order cappuccino''$\rangle$
  \item $\langle$ $\texttt{Contrast}$, ``I love to cook,'' ``I go to restaurant'' $\rangle$
  \item $\langle$ $\texttt{Concession}$, ``I go to gym,'' ``I am tired'' $\rangle$
\end{itemize}
For example, ``You drink alcohol'' usually leads to ``You have to pee.'' Another intriguing case illustrates that if someone loves to cook, he/she tends not to go to the restaurant regularly. The last one describes a diligent and determined person who decides to go to the gym, although he/she is tired.

In the concept graph, the same process is used and the results are stored in Table \ref{tab:concept_one_hop_relation}. 
\begin{itemize}
    \item $\langle$ $\texttt{Contrast}$, ``\textit{Bird} catch \textit{Animal},'' ``\textit{Animal} get cheese'' $\rangle$
    \item $\langle$ $\texttt{Condition}$, ``\textit{PersonX} give lecture,'' ``\textit{PersonX} be \textit{Academic-Title}'' $\rangle$
    \item $\langle$ $\texttt{Result}$, ``\textit{PersonX} play \textit{Sport},'' ``\textit{PersonX} be fit'' $\rangle$
\end{itemize}
We learn from the first example that if birds do not catch these animals (e.g., rats), they would probably get cheese. The second one shows that if someone has an academic title (e.g., professor), he/she will deliver a lecture. The third example tells that the result of $\textit{PersonX}$ plays sport is he/she is fit.



\subsubsection{Two-hop Inference}

Similarly, given two nodes $E_h$ and $E_t$, we define the probability of a two-hop connection ($T_1$, $T_2$) between them as follows:

\begin{align}
    \text{Pr}(T_1, T_2 | E_h, E_t)  &= \sum_{E_m\in \EM_m} \text{Pr}(T_1, T_2, E_m | E_h, E_t) \nonumber\\
    &=\sum_{E_m\in \EM_m} \text{Pr}(T_1 | E_h) \text{Pr}(E_m | T_1, E_h) \text{Pr}(T_2 | E_m, E_t),
\end{align}
where $\text{Pr}(T | E_h)$ is the probability of a relation type $T$ given a head eventuality $E_h$, which is defined as follows:
\begin{equation}\label{eq:relation-probability}
    \text{Pr}(T | E_h)=
    \frac{\sum_{E_t, s.t., (E_h, T, E_t) \in \RM}{w_{\langle E_h, T, E_t \rangle}^{(r)}}}{\sum_{T^\prime\in \TM} \sum_{E_t, s.t., (E_h, T^\prime, E_t) \in \TM}{w_{\langle E_h, T^\prime, E_t \rangle}^{(r)}}}.
\end{equation}

The two-hop relations are inferred from the eventuality and concept graph in Table \ref{tab:event_two_hop_relation} and \ref{tab:concept_two_hop_relation}. Some reasonable results are listed below. In line with two-hop eventuality inference, we give a typical middle node embraced by brackets to show the circumstance more clearly.
\begin{itemize}
    \item $\langle$ $\texttt{Synchronous}$, $\texttt{Conjunction}$, ``We have breakfast,'' [``Our room is ready''], ``The front desk staff is friendly'' $\rangle$
    \item $\langle$ $\texttt{Synchronous}$, $\texttt{Reason}$, ``I sit on chair,'' [``I get my hair washed''], ``Stylist tells me'' $\rangle$
    \item $\langle$ $\texttt{Reason}$, $\texttt{Result}$, ``I go to supermarket,'' [``I have a coupon''], ``The price is great'' $\rangle$
\end{itemize}
In the eventuality graph, we find that some tourists visit a hotel. The guests have breakfast when their room is cleaned and ready. Meanwhile, they find the staff at the front desk is friendly and nice. The second example shows a common thing at the haircut salons. ``I sit on chair'' to get my hair washed because the stylist tells me to do so before haircut. We also find that someone goes to a supermarket because he/she has a coupon to lower the prices of groceries.

\begin{table}[]
\centering
\resizebox{\textwidth}{!}{%
\begin{tabular}{c|c|l|l|l|c}
\toprule  
Relation1 & Relation2 & Head & Middle & Tail & Probability\\
\midrule 
$\texttt{Conjunction}$ & $\texttt{Synchronous}$& I go to bed & [I would sleep] & I heal & 0.25 \\
$\texttt{Result}$ & $\texttt{Reason}$ & I go to bed & [I fall right to sleep] & I am drunk & 0.30 \\
$\texttt{Synchronous}$& $\texttt{Conjunction}$ & We have breakfast & [Our room is ready] & The front desk staffs are friendly & 0.12 \\
$\texttt{Contrast}$ & $\texttt{Precedence}$ & You are an employee & [You get fired] & The contract ends & 0.11 \\
$\texttt{Reason}$ & $\texttt{Conjunction}$ & I drink coffee & [I have severe ADHD] & I do not get any help & 0.33 \\
$\texttt{Synchronous}$& $\texttt{Reason}$ & I sit on chair & [I get my hair wash] & Stylist tell me & 0.25 \\
$\texttt{Contrast}$ & $\texttt{Reason}$ & I am a vegan & [I eat meat] & It tastes good & 0.33\\
$\texttt{Reason}$ & $\texttt{Result}$ & I go to supermarket & [I have a coupon] & The price is great & 0.23\\
$\texttt{Reason}$ & $\texttt{Contrast}$ & I go to restaurant & [The service is great] & The food is mediocre & 0.49\\
$\texttt{Contrast}$ & $\texttt{Contrast}$ & The surgery goes well & [She is in a coma] & She is stabilized & 0.53\\

\bottomrule
\end{tabular}}
\caption{Cases of two-hop relation inference in the eventuality graph. In the table, we provide a typical example of middle nodes (embraced by brackets) to create a scenario for better understanding.}
\label{tab:event_two_hop_relation}
\end{table}

\begin{table}[]
\centering
\resizebox{\textwidth}{!}{%
\begin{tabular}{c|c|l|l|l|c}
\toprule  
Relation1 & Relation2 & Head & Middle & Tail & Probability\\
\midrule 
$\texttt{Precedence}$ & $\texttt{Precedence}$ & \textit{PersonX} be \textit{Stakeholder} & [\textit{PersonX} tell \textit{PersonX}] & \textit{PersonX} sign \textit{Document} & 0.85 \\
$\texttt{Precedence}$ & $\texttt{Precedence}$ & \textit{PersonX} wait for \textit{PersonX} & [\textit{PersonX} send \textit{Information}] & \textit{PersonX} drag \textit{PersonX} away & 0.40 \\
$\texttt{Result}$ & $\texttt{Precedence}$ & \textit{PersonX} be \textit{Vulnerable-Population} & [\textit{PersonX} be homeless] & \textit{Organization} help \textit{PersonX} & 0.73 \\
$\texttt{Precedence}$ & $\texttt{Result}$ & \textit{Predator} catch \textit{Herbivore} & [\textit{Predator} eat \textit{Meat}] & \textit{Mammal} live & 0.79\\
$\texttt{Synchronous}$ & $\texttt{Result}$ & \textit{PersonX} be thirsty & [\textit{PersonX} be hungry] & \textit{PersonX} order \textit{Meat} & 0.56\\
$\texttt{Contrast}$ & $\texttt{Condition}$ & \textit{PersonX} be \textit{Musician} & [\textit{PersonX} play \textit{Sport}] & \textit{PersonX} be tall & 0.42\\
$\texttt{Result}$ & $\texttt{Synchronous}$ & \textit{PersonX} excite & [\textit{PersonX} imagine \textit{Emotion}] & \textit{PersonX} answer \textit{Electronic-Device} & 0.58\\
$\texttt{Reason}$ & $\texttt{Contrast}$ & \textit{PersonX} eat \textit{Animal-Product} & [\textit{PersonX} enjoy it] & It be spicy & 0.25\\
$\texttt{Synchronous}$ & $\texttt{Condition}$ & \textit{PersonX} hear it & [\textit{PersonX} play \textit{Musical-Instrument}] & \textit{PersonX} be \textit{Extracurricular-Activity} & 0.16\\

$\texttt{Alternative}$ & $\texttt{Result}$ & \textit{PersonX} eat \textit{Meat} & [\textit{PersonX} be vegetarian] & \textit{PersonX} starve & 0.50\\

\bottomrule
\end{tabular}}
\caption{Cases of two-hop relation inference in the concept graph. In the table, we provide a typical example of middle nodes (embraced by brackets) to create a scenario for better understanding. The concepts are marked as $\textit{italic}$ texts.}
\label{tab:concept_two_hop_relation}
\end{table}

In Table \ref{tab:concept_two_hop_relation}, the two-hop relations among head, middle, and tail nodes are extracted to show some insights behind the relation inference. 
\begin{itemize}
    \item $\langle$ $\texttt{Precedence}$, $\texttt{Precedence}$, ``\textit{PersonX} wait for \textit{PersonY},'' [``\textit{PersonY} send \textit{Information}''], ``\textit{PersonZ} drag \textit{PersonX} away'' $\rangle$
    \item $\langle$ $\texttt{Result}$, $\texttt{Synchronous}$, ``\textit{PersonX} excite,'' [``\textit{PersonX} imagine \textit{Emotion}''], ``\textit{PersonX} answer \textit{Electronic-Device}'' $\rangle$
    \item $\langle$ $\texttt{Synchronous}$, $\texttt{Condition}$, ``\textit{PersonX} hear it,'' [``\textit{PersonX} play \textit{Musical-Instrument}''], ``\textit{PersonX} be \textit{Extracurricular-Activity}'' $\rangle$
\end{itemize}
In the first example, $\textit{PersonX}$ waits for $\textit{PersonY}$ before $\textit{PersonY}$ sends information (e.g., a letter) to inform $\textit{PersonX}$ not wait for him. $\textit{PersonX}$ is reluctant to leave until his/her friends drag him/her away. In the latter example, the result of a person being excited is that he/she imagines the situation and answers the phone. The third case shows that $\textit{PersonX}$ hears that $\textit{PersonY}$ plays an instrument since $\textit{PersonY}$ is having extracurricular activities.



\subsection{Rule Mining}

\begin{table}[!ht]
    \centering
    \small
    \resizebox{\textwidth}{!}{
    \begin{tabular}{l|l}
    \toprule
    Rule & $\langle E_b \xrightarrow{\texttt{Concession}} E_f \rangle \land \langle E_a \xrightarrow{\texttt{Result}} E_f \rangle \Rightarrow \langle E_a \xrightarrow{\texttt{Contrast}} E_b \rangle$ \\
    \hline
    Instances & $\langle$ I do not know $\rightarrow$ I guess $\rangle \land \langle$ I believe $\rightarrow$ I guess $\rangle \Rightarrow \langle$ I believe $\rightarrow$ I do not know $\rangle$ \\
         & $\langle$ I am not sure $\rightarrow$ I guess $\rangle \land \langle$ I hope so $\rightarrow$ I guess $\rangle \Rightarrow \langle$ I hope so $\rightarrow$ I am not sure $\rangle$ \\
         & $\langle$ I understand $\rightarrow$ I can not speak $\rangle \land \langle$ I am not a lawyer $\rightarrow$ I can not speak $\rangle \Rightarrow \langle$ I am not a lawyer $\rightarrow$ I understand $\rangle$\\
    \midrule
    Rule & $\langle E_f \xrightarrow{\texttt{Contrast}} E_b \rangle \land \langle E_a \xrightarrow{\texttt{Instantiation}} E_f \rangle \Rightarrow \langle E_a \xrightarrow{\texttt{Contrast}} E_b \rangle$\\
    \hline
    Instances & $\langle$ I remember $\rightarrow$ I could not find it $\rangle \land \langle$ I get $\rightarrow$ I remember $\rangle \Rightarrow \langle$ I get $\rightarrow$ I could not find it $\rangle$\\
         & $\langle$ I would say $\rightarrow$ I might be wrong $\rangle \land \langle$ I hope $\rightarrow$ I would say $\rangle \Rightarrow \langle$ I hope $\rightarrow$ I might be wrong $\rangle$\\
         & $\langle$ It have been suggested $\rightarrow$ This is unlikely $\rangle \land \langle$ It is possible $\rightarrow$ It have been suggested $\rangle \Rightarrow \langle$ It is possible $\rightarrow$ This is unlikely $\rangle$\\
    \midrule
    Rule & $\langle E_e \xrightarrow{\texttt{ChosenAlternative}} E_b \rangle \land \langle E_a \xrightarrow{\texttt{ChosenAlternative}} E_e \rangle \Rightarrow \langle E_a \xrightarrow{\texttt{ChosenAlternative}} E_b \rangle$ \\
    \hline
    Instances & $\langle$ I will not go $\rightarrow$ You come here $\rangle \land \langle$ I want to see $\rightarrow$ I will not go $\rangle \Rightarrow \langle$ I want to see $\rightarrow$ You come here $\rangle$\\
         & $\langle$ I want $\rightarrow$ It is $\rangle \land \langle$ I wish $\rightarrow$ I want $\rangle \Rightarrow \langle$ I wish $\rightarrow$ It is $\rangle$\\
         & $\langle$ I want $\rightarrow$ I get $\rangle \land \langle$ I do not get that $\rightarrow$ I want $\rangle \Rightarrow \langle$ I do not get that $\rightarrow$ I get $\rangle$\\
    \midrule
    Rule & $\langle E_a \xrightarrow{\texttt{Reason}} E_e \rangle \land \langle E_e \xrightarrow{\texttt{Restatement}} E_b \rangle \Rightarrow \langle E_a \xrightarrow{\texttt{Reason}} E_b \rangle$ \\
    \hline
    Instances & $\langle$ I have ever see $\rightarrow$ I know $\rangle \land \langle$ I know $\rightarrow$ They are $\rangle \Rightarrow \langle$ I have ever see $\rightarrow$ They are $\rangle$\\
         & $\langle$ I am curious $\rightarrow$ I think $\rangle \land \langle$ I think $\rightarrow$ It seems $\rangle \Rightarrow \langle$ I am curious $\rightarrow$ It seems $\rangle$\\
         & $\langle$ It is not $\rightarrow$ You are lying $\rangle \land \langle$ You are lying $\rightarrow$ I do not believe you $\rangle \Rightarrow \langle$ It is not $\rightarrow$ I do not believe you $\rangle$\\
    \midrule
    Rule & $\langle E_a \xrightarrow{\texttt{Concession}} E_f \rangle \land \langle E_b \xrightarrow{\texttt{Reason}} E_f \rangle \Rightarrow \langle E_a \xrightarrow{\texttt{Contrast}} E_b \rangle$ \\
    \hline
    Instances & $\langle$ I have no clue $\rightarrow$ I hope $\rangle \land \langle$ It be $\rightarrow$ I hope $\rangle \Rightarrow \langle$ I have no clue $\rightarrow$ It is $\rangle$\\
         & $\langle$ I reckon $\rightarrow$ I do not know $\rangle \land \langle$ I can not talk about it $\rightarrow$ I do not know $\rangle \Rightarrow \langle$ I reckon $\rightarrow$ I can not talk about it $\rangle$\\
         & $\langle$ You do not understand it $\rightarrow$ You are admitted $\rangle \land \langle$ That is $\rightarrow$ You are admitted $\rangle \Rightarrow \langle$ You do not understand it $\rightarrow$ That is $\rangle$\\
    \midrule
    Rule & $\langle E_b \xrightarrow{\texttt{Alternative}} E_f \rangle \land \langle E_a \xrightarrow{\texttt{Result}} E_f \rangle \Rightarrow \langle E_a \xrightarrow{\texttt{Contrast}} E_b \rangle$ \\
    \hline
    Instances & $\langle$ I am going $\rightarrow$ I am not going $\rangle \land \langle$ I do not care $\rightarrow$ I am not going $\rangle \Rightarrow \langle$ I do not care $\rightarrow$ I am going $\rangle$\\
         & $\langle$ You do $\rightarrow$ I do $\rangle \land \langle$ I suppose $\rightarrow$ I do $\rangle \Rightarrow \langle$ I suppose $\rightarrow$ You do $\rangle$\\
         & $\langle$ I reckon $\rightarrow$ I guess $\rangle \land \langle$ I wonder $\rightarrow$ I guess $\rangle \Rightarrow \langle$ I wonder $\rightarrow$ I reckon $\rangle$\\
    \midrule
    Rule & $\langle E_a \xrightarrow{\texttt{Reason}} E_f \rangle \land \langle E_b \xrightarrow{\texttt{Succession}} E_f \rangle \Rightarrow \langle E_a \xrightarrow{\texttt{Reason}} E_b \rangle$ \\
    \hline
    Instances & $\langle$ I ask $\rightarrow$ I am not sure $\rangle \land \langle $ I do not know $\rightarrow$ I am not sure $\rangle \Rightarrow \langle$ I ask $\rightarrow$ I do not know $\rangle$\\
         & $\langle$ We are lucky $\rightarrow$ We notice $\rangle \land \langle$ We order $\rightarrow$ We notice $\rangle \Rightarrow \langle$ We are lucky $\rightarrow$ We order $\rangle$\\
         & $\langle$ I remember it $\rightarrow$ I see it $\rangle \land \langle$ I realize $\rightarrow$ I see it $\rangle \Rightarrow \langle$ I remember it $\rightarrow$ I realize $\rangle$\\
    \midrule
    Rule & $\langle E_a \xrightarrow{\texttt{Concession}} E_f \rangle \land \langle E_b \xrightarrow{\texttt{Precedence}} E_f \rangle \Rightarrow \langle E_a \xrightarrow{\texttt{Contrast}} E_b \rangle$ \\
    \hline
    Instances & $\langle$ I am unconscious $\rightarrow$ I wake up $\rangle \land \langle $ I see $\rightarrow$ I wake up $\rangle \Rightarrow \langle$ I am unconscious $\rightarrow$ I see $\rangle$\\
         & $\langle$ I swear $\rightarrow$ I guess $\rangle \land \langle$ I do not know $\rightarrow$ I guess $\rangle \Rightarrow \langle$ I swear $\rightarrow$ I do not know $\rangle$\\
         & $\langle$ I can not believe $\rightarrow$ It is great $\rangle \land \langle$ I think $\rightarrow$ It is great $\rangle \Rightarrow \langle$ I can not believe $\rightarrow$ I think $\rangle$\\
    \midrule
    Rule & $\langle E_a \xrightarrow{\texttt{Alternative}} E_e \rangle \land \langle E_e \xrightarrow{\texttt{Exception}} E_b \rangle \Rightarrow \langle E_a \xrightarrow{\texttt{Exception}} E_b \rangle$ \\
    \hline
    Instances & $\langle$ It is not $\rightarrow$ It is wrong $\rangle \land \langle$ It is wrong $\rightarrow$ It is $\rangle \Rightarrow \langle$ It is not $\rightarrow$ It is $\rangle$\\
         & $\langle$ I really want $\rightarrow$ I think $\rangle \land \langle$ I think $\rightarrow$ I know $\rangle \Rightarrow \langle$ I really want $\rightarrow$ I know $\rangle$\\
         & $\langle$ It is not $\rightarrow$ I suppose $\rangle \land \langle$ I suppose $\rightarrow$ You know $\rangle \Rightarrow \langle$ It is not $\rightarrow$ You know $\rangle$\\
    \midrule
    Rule & $\langle E_a \xrightarrow{\texttt{ChosenAlternative}} E_f \rangle \land E_b \langle \xrightarrow{\texttt{ChosenAlternative}} E_f \rangle \Rightarrow \langle E_a \xrightarrow{\texttt{Restatement}} E_b \rangle$ \\
    \hline
    Instances & $\langle$ I am hoping $\rightarrow$ We get $\rangle \land \langle$ I think $\rightarrow$ We get $\rangle \Rightarrow \langle$ I am hoping $\rightarrow$ I think $\rangle$\\
         & $\langle$ I suppose $\rightarrow$ He is $\rangle \land \langle$ I think $\rightarrow$ He is $\rangle \Rightarrow \langle$ I suppose $\rightarrow$ I think $\rangle$\\
         & $\langle$ I am glad $\rightarrow$ I think $\rangle \land \langle$ The food is good $\rightarrow$ I think $\rangle \Rightarrow \langle$ I am glad $\rightarrow$ The food is good $\rangle$\\
    \bottomrule
    \end{tabular}}
    \caption{Cases of AMIE+ rule mining in the eventuality graph. For the simplicity of formatting, we represent $\langle E_h, T, E_t \rangle$ triples as $\langle E_h \xrightarrow{\texttt{T}} E_t \rangle$.}
    \label{tab:amie_aser}
\end{table}

AMIE+ \cite{DBLP:journals/vldb/GalarragaTHS15} aims at mining close and connected Horn Rules in the form of
$$\langle E_a, T_1, E_b \rangle \land \langle E_b, T_2, E_c \rangle \Rightarrow \langle E_a, T_3, E_b \rangle$$
where $E_a, E_b, E_c$ and $T_1, T_2, T_3$ are eventuality variables and relation variables, respectively. As demonstrated through experiments on different KBs, AMIE+ provides an effective approach for the investigation and inference over KB from a logic-rule perspective. We therefore apply AMIE+ on ASER to probe whether it preserves logical properties among multi-hop eventualities and relations. To make the notation consistent with AMIE+, we denote a fact triple with variables at head and/or tail eventuality positions, such as $\langle E_a, T_1, E_b \rangle$, as an \textit{atom}. A rule of interests 
comprise of a body of a set of atoms $B_1, B_2, \cdots, B_n$ (n = 2 in this case), and a head of a single atom $\langle E_h, T, E_t \rangle$. We abbreviate the rule as $\Vec{B} \, \Rightarrow \, \langle E_h, T, E_t \rangle$

The Algorithm of AMIE+ adopts an iterative procedure: starting with a queue of all possible items (rule of size 1), it dequeues a rule in each iteration, and outputs/grows/prunes the rule with certain criterion (listed below), and enqueues the grown new rules back. Throughout the iteration, AMIE+ sets the following significance criterion:

\textit{Head coverage}: 
$$hc(\Vec{B} \, \Rightarrow \, \langle E_h, T, E_t \rangle) := \frac{supp(\Vec{B} \, \Rightarrow \, \langle E_h, T, E_t \rangle)}{size(T)},$$
where $$supp(\Vec{B} \, \Rightarrow \, \langle E_h, T, E_t \rangle) := \#(E_h, E_t): \exists \, z_1, \cdots, z_m: \Vec{B} \land \langle E_h, T, E_t \rangle: $$ denotes the support, i.e., the number of correct prediction yielded with the rule in the current KB, and $size(T)$ denotes the number of facts with $T$ as relations.

\textit{Standard confidence}: 
$$conf(\Vec{B} \, \Rightarrow \, \langle E_h, T, E_t \rangle) := \frac{supp(\Vec{B} \, \Rightarrow \, \langle E_h, T, E_t \rangle)}{\#(E_h,E_t): \exists \, z_1, \cdots, z_m : \Vec{B}},$$
where $\#(E_h,E_t): \exists z_1, \cdots, z_m : \Vec{B}$ denotes all possible predictions of the rule.

\textit{PCA confidence}: 
$$conf_{pca}(\Vec{B} \, \Rightarrow \, \langle E_h, T, E_t \rangle) := \frac{supp(\Vec{B} \, \Rightarrow \, \langle E_h, T, E_t \rangle)}{\#(E_h,E_t): \exists \, z_1, \cdots, z_m : \Vec{B} \, \land \langle E_h, T, E^{\prime}_t \rangle},$$
where $\#(E_h,E_t): \exists \, z_1, \cdots, z_m : \Vec{B} \, \land \langle E_h, T, E^{\prime}_t \rangle$ denotes the number of pairs of $(E_h, E_t)$ predicted by corresponding relation body $\Vec{B}$ but with an existing pair of $\langle E_h, T, E^{\prime}_t \rangle$ in KB.

AMIE+ focuses on RDF Knowledge Bases, where an RDF KB could be represented as a set of facts in the form $\langle {Subject}, \texttt{Relation}, {Object} \rangle$. To pair ASER Graph with AMIE+, we extract all $\langle E_1, \texttt{Relation}, E_2 \rangle$ triples from the relation table as our set of facts. To preserve the frequency information, we duplicate each extracted fact for $f$ times where $f$ is the corresponding triple frequency in the relation table. During our experiments, we set the threshold of minimal PCA confidence \textit{minPCA}=0.1 and minimal head coverage \textit{minHC}=0.01, and run AMIE+ on both the eventuality graph and concept graph of ASER. 

Some of the mined rules and instantiated cases are shown in Table \ref{tab:amie_aser} and \ref{tab:amie_concept}, respectively. 
For the eventuality graph, the rule ``$\langle E_b \xrightarrow{\texttt{Concession}} E_f \rangle \land \langle E_a \xrightarrow{\texttt{Result}} E_f \rangle \Rightarrow \langle E_a \xrightarrow{\texttt{Contrast}} E_b \rangle$'' demonstrates that if the result of some event is opposite to what should happen, then the reason that induces this ``opposite'' result is likely to contrast the original event in some core properties. The last instance ``$\langle$ I understand $\rightarrow$ I can not speak $\rangle \land \langle$ I am not a lawyer $\rightarrow$ I can not speak $\rangle \Rightarrow \langle$ I am not a lawyer $\rightarrow$ I understand $\rangle$'' illustrates this rule with a lawsuit scenario where the ``opposite'' result "I can not speak" is induced by the reason ``I am not a lawyer,'' which contrasts the original event ``I understand.''
And the rule ``$\langle E_a \xrightarrow{\texttt{ChosenAlternative}} E_f \rangle \land E_b \langle \xrightarrow{\texttt{ChosenAlternative}} E_f \rangle \Rightarrow \langle E_a \xrightarrow{\texttt{Restatement}} E_b \rangle$'' captures that if the subject consistently prefers two events as alternative to a third event, then it is possible that the first two events have very similar semantic meaning. This is illustrated with its last instance ``$\langle$ I am glad $\rightarrow$ I think $\rangle \land \langle$ The food is good $\rightarrow$ I think $\rangle \Rightarrow \langle$ I am glad $\rightarrow$ The food is good $\rangle$'' where both ``I am glad'' and ``The food is good'' express similar meanings in which the subject consistently prefers to ``I think.''

For the concept graph, the rule ``$\langle E_e \xrightarrow{\texttt{Instantiation}} E_a \rangle \land \langle E_e \xrightarrow{\texttt{Instantiation}} E_b \rangle \Rightarrow \langle E_a \xrightarrow{\texttt{Conjunction}} E_b \rangle$'' shows that the concepts that both describe a shared concept in details are likely to both happen. Its first instance ``$\langle$ \textit{PersonX} realize $\rightarrow$ \textit{PersonX} point out $\rangle \land \langle$ \textit{PersonX} realize $\rightarrow$ \textit{PersonX} have \textit{Information} $\rangle \Rightarrow \langle$ \textit{PersonX} point out $\rightarrow$ \textit{PersonX} have \textit{Information}'' demonstrates this through a information-capture process, where the two concepts ``\textit{PersonX} point out'' and ``\textit{PersonX} have \textit{Information} $\rangle$'' that both describe the concept ``\textit{PersonX} realize'' happen together. 
The rule ``$\langle E_e \xrightarrow{\texttt{Exception}} E_b \rangle \land \langle E_e \xrightarrow{\texttt{Succession}} E_a \rangle \Rightarrow \langle E_a \xrightarrow{\texttt{Contrast}} E_b \rangle$'' shows the association between $\texttt{Exception}$ and $\texttt{Contrast}$ that is bridged via $\texttt{Succession}$. This is demonstrated with its first instance ``$\langle$ \textit{Item} be ready $\rightarrow$ \textit{PersonX} wait $\rangle \land \langle$ \textit{Item} be ready $\rightarrow$ \textit{PersonX} check $\rangle \Rightarrow \langle$ \textit{PersonX} check $\rightarrow$ \textit{PersonX} be wait $\rangle$'' that shows a scenario of a customer checking and waiting for products.

\begin{table}[]
    \centering
    \small
    \resizebox{\textwidth}{!}{
    \begin{tabular}{l|l}
    \toprule
    Rule & $\langle E_e \xrightarrow{\texttt{Restatement}} E_a \rangle \land \langle E_e \xrightarrow{\texttt{Restatement}} E_b \rangle \Rightarrow \langle E_a \xrightarrow{\texttt{Conjunction}} E_b \rangle$ \\
    \hline
    Instances & $\langle$ \textit{PersonX} laugh $\rightarrow$ \textit{PersonX} smile $\rangle \land \langle$ \textit{PersonX} laugh $\rightarrow$ \textit{PersonX} open \textit{Facial-Feature} $\rangle \Rightarrow \langle$ \textit{PersonX} smile $\rightarrow$ \textit{PersonX} open \textit{Facial-Feature} $\rangle$\\
         & $\langle$ \textit{PersonX} love it $\rightarrow$ It be good $\rangle \land \langle$ \textit{PersonX} love it $\rightarrow$ It be tasty $\rangle \Rightarrow \langle$ It be good $\rightarrow$ It be tasty $\rangle$\\
         & $\langle$ \textit{PersonX} wish $\rightarrow$ \textit{PersonX} need $\rangle \land \langle$ \textit{PersonX} wish $\rightarrow$ \textit{PersonX} need $\rangle \Rightarrow \langle$ \textit{PersonX} need $\rightarrow$ \textit{PersonX} need $\rangle$\\
    \midrule
    Rule & $\langle E_e \xrightarrow{\texttt{Instantiation}} E_a \rangle \land \langle E_e \xrightarrow{\texttt{Instantiation}} E_b \rangle \Rightarrow \langle E_a \xrightarrow{\texttt{Conjunction}} E_b \rangle$ \\
    \hline
    Instances & $\langle$ \textit{PersonX} realize $\rightarrow$ \textit{PersonX} point out $\rangle \land \langle$ \textit{PersonX} realize $\rightarrow$ PersonX have \textit{Information} $\rangle \Rightarrow \langle$ \textit{PersonX} point out $\rightarrow$ \textit{PersonX} have \textit{Information} $\rangle$\\
         & $\langle$ \textit{PersonX} have $\rightarrow$ \textit{PersonX} get $\rangle \land \langle$ \textit{PersonX} have $\rightarrow$ \textit{PersonX} own $\rangle \Rightarrow \langle$ \textit{PersonX} get $\rightarrow$ \textit{PersonX} own $\rangle$\\
         & $\langle$ \textit{PersonX} know $\rightarrow$ \textit{PersonX} be sure $\rangle \land \langle$ \textit{PersonX} know $\rightarrow$ \textit{PersonX} remember $\rangle \Rightarrow \langle$ \textit{PersonX} be sure $\rightarrow$ \textit{PersonX} remember $\rangle$\\
    \midrule
    Rule & $\langle E_e \xrightarrow{\texttt{Concession}} E_b \rangle \land \langle E_e \xrightarrow{\texttt{Restatement}} E_a \rangle \Rightarrow \langle E_a \xrightarrow{\texttt{Contrast}} E_b \rangle$ \\
    \hline
    Instances & $\langle$ \textit{PersonX} order \textit{Dish} $\rightarrow$ \textit{PersonX} be hungry $\rangle \land \langle$ \textit{PersonX} order \textit{Dish} $\rightarrow$ \textit{PersonX} order $\rangle \Rightarrow \langle$ \textit{PersonX} order $\rightarrow$ \textit{PersonX} be hungry $\rangle$\\
         & $\langle$ \textit{PersonX} wish $\rightarrow$ \textit{PersonX} doubt $\rangle \land \langle$ \textit{PersonX} wish $\rightarrow$ \textit{PersonX} need $\rangle \Rightarrow \langle$ \textit{PersonX} doubt $\rightarrow$ \textit{PersonX} need $\rangle$\\
         & $\langle$ \textit{PersonX} love it $\rightarrow$ \textit{PersonX} hate it $\rangle \land \langle$ \textit{PersonX} love it $\rightarrow$ It be good $\rangle \Rightarrow \langle$ \textit{PersonX} hate it $\rightarrow$ It be good $\rangle$\\
    \midrule
    Rule & $\langle E_e \xrightarrow{\texttt{Exception}} E_b \rangle \land \langle E_e  \xrightarrow{\texttt{Succession}} E_a \rangle \Rightarrow E_a \langle \xrightarrow{\texttt{Contrast}} E_b \rangle$ \\
    \hline
    Instances & $\langle$ \textit{Item} be ready $\rightarrow$ \textit{PersonX} wait $\rangle \land \langle$ \textit{Item} be ready $\rightarrow$ \textit{PersonX} check $\rangle \Rightarrow \langle$ \textit{PersonX} check $\rightarrow$ \textit{PersonX} be wait $\rangle$\\
         & $\langle$ \textit{PersonX} say $\rightarrow$ \textit{PersonX} be sorry $\rangle \land \langle$ \textit{PersonX} say $\rightarrow$ \textit{PersonX} be surprised $\rangle \Rightarrow \langle$ \textit{PersonX} be sorry $\rightarrow$ \textit{PersonX} be surprised $\rangle$\\
         & $\langle$ It be $\rightarrow$ \textit{PersonX} guess $\land$ It be $\rightarrow$ It be \textit{factor} $\rangle \Rightarrow \langle$ \textit{PersonX} guess $\rightarrow$ It be \textit{factor} $\rangle$\\
    \midrule
    Rule & $\langle E_a \xrightarrow{\texttt{Restatement}} E_f \rangle \land \langle E_b \xrightarrow{\texttt{Restatement}} E_f \rangle \Rightarrow \langle E_a \xrightarrow{\texttt{Synchronous}} E_b \rangle$ \\
    \hline
    Instances & $\langle$ \textit{PersonX} love it $\rightarrow$ It be good $\rangle \land \langle$ \textit{PersonX} feel $\rightarrow$ It be good $\rangle \Rightarrow \langle$ \textit{PersonX} love it $\rightarrow$ \textit{PersonX} feel $\rangle$\\
         & $\langle$ It be cool $\rightarrow$ It be good $\rangle \land \langle$ \textit{PersonX} think $\rightarrow$ It be okay $\rangle \Rightarrow \langle$ It be cool $\rightarrow$ It be okay $\rangle$\\
         & $\langle$ \textit{PersonX} like it $\rightarrow$ It be good $\rangle \land \langle$ \textit{PersonX} be happy $\rightarrow$ It be good $\rangle \Rightarrow \langle$ \textit{PersonX} like it $\rightarrow$ \textit{PersonX} be happy $\rangle$\\
    \bottomrule
    \end{tabular}}
    \caption{Cases of AMIE+ rule mining in the concept graph. For the simplicity of formatting, we represent $\langle E_h, T, E_t \rangle$ triples as $\langle E_h \xrightarrow{\texttt{T}} E_t \rangle$.}
    \label{tab:amie_concept}
\end{table}



\subsection{Meta-path Mining}

ASER is a complex heterogeneous graph that encodes the commonsense knowledge. ASER is composed of two types of nodes (i.e., extracted eventuality and conceptualized eventuality) and 15 types of edges (e.g., $\texttt{Reason}$ and $\texttt{Precedence}$). We leverage meta-path \cite{sun2012metapath} mining which studies the semantic meanings behind paths to tackle the heterogeneity of ASER. A meta-path is a path that consists of a sequence of different relations defined among various node types. Formally, a meta-path $P$ is defined as a path $ E_1 \xrightarrow{T_1} E_2 \xrightarrow{T_2} \cdots \xrightarrow{T_{l-1}} E_l$, in which $T=T_1 \circ T_2 \circ \cdots \circ T_l$ is the composite relation between $N_1$ and $N_l$. Take an example from Table \ref{tab:metapath}, the meta-path ``$E_1 \xrightarrow{\texttt{Conceptualization}} C_1   \xrightarrow{\texttt{ ConceptInstantiation}} E_2$'' defines a composite relation in which the two eventuality $E_1$ and $E_2$ are conceptualized to the same concept $C_1$.

To automatically select the most frequent and influential meta-paths, we first perform a random walk on the hybrid graph. Specifically, 50,000 seed nodes are chosen independently and uniformly from the nodes of ASER. Starting from each seed node, a random walk is used to generate 50 multi-hop paths of different nodes and relations. The nodes in a path are represented by their types rather than their contents. For example, a path ``I drink coffee $\xrightarrow{\texttt{Result}}$ I stay up late'' is converted into a meta-path ``$E_1 \xrightarrow{\texttt{Result}} E_2$.'' After collecting meta-paths, we search for the frequent patterns of 2-hop and 3-hop meta-paths. The appearance of meta paths is counted. The 2-hop and 3-hop meta paths are later ranked by their frequencies. The frequent meta paths are selected for the further case study. We list the intriguing instances from these meta-paths in Table \ref{tab:metapath}.   

For 2-hop meta paths, the results are very similar to the ones of eventuality/relation retrieval inference. For example, ``$E_1 \xrightarrow{\texttt{Reason}} E_2 \xrightarrow{\texttt{Result}} E_3$'' describes paths following cause and effect relations. A typical instance in daily life is ``I am in pain$\rightarrow$I am alone$\rightarrow$I sit at bar,'' describing a scenario in which a man suffers from loneliness and goes to the bar to numb the pains. In addition to relations among eventualities, the interaction between concepts and eventualities are also discovered by the meta-paths. In the cases of ``$E_1 \xrightarrow{\texttt{Conceptualization}} C_1 \xrightarrow{\texttt{ConceptInstantiation}} E_2$,'' two semantically distinct eventualities are unified in the concept-level. For example, ``He is psychiatrist'' and ``I am attorney'' follows the same pattern, ``$\textit{PersonX}$ be specialist.''

For 3-hop meta paths, the reasoning paths are longer and illustrates the daily life in more details. For example, in the meta-path ``$E_1 \xrightarrow{\texttt{Result}} E_2 \xrightarrow{\texttt{Contrast}} E_3 \xrightarrow{\texttt{Conjunction}} E_4$,'' an instantiated example, ``I have you number$\rightarrow$I call you$\rightarrow$I have a meeting$\rightarrow$I have a presentation,'' shows that a person wants to call his friends with the phone number. However, he has to do a presentation in the coming meeting and decides to call his friend later. As for the hybrid meta-path with extracted and conceptualized eventualities, ``$E_1 \xrightarrow{\texttt{Conjunction}} E_2 \xrightarrow{\texttt{Conceptualization}} C_1 \xrightarrow{\texttt{ConceptInstantiation}} E_3$,'' we find out that someone is sweating because of the hot weather while someone is unfortunately in a coma. Both of them are unified under the concept ``$\textit{PersonX}$ be $\textit{Symptom}$.''

\begin{table}[]
\centering
\resizebox{\textwidth}{!}{
\begin{tabular}{c|l|l}
\toprule  
\#Hop & meta-path & Instances\\
\midrule  
\multirow{15}*{2} & \multirow{3}*{$E_1$ $\xrightarrow{\texttt {Conjunction}}$ $E_2$ $\xrightarrow{\texttt {Contrast}}$ $E_3$} &  I go to bed $\rightarrow$ I go to sleep $\rightarrow$ I wake up\\
~ & ~ & I have breakfast $\rightarrow$ I have milk $\rightarrow$ I feel sick \\
~ & ~ & I take bus $\rightarrow$ I go to work $\rightarrow$ I go home \\
\cline{2-3}
~ & \multirow{3}*{$E_1$ $\xrightarrow{\texttt{Precedence}}$ $E_2$ $\xrightarrow{\texttt{Precedence}}$ $E_3$} & You go to sleep $\rightarrow$ You wake up $\rightarrow$ You hit the ground  \\
~ & ~ & You drink alcohol $\rightarrow$ You go to toilet $\rightarrow$ You have to pee \\
~ & ~ & You go to restaurant $\rightarrow$ You are sick $\rightarrow$ You go to hospital \\
\cline{2-3}
~ & \multirow{3}*{$E_1$ $\xrightarrow{\texttt {Conceptualization}}$ $C_1$ $\xrightarrow{\texttt{ ConceptInstantiation}}$ $E_2$} & He is psychiatrist $\rightarrow$ \textit{PersonX} is \textit{Specialist} $\rightarrow$ I am attorney \\
~ & ~ & I want milk $\rightarrow$ \textit{PersonX} want \textit{Animal-Product} $\rightarrow$ He wants burgers \\
~ & ~ & You make reservation $\rightarrow$ \textit{PersonX} make \textit{Service} $\rightarrow$ He makes statement \\
\cline{2-3}
~ & \multirow{3}*{$E_1$ $\xrightarrow{\texttt{Conjunction}}$ $E_2$ $\xrightarrow{\texttt{Conjunction}}$ $E_3$} &  I go to gym $\rightarrow$ I have to wait $\rightarrow$ I go home\\
~ & ~ & I am vegan $\rightarrow$ My wife is vegan $\rightarrow$ I used to eat meat \\
~ & ~ & It is a cat $\rightarrow$ It is fine $\rightarrow$ It is beautiful \\
\cline{2-3}
~ & \multirow{3}*{$E_1$ $\xrightarrow{\texttt{Reason}}$ $E_2$ $\xrightarrow{\texttt{Result}}$ $E_3$} & I go to bar $\rightarrow$ I have many friends $\rightarrow$ I have parties \\
~ & ~ & I go to school $\rightarrow$ We could afford $\rightarrow$ I get my first job\\
~ & ~ & I am in pain $\rightarrow$ I am alone $\rightarrow$ I sit at bar \\
\hline
\multirow{15}*{3} & \multirow{3}*{$E_1$ $\xrightarrow{\texttt{Precedence}}$ $E_2$ $\xrightarrow{\texttt{Conjunction}}$ $E_3$ $\xrightarrow{\texttt{Precedence}}$ $E_4$} & The rain comes down $\rightarrow$ The engine whistles $\rightarrow$ The train starts $\rightarrow$ The train moves on\\
~ & ~ & The moon arises $\rightarrow$ The weather is pleasant $\rightarrow$ The snow ceases $\rightarrow$ The night is still\\
~ & ~ & She sleeps $\rightarrow$ The phone rings $\rightarrow$ We gets home $\rightarrow$ She hangs up the phone\\
\cline{2-3}
~ & \multirow{3}*{$E_1$ $\xrightarrow{\texttt{Conjunction}}$ $E_2$ $\xrightarrow{\texttt{Conceptualization}}$ $C_1$ $\xrightarrow{\texttt{ConceptInstantiation}}$ $E_3$} & I play piano $\rightarrow$ I am musician $\rightarrow$ \textit{PersonX} be \textit{Artist} $\rightarrow$ He is actor\\
~ & ~ & I am chill $\rightarrow$ It is a snake $\rightarrow$ It be \textit{Predator} $\rightarrow$ It is a bear\\
~ & ~ & It is hot $\rightarrow$ I am sweating $\rightarrow$ \textit{PersonX} be \textit{Symptom} $\rightarrow$ She is in a coma\\
\cline{2-3}
~ & \multirow{3}*{$E_1$ $\xrightarrow{\texttt{Condition}}$ $E_2$ $\xrightarrow{\texttt{Reason}}$ $E_3$ $\xrightarrow{\texttt{Conjunction}}$ $E_4$
} & Everyone knows him $\rightarrow$ He comes off the bench $\rightarrow$ He makes his debut for club $\rightarrow$ He scores his first goal\\
~ & ~ & I am healthy $\rightarrow$ I sleep $\rightarrow$ I am exhausted $\rightarrow$ I am cold\\
~ & ~ & We get the check $\rightarrow$ We order dessert $\rightarrow$ I am still hungry $\rightarrow$ We eat everything\\
\cline{2-3}
~ & \multirow{3}*{$E_1$ $\xrightarrow{\texttt{Result}}$ $E_2$ $\xrightarrow{\texttt {Contrast}}$ $E_3$ $\xrightarrow{\texttt{Conjunction}}$ $E_4$
} & I am tired $\rightarrow$ I go to bed $\rightarrow$ The sun is shining $\rightarrow$ The wind blows\\
~ & ~ & There is a storm coming $\rightarrow$ The rain falls $\rightarrow$ The sky is clear $\rightarrow$ The air is warm\\
~ & ~ & I have you number $\rightarrow$ I call you $\rightarrow$ I have a meeting $\rightarrow$ I have a presentation\\
\cline{2-3}
~ & \multirow{3}*{$E_1$ $\xrightarrow{\texttt{Contrast}}$ $E_2$ $\xrightarrow{\texttt{Reason}}$ $E_3$ $\xrightarrow{\texttt{Reason}}$ $E_4$
} & I am a vegan $\rightarrow$ I eat meat $\rightarrow$ I enjoy it $\rightarrow$ It tastes good\\
~ & ~ & The painting is controversial $\rightarrow$ It is a masterpiece $\rightarrow$ It belongs to museum $\rightarrow$ It is valuable\\
~ & ~ & I get over it quickly $\rightarrow$ I go to mall $\rightarrow$ I buy clothes $\rightarrow$ I have a job interview\\

\bottomrule
\end{tabular}}
\caption{Instances of meta paths generated by random walk. $E$ represents eventuality while $C$ represents concept. The concepts in the instances are marked as $\textit{italic}$ texts. For example, ``I go to bed  $\rightarrow$ I go to sleep  $\rightarrow$ I sleep'' is an instance of the meta-path ``$E_1$ $\xrightarrow{\texttt{Conjunction}}$ $E_2$ $\xrightarrow{\texttt{Contrast}}$ $E_3$.''}
\label{tab:metapath}
\end{table}


\begin{figure}[t]
    \centering
    \includegraphics[width=0.75\linewidth]{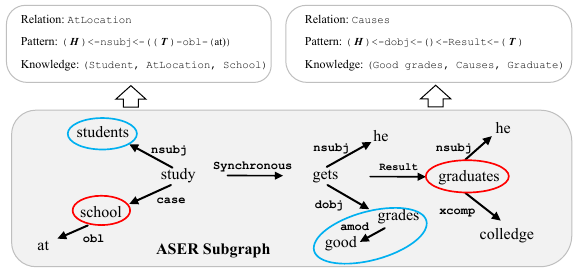}
    \caption{Example of transferring selectional preference knowledge in ASER to commonsense knowledge. By exploring different dependency patterns, we can acquire different forms of candidate knowledge.  Extracted head and tail concepts are indicated with blue and red circles respectively, and are denoted as \textbf{\textit{H}} and \textbf{\textit{T}} placeholders in the patterns. }
    \label{fig:TransOMCS}
\end{figure}

\begin{table}[t]
\renewcommand\arraystretch{1.0}
 \centering
\scriptsize
 \begin{tabular}{l||cc|cc|cc}
  \toprule
    Model & \# Vocab & \# Tuple & Novel$_t$ & Novel$_c$ & ACC$_n$ & ACC$_o$ \\
    \midrule
    COMET$_{Original}$ (Greedy decoding) & 715 & 1,200 & 33.96\% & 5.27\% & 58\% & 90\% \\
    COMET$_{Original}$ (Beam search - 10 beams) & 2,232 & 12,000 & 64.95\% & 27.15 \% & 35 \% & 44\%  \\
    COMET$_{Extended}$ (Greedy decoding) & 3,912 & 24,000 & \textbf{99.98}\% & 55.56\% & 34\% & 47\%  \\
    COMET$_{Extended}$ (Beam search - 10 beams) & 8,108 & 240,000 & \textbf{99.98\%} & 78.59\% & 23\% & 27\%\\
    \midrule
    LAMA$_{Original}$ (Top 1) & 328 & 1,200 & - & - & - & 49\% \\
    LAMA$_{Original}$ (Top 10) &  1,649 & 12,000 & - & - & - & 20\%\\
    LAMA$_{Extended}$ (Top 1) &  1,443 & 24,000 & - & - & - & 29\%\\
    LAMA$_{Extended}$ (Top 10) &  5,465 & 240,000 & - & - & - & 10\%\\
    \midrule
    TransOMCS$_{Original}$ (No Ranking) & 33,238 & 533,449 & 99.53\% & 89.20\% & 72\% & 74\% \\
    TransOMCS (Top 1\%) & 37,517 & 184,816 & 95.71\% & 75.65\% & \textbf{86\%} & 87\%\\
    TransOMCS (Top 10\%) & 56,411 & 1,848,160 & 99.55\% & 92.17\% & 69\% & 74\% \\
    TransOMCS (Top 30\%) & 68,428 & 5,544,482 & 99.83\% & 95.22\% & 67\% & 69\% \\
    TransOMCS (Top 50\%) & 83,823 & 9,240,803 & 99.89\% & 96.32\% & 60\% & 62\% \\
    TransOMCS (No Ranking) & 100,659 & 18,481,607 & 99.94\% & \textbf{98.30\%} & 54\% & 56\% \\
    \midrule
    OMCS in ConceptNet 5.0 & 36,954 & 207,427 & - & - & - & \textbf{92\%}\\
    \bottomrule
  \end{tabular}
 \caption{Main evaluation results of TransOMCS compared with COMET and LAMA. }\label{table:TransOMCS-evaluation}
\end{table}

\begin{figure}[t]
    \centering
    \includegraphics[width=0.7\linewidth]{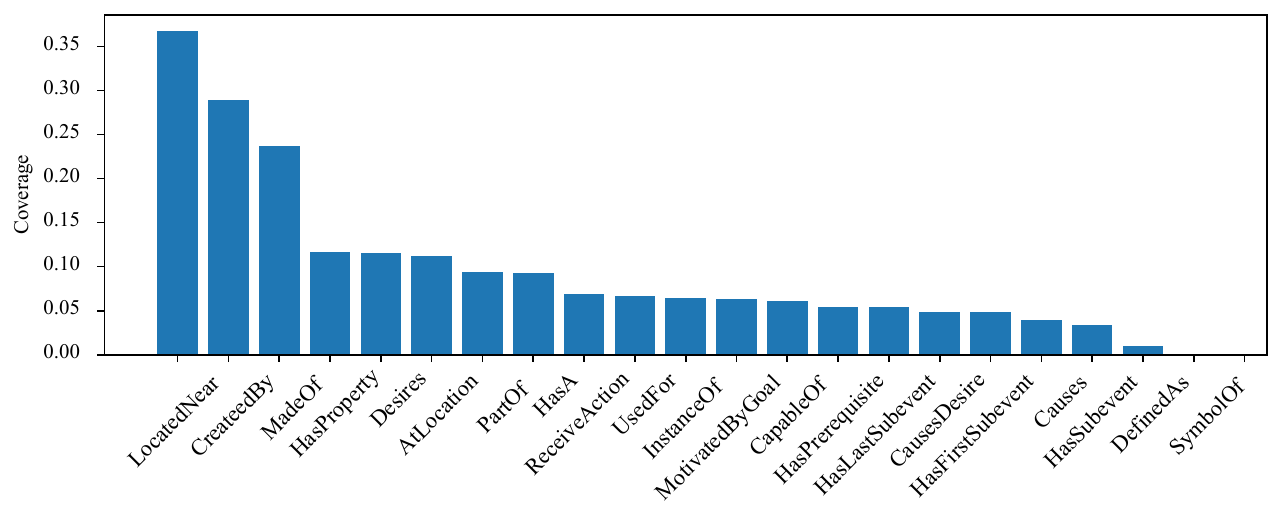}
    \caption{
    The matching statistics of ConceptNet assertions in ASER grouped by each relation. The coverage indicates the proportion of ConceptNet assertions where both heads and tails can be matched to an ASER unit, i.e., a discourse edge or an eventuality.
    }
    \label{fig:TransOMCS-stat}
\end{figure}

\begin{table}[t]
 \begin{minipage}[t]{0.45\textwidth}
  \centering
  \footnotesize
       \begin{tabular}{l|l}
\toprule
Relation & Dependency Pattern \\
\midrule
AtLocation &  \verb|()->compound->()| \\
CapableOf &  \verb|()<-nsubj<-()| \\
Causes &  \verb|(-compound-)<-pobj<-of<-prep<-()| \\
CausesDesire &  \verb|()<-pobj<-to<-prep<-()| \\
CreatedBy &  \verb|()<-dobj<-make->nsubj->()| \\
DefinedAs &  \verb|()<-nsubj<-be->attr->(-amod-)| \\
Desires &  \verb|()<-nsubj<-()| \\
HasA &  \verb|()<-nsubj<-have->dobj->()| \\
HasPrerequisite &  \verb|()->dobj->()| \\
HasProperty &  \verb|()<-nsubj<-be->acomp->()| \\
\bottomrule
\end{tabular}
  \end{minipage}
  \hspace{0.05\textwidth}
  \begin{minipage}[t]{0.45\textwidth}
   \centering
   \footnotesize
\begin{tabular}{l|l}
\toprule
Relation & Dependency Pattern \\
\midrule
HasSubevent &  \verb|(-dobj-)->neg->()| \\
HasFirstSubevent &  \verb|(-prep-)<-Succession<-()| \\
HasLastSubevent &  \verb|()<-acomp<-be<-Reason<-()| \\
InstanceOf &  \verb|()<-acomp<-be->nsubj->()| \\
LocatedNear &  \verb|()<-nsubj<-be->prep->on->pobj->()| \\
MadeOf &  \verb|()<-compound<-()| \\
MotivatedByGoal &  \verb|()<-xcomp<-()| \\
PartOf &  \verb|()->compound->()| \\
ReceivesAction &  \verb|()<-dobj<-()| \\
UsedFor &  \verb|()<-pobj<-(-prep-)| \\
\bottomrule
\end{tabular}
   \end{minipage}
\makeatletter\def\@captype{table}\makeatother\caption{Examples of extracted dependency patterns in TransOMCS. We select the pattern ranked as most plausible for each relation. \texttt{()} are placeholders for words, and attributes like \texttt{nsubj} are names of the dependency edges.} \label{tab:OMCS-pattern}
\end{table}


\begin{table}[t]
 \begin{minipage}[t]{0.45\textwidth}
  \centering
  \footnotesize
       \begin{tabular}{l|l|l}
\toprule
Head& Relation & Tail \\
\midrule
student & \texttt{AtLocation} & school \\
curator & \texttt{AtLocation} & museum \\
leader & \texttt{AtLocation} & group \\
glue & \texttt{CapableOf} & dry \\
anyone & \texttt{CapableOf} & think \\
door & \texttt{CapableOf} & open \\
love & \texttt{Causes} & be friendly \\
attract & \texttt{Causes} & be vulgar \\
want & \texttt{Causes} & be closer \\
music & \texttt{CausesDesire} & listen \\
friend & \texttt{CausesDesire} & talk \\
choice & \texttt{CausesDesire} & entitle \\
art & \texttt{CreatedBy} & artist \\
playoff & \texttt{CreatedBy} & team \\
money & \texttt{CreatedBy} & bank \\
earth & \texttt{DefinedAs} & world \\
god & \texttt{DefinedAs} & truth \\
door & \texttt{DefinedAs} & entrance \\
idea & \texttt{Desires} & come \\
word & \texttt{HasA} & meaning \\
house & \texttt{HasA} & wall \\
bathroom & \texttt{HasA} & sink \\
save & \texttt{HasPrerequisite} & do part \\
enter & \texttt{HasPrerequisite} & ask i \\
\bottomrule
\end{tabular}
  \end{minipage}
  \hspace{0.05\textwidth}
  \begin{minipage}[t]{0.45\textwidth}
   \centering
   \footnotesize
\begin{tabular}{l|l|l}
\toprule
Head& Relation & Tail \\
\midrule
talk & \texttt{HasProperty} & cheap  \\
future & \texttt{HasProperty} & uncertain  \\
be sure & \texttt{HasSubevent} & ask  \\
be hungry & \texttt{HasSubevent} & eat  \\
intrude into & \texttt{HasFirstSubevent} & shoot  \\
go at & \texttt{HasFirstSubevent} & work  \\
closer & \texttt{HasLastSubevent} & go  \\
world & \texttt{MadeOf} & country  \\
whole & \texttt{MadeOf} & part  \\
run & \texttt{MotivatedByGoal} & afraid  \\
eat & \texttt{MotivatedByGoal} & hungry  \\
sleep & \texttt{MotivatedByGoal} & tired  \\
wall & \texttt{PartOf} & house  \\
child & \texttt{PartOf} & family  \\
bone & \texttt{PartOf} & fish  \\
crime & \texttt{ReceivesAction} & commit  \\
game & \texttt{ReceivesAction} & play  \\
video & \texttt{ReceivesAction} & watch  \\
table & \texttt{UsedFor} & sit at  \\
radio & \texttt{UsedFor} & listen to  \\
pool & \texttt{UsedFor} & swim in  \\
nose & \texttt{LocatedNear} & eye  \\
heat & \texttt{LocatedNear} & fire  \\
beaver & \texttt{LocatedNear} & dam  \\
\bottomrule
\end{tabular}
   \end{minipage}
\makeatletter\def\@captype{table}\makeatother\caption{TransOMCS generated ConceptNet-like commonsense knowledge tuples. } \label{tab:OMCS-case}
\end{table}

\begin{figure}[t]
    \centering
    \includegraphics[width=0.8\linewidth]{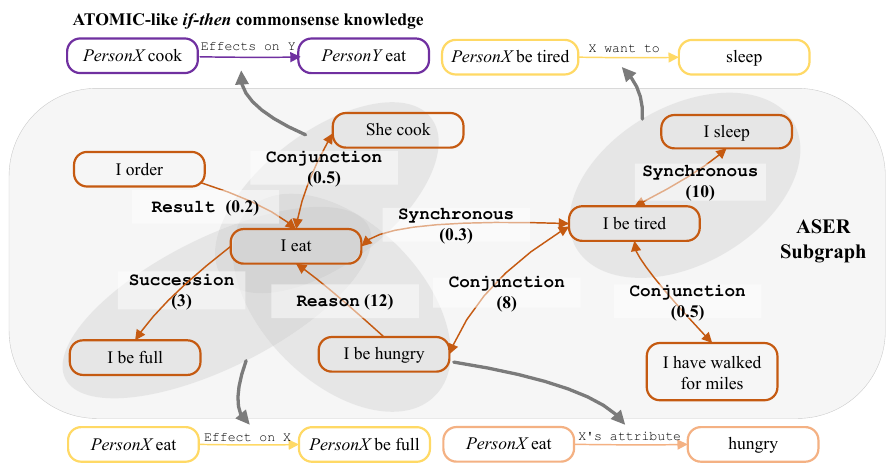}
    \caption{An illustration about exploring novel inferential commonsense knowledge about events. The center of the figure is a real subgraph of ASER. The grey ovals across ASER nodes are the relations that can be transferred to plausible \textit{if-then} commonsense relations. For example, the $\langle$``I eat,'' \texttt{Succession}, ``I eat''$\rangle$ tuple in ASER can be intuitively written as the ATOMIC format, $\langle$``\textit{PersonX} eats,'' \texttt{Effects on X}, ``be full''$\rangle$.}
    \label{fig:aser-atomic-sketch}
\end{figure}

\begin{table}[t]
\centering
\small
\begin{tabular}{c|c|p{10cm}}
  \toprule
  \multicolumn{2}{c|}{} & Mapping rules \\
  \midrule
  \multicolumn{2}{c|}{Head} & Replace \textit{PersonX} and \textit{\textit{PersonY}} with concrete singular personal pronouns, i.e., I/he/she/man/women/person \\
  \midrule
  \multirow{4}{0.7cm}{Tail} &\tabincell{c}{\texttt{xWant}/\texttt{oWant}/\\\texttt{xIntent}/\texttt{xNeed}} & Add a personal pronoun in front of the tail and remove the initial ``to''  \\
  &\texttt{xEffect}/\texttt{oEffect}&Add a personal pronoun in front of the tail \\
  &\texttt{xReact}/\texttt{oReact}&Add a personal pronoun and ``be'' in front of the tail \\
  &\texttt{xAttr}& Add a personal pronoun and ``be'' in front of the tail \\
  \bottomrule
 \end{tabular}
 \caption{Mapping rules from ATOMIC to ASER.}
 \label{table:atomic-preprocess}
\end{table}

\begin{table}[t]
\small
\renewcommand\arraystretch{1.0}
 \centering
 \begin{tabular}{l|c|c|c}
  \toprule
  Relation & Nodes & Edges & Avg. Shortest Path Length \\
  \midrule
oEffect&31.1\% &  25.36\% &  2.41  \\
oReact&87.3\% &  51.53\% &  2.22  \\
oWant&61.6\% &  36.95\% &  2.47  \\
xAttr&95.8\% &  53.67\% &  2.38  \\
xEffect&33.1\% &  21.81\% &  2.51  \\
xIntent&33.8\% & 21.06\% &  2.56  \\
xNeed&52.9\% &  24.91\% &  2.67  \\
xReact&88.7\% &  52.66\% &  2.25  \\
xWant&58.8\% &  30.60\% &  2.59  \\

    \midrule
    Average & 62.9\% & 35.91\% & 2.44 \\
    \bottomrule
  \end{tabular}
 \caption{Mapping statistics of ATOMIC nodes and edges in ASER. The \textit{Nodes} and \textit{Edges} columns denote the percentage of ATOMIC nodes or edges that can be found in ASER. The \textit{Avg. Shortest Path Length} column presents the average shortest path length of the matched ATOMIC edges in ASER.}\label{table:mapping_stat}
\end{table}

\section{From ASER to Commonsense Knowledge}\label{sec:transferability}

In this section, we investigate the connection between ASER and existing commonsense knowledge bases. Specifically, we check the coverage and similarities between the selectional preference knowledge in ASER and the human-defined commonsense knowledge in ConceptNet~\cite{liu2004conceptnet} and ATOMIC~\cite{Maarten2019Atomic}.


\begin{table}[t]
\renewcommand\arraystretch{1.0}
 \centering
\small
 \begin{tabular}{l|l|c|c}
  \toprule
Head& Tail& ATOMIC-Rel& ASER-Rel\\
\midrule
\rowcolor{Gray}
\textit{PersonX} bites \textit{PersonX}'s tongue&   \textit{PersonX} cries& \texttt{xWant}& \texttt{Precedence}\\
\rowcolor{Gray}
\textit{PersonX} feels hungry& \textit{PersonX} eats& \texttt{xWant}& \texttt{Conjunction}\\
\rowcolor{Gray}
\textit{PersonX} opens the envelope & \textit{PersonX} read the letter& \texttt{xWant}& \texttt{Co\_Occurance}\\
\rowcolor{Gray}
\textit{PersonX} pays \textit{PersonX}'s bill& \textit{PersonX} leaves the restaurant& \texttt{xWant}& \texttt{Co\_Occurance}\\
\textit{PersonX} bleeds profusely& \textit{PersonX} passes out& \texttt{xEffect}& \texttt{Co\_Occurance}\\
\textit{PersonX} goes to party& \textit{PersonX} gets drunk& \texttt{xEffect}& \texttt{Conjunction}\\
\textit{PersonX} plays well& \textit{PersonX} wins& \texttt{xEffect}& \texttt{Co\_Occurance}\\
\textit{PersonX} wins the lottery& \textit{PersonX} becomes rich& \texttt{xEffect}& \texttt{Co\_Occurance}\\
\rowcolor{Gray}
\textit{PersonX} would better go& \textit{PersonX} is busy& \texttt{xAttr}& \texttt{Condition}\\
\rowcolor{Gray}
\textit{PersonX} bites \textit{PersonX}'s nail& \textit{PersonX} is nervous& \texttt{xAttr}& \texttt{Synchronous}\\
\rowcolor{Gray}
\textit{PersonX} eats \textit{PersonX}'s breakfast& \textit{PersonX} is hungry& \texttt{xAttr}& \texttt{Condition}\\
\rowcolor{Gray}
\textit{PersonX} holds \textit{PersonX}'s tongue& \textit{PersonX} is quiet& \texttt{xAttr}& \texttt{Co\_Occurance}\\
\textit{PersonX} can not sleep& \textit{PersonX} is stressed & \texttt{xReact}& \texttt{Reason}, \texttt{Condition}\\
\textit{PersonX} is away from home & \textit{PersonX} is lonely& \texttt{xReact}& \texttt{Conjunction}\\
\textit{PersonX} is looking forward to it& \textit{PersonX} is excited& \texttt{xReact}& \texttt{Conjunction}\\
\textit{PersonX} tells \textit{PersonY} everything& \textit{PersonX} is trusted& \texttt{xReact}& \texttt{Conjunction}\\
\rowcolor{Gray}
\textit{PersonX} accepts the challenge& \textit{PersonX} wins & \texttt{xIntent}& \texttt{Co\_Occurance}\\
\rowcolor{Gray}
\textit{PersonX} bows \textit{PersonX}'s head& \textit{PersonX} prays & \texttt{xIntent}& \texttt{Co\_Occurance}\\
\rowcolor{Gray}
\textit{PersonX} removes \textit{PersonX}'s hat& \textit{PersonX} shows respect& \texttt{xIntent}& \texttt{Co\_Occurance}\\
\rowcolor{Gray}
\textit{PersonX} sits in car& \textit{PersonX} waits for \textit{PersonY} & \texttt{xIntent}& \texttt{Co\_Occurance}\\
\textit{PersonX} begins \textit{PersonX}'s work& \textit{PersonX} gets up& \texttt{xNeed}& \texttt{Contrast}, \texttt{Conjunction}\\
\textit{PersonX} closes the door& \textit{PersonX} has opened it & \texttt{xNeed}& \texttt{Synchronous}\\
\textit{PersonX} gets a divorce& \textit{PersonX} gets married& \texttt{xNeed}& \texttt{Reason}\\
\textit{PersonX} makes amends& \textit{PersonX} apologizes& \texttt{xNeed}& \texttt{Conjunction}\\
\rowcolor{Gray}
\textit{PersonX} calls \textit{PersonY}'s name& \textit{PersonY} turns around& \texttt{oEffect}& \texttt{Co\_Occurance}\\
\rowcolor{Gray}
\textit{PersonX} receives a text& \textit{PersonY} waits& \texttt{oEffect}& \texttt{Synchronous}\\
\rowcolor{Gray}
\textit{PersonX} takes \textit{PersonY} a picture& \textit{PersonY} smiles& \texttt{oEffect}& \texttt{Co\_Occurance}\\
\rowcolor{Gray}
\textit{PersonX} tries to tell \textit{PersonY}& \textit{PersonY} refuses to listen& \texttt{oEffect}& \texttt{Contrast}\\
\textit{PersonX} gets pregnant& \textit{PersonY} wants to marry \textit{PersonX}& \texttt{oWant}& \texttt{Precedence}\\
\textit{PersonX} has not seen \textit{PersonY} in years& \textit{PersonY} wants to see \textit{PersonX}& \texttt{oWant}& \texttt{Co\_Occurance}\\
\textit{PersonX} puts \textit{PersonX}'s arm around \textit{PersonY}& \textit{PersonY} pushes \textit{PersonX} away& \texttt{oWant}& \texttt{Conjunction}\\
\textit{PersonX} steals \textit{PersonY}'s wallet& \textit{PersonY} calls the police& \texttt{oWant}& \texttt{Conjunction}\\
\rowcolor{Gray}
\textit{PersonX} complains to the manager& \textit{PersonY} is sorry& \texttt{oReact}& \texttt{Co\_Occurance}\\
\rowcolor{Gray}
\textit{PersonX} did an excellent job& \textit{PersonY} is happy& \texttt{oReact}& \texttt{Conjunction}\\
\rowcolor{Gray}
\textit{PersonX} gives \textit{PersonY} money& \textit{PersonY} is grateful& \texttt{oReact}& \texttt{Conjunction}\\
    \bottomrule
  \end{tabular}
 \caption{Overlaps of ASER and ATOMIC. }\label{table:ASER-ATOMIC-cases}
\end{table}

\subsection{Relationship with ConceptNet}\label{sec:relationship_with_conceptnet}

After around 20 years development, ConceptNet 5.0~\cite{speer2013conceptnet} now contains 21 million edges over 8 million nodes, built from the original ConceptNet~\cite{liu2004conceptnet}.
The core of ConceptNet, which is inherited from the Open Mind CommonSense (OMCS) project~\cite{liu2004conceptnet}, only contains 600K pieces of high-quality commonsense knowledge in the format of tuples, e.g., (`song', \textit{UsedFor}, `sing'). However, there is a huge gap between the small scale of existing commonsense knowledge resources and the broad demand of downstream applications, motivating us to acquire more commonsense knowledge cheaply and feasibly.

Based on the observation that selectional preference can naturally reflect commonsense knowledge about word choice in various contexts \cite{resnik1997selectional}, we proposed TransOMCS \cite{DBLP:conf/ijcai/ZhangKSR20} to transfer the selectional preference knowledge in ASER to ConceptNet-like commonsense tuples. 
Specifically, we adopt the English subset of ConceptNet 5 \cite{speer2013conceptnet} as seed commonsense knowledge, and only relations covered by the original OMCS project~\cite{liu2004conceptnet} are selected. 
Different from OpenIE~\cite{angeli-etal-2015-leveraging} and Hearst patterns~\cite{hearst-1992-automatic}, where human-defined patterns are leveraged to extract relations, we develop a pipeline to discover dependency patterns automatically.
As shown in Figure~\ref{fig:TransOMCS}, for each commonsense relation $r$ in ConceptNet, we first try to find patterns over dependency and discourse relations in ASER automatically from the overlap of ConceptNet assertions and ASER sub-graphs. 
The percentage of ConceptNet knowledge that can be matched in ASER is presented in Figure~\ref{fig:TransOMCS-stat}.
After that, a pattern selection scoring function is designed to select highly plausible patterns. We present the most plausible dependency patterns for each relation as an illustration in Table~\ref{tab:OMCS-pattern}.   Based on the patterns, we can traverse the whole ASER to acquire a large-scale commonsense knowledge graph in the format of ConceptNet.

A running example of TransOMCS is shown in Figure~\ref{fig:TransOMCS}.
For the OMCS-like assertion $\langle$``Good grades,'' \texttt{Causes}, ``Graduate''$\rangle$, we can extract the corresponding dependency relation from the ASER edge $\langle$``he gets good grades,'' \texttt{Result}, ``he graduates colledge''$\rangle$. Such dependency pattern is in turn used for other ASER edges to extract novel knowledge.
As a result, we successfully acquire 18 million ConceptNet-like commonsense assertions with high novelty and accuracy. 
From the case study in Table~\ref{tab:OMCS-case} we can see that ConceptNet-like commonsense knowledge is indeed contained in ASER.
For example, with the help of ASER, we can know that students are often at school, artists often create art, and the wall is part of the house.

We also conducted qualitative analysis regarding the accuracy, novelty, and quantity of the acquired commonsense knowledge in TransOMCS. 
For accuracy, human annotators from Amazon Mechanical Turk are invited to evaluate whether the 100 randomly sampled generated commonsense triple are plausible or not. If at least four annotators out of five agree that the triple is plausible, then it is considered plausible. ACC$_n$ is the accuracy of the novel triples and ACC$_o$ is the overall accuracy of all triples.
For novelty, the proportion of all generated tuples that are novel (Novel$_t$) and that have a novel object/tail (Novel$_c$) are used to measure novelty of the generated knowledge.
We also include the quantity of generated triples for reference.

For baseline models, we compare COMET~\cite{DBLP:conf/acl/BosselutRSMCC19} and LAMA~\cite{DBLP:conf/emnlp/PetroniRRLBWM19} with TransOMCS. When decoding, head-relation pairs in OMCS are fed into the language models to acquire tail outputs, where this setting is denoted as COMET$_{Original}$ and LAMA$_{Original}$. Due to the small size of OMCS (around 1.2K), we also include 24K additional head-relation pairs from the concepts extracted by TransOMCS as additional inputs, where the corresponding models are denoted as COMET$_{Extended}$ and LAMA$_{Extended}$.

The evaluation results of TransOMCS are shown in Table~\ref{table:TransOMCS-evaluation} (Results are from Table 2 in the original paper of TransOMCS~\cite{DBLP:conf/ijcai/ZhangKSR20}). 
In the meantime of producing commonsense knowledge with two more orders of magnitude in terms of quantity, TransOMCS can produce commonsense tails with more novelty and accuracy. For COMET, as it's a pure machine learning based approach which can fit the training data too well to generate novel tails. 
For quality, when the test data is similar to the training set, COMET provides the best quality. 
For example, in the COMET$_{Original}$ setting under greedy decoding, it achieves 90\% overall accuracy. 
The quality of LAMA is whereas less satisfying as a matter of the unsupervised setting and over simple prompts. 
Compared with them, TransOMCS (top 1\%) can generate commonsense knowledge with comparable quality as COMET.

\subsection{Relationship with ATOMIC}

\subsubsection{Overlaps}

Besides ConceptNet, another substantial commonsense knowledge base is ATOMIC \cite{Maarten2019Atomic}, a large-scale human-annotated commonsense knowledge graph that provides inferential knowledge about daily events. Like ASER, the ATOMIC nodes are events described in free-form text, while not parsed to be canonical. There are nine \textit{if-then} relationships defined across ATOMIC, measuring the daily causes and effects for certain base events. To tackle the limitations in terms of novelty and coverage of current \textit{if-then} commonsense acquisition methods, we proposed a novel framework DISCOS (from DIScourse to COmmonSense)~\cite{fang2020discos, DBLP:conf/emnlp/FangWCHZSH21}, which transfers selectional preference knowledge in ASER to complex commonsense knowledge in ATOMIC.
As a result, we acquire 3.4 Million \textit{if-then} commonsense knowledge in the format of ATOMIC. An illustration of the process in DISCOS is presented in Figure~\ref{fig:aser-atomic-sketch}.

Specifically, we first conduct an alignment from ATOMIC to ASER. In ATOMIC, the personal pronouns are represented with wildcards like ``\textit{PersonX}'' and ``\textit{PersonY},'' and in ASER, the subjects of events are concrete personal pronouns like ``she'' and ``he.'' Moreover, as all of the tail events in ATOMIC are written by human annotators, the form of ATOMIC tails can be arbitrary and sometimes subjects are omitted. Based on those observations, we develop some string substitution rules to align the nodes in ATOMIC and ASER, as illustrated in Table~\ref{table:atomic-preprocess}. After conducting the string substitution operations, we use the parser in ASER to parse the acquired text into standard ASER format.

Table~\ref{table:mapping_stat} presents the coverage statistics between ATOMIC and ASER.
We first conduct the string match to check the coverage of ATOMIC nodes in ASER, and find that the average percentage of ATOMIC nodes found in ASER is 62.9\%. For edges, we present the percentage of ATOMIC edges whose head and tail are both covered by ASER, which is 35.91\% on average. 
On top of the matched edges, we check the shortest path length between the matched head and tail in ASER and report the average among all edges in the \textit{Avg. Shortest Path Length} column.
The range of shortest path length starts from 1, where the shortest path length between two directly connected nodes is 1.
We can conclude that, within a few hops of reasoning in ASER, a decent percentage of ATOMIC relations can be inferred. 
Some examples are presented in Table~\ref{table:ASER-ATOMIC-cases}. For instance, the knowledge that if \textit{\textit{PersonX} bites \textit{PersonX}'s tongue}  then the person would want to \textit{cry}, can be entailed from the \texttt{Precedence} discourse relation in ASER. 

\subsubsection{Mining ATOMIC-like Knowledge from ASER}

As the heads and tails in ATOMIC are all arbitrary sentences, the aforementioned pattern mining approach used in TransOMCS is no longer suitable.
To effectively convert ASER knowledge into the ATOMIC format, we propose to use a neural network based classifier instead of hard patterns.
After we match ATOMIC and ASER, we will use the matched eventualities and associated sub-graph as the positive training examples.
For each matched eventuality, we consider its one-hop or two-hop neighbors in ASER to be the candidate eventualities for populating commonsense knowledge of the corresponding ATOMIC relation, whose examples are shown in Table~\ref{table:ASER-neighbor}.
With the help of a graph-based knowledge graph population model and the random negative example sampling, we successfully acquire large-scale commonsense knowledge in the format of ATOMIC.
As demonstrated in Figure~\ref{fig:aser-atomic-sketch} and Table~\ref{table:ASER-concept}, both the original extracted eventualities and edges and those after the conceptualization can help us find rich commonsense about daily events.
For example, before the conceptualization, we can find some knowledge like $\langle$``She takes antibiotic,'' \texttt{Result}, ``She gets better''$\rangle$, which is rather specific. After the conceptualization, we can get a more abstract level commonsense that $\langle$``\textit{PersonX} takes medicine,'' \texttt{Result}, ``\textit{PersonX} gets better''$\rangle$.
Further experiments in~\cite{fang2020discos} also show that compared with a pure supervised model, the knowledge populated with our approach is much more novel and diverse with the comparable high quality. 

Similar but a bit different with that in TransOMCS, we evaluate the acquired commonsense knowledge by DISCOS using accuracy, novelty, and diversity. 
For accuracy, we ask annotators from Amazon Mechanical Turk to determine whether the commonsense tails generated by either COMET or DISCOS are plausible or not. We randomly sampled 50 heads for each relations, and acquire the top 10 results provided by the two models. For COMET, the top 10 results are acquired by Top 10 results using beam search with beam size 10. For DISCOS, the results are acquired by selecting the top 10 neighbors from ASER that are ranked the highest by \textsc{BertSAGE}, a graph-aware model for populating commonsense knowledge on ASER. 
For novelty, we report the proportion of generated tails that are novel (Novelty$_{tail}$), and the proportion of novel tails in the set of all the unique generated tails (Novelty$_{unique}$). The second novelty metric is also expected to be high to avoid the situation when Novelty$_{tail}$ is high while the \textit{novel} tails are all the same.
We also check the diversity among the ten generated tails for each head-relation pair. The proportion of distinct unigrams (Dist-1) and bigrams (Dist-2) among the total number of generated unigrams and bigrams are used here.

Table~\ref{table:DISCOS-evaluation} shows the performance of DISCOS compared with COMET. 
DISCOS can achieve comparable or even better human annotated accuracy on \texttt{oEffect}, \texttt{oReact}, \texttt{oWant}, \texttt{xIntent}, and \texttt{xNeed}) among the nine relations. 
These relations are either of a smaller amount of annotations in ATOMIC or require more temporal knowledge (i.e., \texttt{xIntent}, and \texttt{xNeed} are the causes of the head event instead of effects).
In addition, DISCOS can significantly outperform COMET in terms of novelty.
The reason behind this is similar to that in TransOMCS, which is that COMET fits the training data too well and can suffer from selection bias~\cite{DBLP:conf/icml/Zadrozny04}. 
Due to the limitation of using beam search to generate multiple tails given a head-relation pair, COMET also performs poorly on both diversity metrics than COMET.
As DISCOS adopts an information extraction plus classification framework instead of a text generation framework, it does not suffer from that problem.

\begin{table}[t]
\renewcommand\arraystretch{1.0}
 \centering
\scriptsize
 \begin{tabular}{ll||cc|cc|cc}
  \toprule
    Relation & Model & Novelty$_{tail}$ & Novelty$_{unique}$ & Dist-1 & Dist-2 & Accuracy \\
    \midrule
    \multirow{2}{0.7cm}{oEffect} & COMET@10 & 16.8 & 40.4 & 60.3 & 76.3 & 59.8 \\
    & DISCOS@10 & \textbf{62.9} & \textbf{76.2}  & \textbf{66.7} & \textbf{89.3} & \textbf{68.3}  \\
    \hline 
    \multirow{2}{0.7cm}{oReact} & COMET@10 & 0.4 & 4.9 & \textbf{35.5} & 13.5 & \textbf{69.6} \\
    & DISCOS@10 & \textbf{22.5} & \textbf{50.4}  & 33.5 & \textbf{35.9} & 67.1\\
    \hline 
    \multirow{2}{0.7cm}{oWant} & COMET@10 & 9.8 & 32.4& 46.6 & 84.1 & 69.0 \\
    & DISCOS@10 & \textbf{55.8} & \textbf{75.4} & \textbf{69.0} & \textbf{93.8} & \textbf{69.9}\\
    \hline 
    \multirow{2}{0.7cm}{xAttr} & COMET@10 & 0.1 & 0.8& 8.3 & 4.2 & \textbf{77.7} \\
    & DISCOS@10 & \textbf{12.0} & \textbf{30.4} & \textbf{26.0} & \textbf{27.4} & 66.7\\
    \hline 
    \multirow{2}{0.7cm}{xEffect} & COMET@10 & 8.0 & 24.1& 58.4 & 81.8 & \textbf{75.4} \\
    & DISCOS@10 & \textbf{54.5} & \textbf{71.1} & \textbf{67.2} & \textbf{90.4} & 60.9 \\
    \hline 
    \multirow{2}{0.7cm}{xIntent} & COMET@10 & 12.7 & 31.2 & 42.9 & 75.7 & 86.2 \\
    & DISCOS@10 & \textbf{51.7}& \textbf{74.1} & \textbf{61.5} & \textbf{87.3} & \textbf{87.8} \\
    \hline 
    \multirow{2}{0.7cm}{xNeed} & COMET@10 & 18.6 & 41.0 & 41.4 & 75.7 & 80.7 \\
    & DISCOS@10 & \textbf{44.2} & \textbf{66.2} & \textbf{63.6} & \textbf{88.4} & \textbf{84.9} \\
    \hline 
    \multirow{2}{0.7cm}{xReact} & COMET@10 & 0.4 & 4.7 & 27.1 & 12.1 & \textbf{75.6} \\
    & DISCOS@10 & \textbf{9.1} & \textbf{42.8} & \textbf{29.3} & \textbf{32.9} & 68.4\\
    \hline 
    \multirow{2}{0.7cm}{xWant} & COMET@10 & 12.3 & 30.5 & 42.2 & 78.7 & \textbf{78.9} \\
    & DISCOS@10 & \textbf{38.1} & \textbf{62.0} & \textbf{65.3} & \textbf{91.5} & 73.4 \\
    \bottomrule
  \end{tabular}
 \caption{Main evaluation results of DISCOS and COMET. }\label{table:DISCOS-evaluation}
\end{table}

\begin{table}[t]
 \centering
 
 \footnotesize
 \begin{tabular}{l|l|p{0.1\textwidth}<{\centering}|l}
  \toprule
ATOMIC Head& ATOMIC Tail& ATOMIC-Rel& Add. Neigh. by ASER\\
\midrule
\rowcolor{Gray}
\textit{PersonX} bites \textit{PersonX}'s tongue& \textit{PersonX} cries& xWant& \textit{PersonY} strikes \textit{PersonX} carefully with back\\
\rowcolor{Gray}
\textit{PersonX} bows \textit{PersonX}'s head& \textit{PersonX} prays& xWant& \textit{PersonX} cover \textit{PersonX}'s face with hands\\
\rowcolor{Gray}
\textit{PersonX} catch \textit{PersonY}'s eye& \textit{PersonX} makes an impression & xWant& \textit{PersonY} is interested\\
\textit{PersonX} becomes angry& \textit{PersonX} yells& xEffect& \textit{PersonX} asks for explanation\\
\textit{PersonX} goes to the party& \textit{PersonX} gets drunk& xEffect& \textit{PersonX}'s stomach hurts\\
\textit{PersonX} wins the lottery& \textit{PersonX} becomes rich& xEffect& \textit{PersonX} would quit \textit{PersonX}'s job\\
\rowcolor{Gray}
\textit{PersonX} can not sleep& \textit{PersonX} is stressed& xReact& \textit{PersonX} had a bad day at work\\
\rowcolor{Gray}
\textit{PersonX} is away from home& \textit{PersonX} is lonely& xReact& \textit{PersonX} tries to talk to people\\
\rowcolor{Gray}
\textit{PersonX} is looking forward to it& \textit{PersonX} is excite& xReact& \textit{PersonX} is working hard to get there\\
\textit{PersonX} accepts the challenge& \textit{PersonX} win& xIntent & \textit{PersonY} plays\\
\textit{PersonX} bows \textit{PersonX}'s head& \textit{PersonX} prays& xIntent& \textit{PersonX} is silent\\
\textit{PersonX} sits in car& \textit{PersonX} waits for \textit{PersonY}& xIntent & the police gets \textit{PersonX} out\\
\rowcolor{Gray}
\textit{PersonX} gets a divorce& \textit{PersonX} gets married& xNeed& \textit{PersonX}'s spouse cheats on \textit{PersonX}\\
\rowcolor{Gray}
\textit{PersonX} makes amends& \textit{PersonX} apologizes& xNeed& \textit{PersonX} did wrong\\
    \bottomrule
  \end{tabular}
 \caption{Additional commonsense neighbors that ASER can provide, which can be learned by a knowledge graph population model. }\label{table:ASER-neighbor}
\end{table}

\begin{table}[t]
\renewcommand\arraystretch{1.0}
 \centering
 
 \footnotesize
 \begin{tabular}{l|l|l}
  \toprule
&Head& Tail\\
\midrule
Extracted & she take antibiotic& she get better\\
Conceptualized & \textit{PersonX} take \textit{Medicine}& \textit{PersonX} get better\\
\midrule
Extracted & he pay he bill& money be not plentiful with he\\
Conceptualized & \textit{PersonX} pay \textit{Short-Dated-Asset}& money be not plentiful with \textit{PersonX}\\
\midrule
Extracted & i win the lottery& i become rich\\
Conceptualized & \textit{PersonX} win \textit{Form-of-Gambling}& \textit{PersonX} become rich\\
\midrule
Extracted & he spill coffe& i ask for refill\\
Conceptualized & \textit{PersonX} spill \textit{Beverage}& \textit{PersonY} ask for refill\\

\bottomrule
\end{tabular}
\caption{Examples of \textit{if-then} commonsense knowledge in ASER.The knowledge before and after with Conceptualization are indicated with ``Extracted'' and ``Conceptualized.''}\label{table:ASER-concept}
\end{table}

\section{Applications on Downstream Tasks}\label{sec:extrinsic-evaluation}

After the release of the ASER database\footnote{https://github.com/HKUST-KnowComp/ASER}, many efforts have been devoted to applying the ASER knowledge for downstream tasks. In this section, we briefly introduce representative works of applying the ASER knowledge for downstream tasks and their key observations. More technical details can be found in the original papers.

\subsection{Converting ASER into the Format of Human-crafted Commonsense Knowledge Graph}


As discussed in Section~\ref{sec:relationship_with_conceptnet}, we explored how to convert ASER knowledge into the format of ConceptNet~\cite{liu2004conceptnet}.
To test whether the converted ASER knowledge can help downstream tasks, we conduct experiments on two downstream tasks: commonsense reading comprehension~\cite{ostermann2018semeval} and dialogue generation~\cite{DBLP:conf/ijcnlp/LiSSLCN17}.
Besides the original ConceptNet knowledge base, we also compare with other commonsense knowledge retrieval methods (i.e., COMET~\cite{DBLP:conf/acl/BosselutRSMCC19} and LAMA~\cite{DBLP:conf/emnlp/PetroniRRLBWM19}).

\begin{table*}[t]
	\centering
	\small
	\resizebox{0.68\linewidth}{!}{%
	\begin{tabular}{l||c|c||c|c}
		\toprule
		\multirow{2}{*}{Commonsense Knowledge Resource} & \multicolumn{2}{c||}{Reading Comprehension} & \multicolumn{2}{c}{Dialog Generation}\\
		\cline{2-5}
		& Accuracy (\%) & $\Delta$ (\%) & BLEU & $\Delta$\\
		\midrule
		Base model (no external knowledge resource)         & 82.90 & -     & 0.54 & - \\
		\midrule
		\quad +OMCS                                   & 83.11 & +0.21 & 0.72 & +0.18 \\
		\midrule
		\qquad+COMET         & 83.12 & +0.22 & 0.61 & +0.07 \\
		\midrule
		\qquad+LAMA                & 83.13 & +0.23 & 0.56 & +0.02 \\
		\midrule
		\qquad+ASER knowledge                   & \textbf{83.27} & \textbf{+0.37} & \textbf{1.85} & \textbf{+1.31}\\
		\bottomrule
	\end{tabular}
	}
	\caption{Effect of different knowledge resources on commonsense reading comprehension~\cite{ostermann2018semeval} and dialogue generation~\cite{DBLP:conf/ijcnlp/LiSSLCN17} tasks. }
	\label{tab:transomcs_downstream}
\end{table*}

The experimental results are shown in Table~\ref{tab:transomcs_downstream}. 
For the reading comprehension task, 
adding the ASER knowledge contributes 0.37 overall accuracy, compared to 0.21 contribution of OMCS. 
Meanwhile, the contributions of COMET and LAMA are minor for this task.
For the dialogue generation task,
ASER knowledge also shows remarkable improvement in
the quality of generated responses.
At the same time, adding other knowledge resources to OMCS does not provide any meaningful improvements to the performance. 
The reason behind this
could be
that COMET and LAMA 
provide limited 
high quality novel commonsense knowledge.
For example, the original OMCS on average contributes 1.46 supporting tuples\footnote{Here by supporting tuple, we mean that the head and tail concept appear in the post and response respectively.} and ASER knowledge contributes another 3.36 supporting tuples. As a comparison, COMET and LAMA only provide 0.01, 0.49 additional tuples respectively.

\subsection{Combining ASER Knowledge with Language Models}


Besides converting ASER into commonsense triplets, another work~\cite{DBLP:journals/corr/abs-2012-15643} tries to combine the structured knowledge and pre-trained language models. Motivated by the observation that while language models have already captured rich knowledge, they often only perform well when the semantic unit is a single token while poorly when the semantic unit is more complex (e.g., a multi-token named entity or an eventuality~\cite{DBLP:journals/corr/abs-2007-00849}. For example, if we follow LAMA~\cite{DBLP:conf/emnlp/PetroniRRLBWM19} to analyze the knowledge contained in BERT-large~\cite{DBLP:conf/naacl/DevlinCLT19} with a token prediction task, we can find out that BERT can understand that birds can fly, and a car is used for transportation, but it fails to understand the relation between ``Jim yells at Bob'' and relevant eventualities.
An important reason behind this is that current language models heavily rely on token-level masked language models (MLMs) as the loss function, which can effectively represent and memorize token co-occurrence statistics\footnote{Sinha et al.,~\cite{sinha2021masked} also explains the success of LMs due to distributional information. These models pre-trained over sentences with shuffled word order still achieve high accuracy.} but struggle at perceiving multi-token concepts.
To address this issue, \cite{DBLP:journals/corr/abs-2012-15643} proposed to first verbalize the sub-graphs in ASER into sentences and then further fine-tune the pre-trained language models. A specific loss is added during the training phase to help the models to learn the complex eventuality knowledge in ASER.

 


To test whether the knowledge in ASER can help improve language models' commonsense reasoning ability, \cite{DBLP:journals/corr/abs-2012-15643} conducted experiments on three popular commonsense reasoning tasks: (1) ROCStories~\cite{DBLP:conf/naacl/MostafazadehCHP16}, which is widely used for story comprehension tasks such as Story Cloze Test; (2) MATRES~\cite{DBLP:conf/acl/RothWN18}, that focuses on the temporal commonsense between events; (3) COPA~\cite{DBLP:conf/semeval/GordonKR12} that works on the causal commonsense. Experimental results show that the ASER knowledge can significantly improve the performance of pre-trained language models on these downstream tasks. It also supports our assumption that due to the limitation of the training loss, language models still need the support of structured knowledge to understand those complex commonsense knowledge.

\subsection{Leveraging the Knowledge in ASER for Script Learning}

ASER has been found useful for the task of Script Learning~\cite{DBLP:conf/coling/LvZH20,DBLP:conf/acl/ZhouGSPZJ21}.
The task of Script Learning aims to predict plausible subsequent events given an event chain describing previous states~\cite{ChambersJ08}.
For example, a script depicting someone going to the restaurant may contain ``\textit{PersonX} goes to the restaurant,'' ``\textit{PersonX} reads the menu,'' and ``\textit{PersonX} orders food.'' Script learning aims to predict the following events given the known event chain, for example in the previous case the next step can be ``\textit{PersonX} eats food.'' Understanding scripts can be of vital importance on tasks such as storytelling, dialogue generation, and event understanding.

Lv et al.~\cite{DBLP:conf/coling/LvZH20} use Elastic Search to match the events from event chains to ASER nodes, and select relevant supporting knowledge from their neighbors in ASER. The retrieved knowledge from ASER is then encoded with RoBERTa~\cite{DBLP:journals/corr/abs-1907-11692} and aggregated using an attention mechanism. The knowledge representation is then concatenated with the representation of the event chain as the final representation.
Such a knowledge-aware model can boost the performance of  RoBERTa-Large by over 2 points in terms of accuracy on the Multi-Choice Narrative Cloze (MCNC) dataset~\cite{DBLP:conf/ijcai/LiDL18}. 
Furthermore, instead of only focusing on related subgraphs from ASER of a certain event chain, which may not be enough to equip the model with general script reasoning ability, Zhou et al.~\cite{DBLP:conf/acl/ZhouGSPZJ21} proposed to pre-train a discriminative knowledge model on ASER, where the task is to classify the relationship given head and tail in a $(h, r, t)$ triple. The head and tail are encoded separately with pre-trained language models and an interactive concatenation is applied to model their inner relationship. The finetuned encoder is then used as the encoder for events in the event chains. A chain-contextualized Bi-LSTM is then applied to deal with event chains. 
This model can learn rich relational patterns in the ASER graph for a script in a more supportive way than including local sub-structures only.
Experimental results show that it can further boost the performance of Lv et al.~\cite{DBLP:conf/acl/ZhouGSPZJ21} by 5 points.

\section{Related Works}\label{sec:related_work}

In this section, we introduce related works about commonsense knowledge acquisition, linguistic relation based information extraction systems, and conceptualization. 

\subsection{Commonsense Knowledge Acquisition}

The acquisition of commonsense knowledge can be categorized into three main categories, crowdsourcing~\cite{researchCyc, lenat1995cyc, liu2004conceptnet, Maarten2019Atomic, DBLP:conf/aaai/HwangBBDSBC21, DBLP:conf/emnlp/MostafazadehKMB20}, automatic construction from large-scale corpora~\cite{TandonMSW14WebChild, TandonMW17WebChild2, romero2019commonsense, liu2020mining}, and more recently, mining from pre-trained language models~\cite{DBLP:conf/emnlp/PetroniRRLBWM19, davison2019commonsense, DBLP:conf/nips/BrownMRSKDNSSAA20, DBLP:journals/corr/abs-2110-07178}. 
Details are as follows.


\noindent\textbf{Crowdsourcing Commonsense Knowledge Bases}:
Commonsense knowledge, primarily possessed by ordinary people, was first formalized and collected from human beings ourselves~\cite{lenat1995cyc} with specific guidance towards specific domains. The CYC project asked knowledge engineers to write assertions and formalize the text to logical formats to support logical reasoning. ConceptNet~\cite{liu2004conceptnet} is originated from the Open-Mind CommonSense (OMCS)~\cite{DBLP:conf/coopis/SinghLMLPZ02} project, human annotations are applied to acquire over 400K commonsense assertions among world entities. The latest version of ConceptNet 5~\cite{speer2013conceptnet} now involves the English version of previous ConceptNets, as well as millions of facts from other taxonomy like WordNet and DBPedia. For each entity in ConceptNet, it can be linked to WordNet, Wiktionary, OpenCyc, and DBPedia. Moreover, ConceptNet is now a multi-lingual knowledge base that can also build connections between 83 languages. While ConceptNet focuses on commonsense relations among entities or noun phrases, ATOMIC~\cite{Maarten2019Atomic} is proposed to investigate rich \textit{if-then} relationships among daily social events. Nine social interaction related relations are developed and human annotators are asked to write the corresponding causes or effects of a certain base event. ATOMIC$_{20}^{20}$~\cite{DBLP:conf/aaai/HwangBBDSBC21} is further proposed to unify the triples from ConceptNet and ATOMIC, together with some newly developed relations. 
GLUCOSE~\cite{DBLP:conf/emnlp/MostafazadehKMB20} is a commonsense knowledge base constructed based on ROC Story~\cite{DBLP:conf/naacl/MostafazadehCHP16}. The commonsense causal relations in GLUCOSE are based on cognitive psychology theories that humans primarily focus on events, their timeline, locations of entities, causes and motivations of the event, and emotional trajectory of the character, when focusing on a piece of narrative.


\noindent\textbf{Commonsense Knowledge by Information Extraction}:
Though in general, commonsense knowledge is not explicitly expressed, there is still a non-negligible amount of commonsense knowledge of certain types that can be mined using information extraction tools, such as salient properties of objects~\cite{TandonMSW14WebChild, romero2019commonsense}, verb-oriented selectional preference commonsense~\cite{liu2020mining}, and general statements~\cite{DBLP:journals/corr/abs-2005-00660, nguyen2021advanced}. 
WebChild~\cite{TandonMSW14WebChild} uses semi-supervised label propagation over constructed graphs from web contents, where the seed commonsense knowledge is derived from WordNet. 
Quasimodo~\cite{romero2019commonsense} derives commonsense knowledge from search-engine query logs and QA forums. Syntactical patterns are designed to capture salient properties of objects, for example, detecting questions starting with \textit{Why} and some specific auxiliary verbs of a certain object. 
Verb-Oriented Commonsense Knowledge~\cite{liu2020mining} explores plausible subjects and objects of certain verbs. A large-scale probabilistic taxonomy, Probase~\cite{wu2011taxonomy}, is used to conceptualize subject and object in a verb phrase to get a general s-v-o phrase. An entropy-based filter is applied to determine the appropriate level of conceptualization and a language model is used to score the quality of the provided s-v-o triples.
To capture knowledge that goes beyond $(h, r, t)$ triples, some knowledge bases storing general statements are proposed to be more flexible in representing commonsense knowledge.
GenericsKB~\cite{DBLP:journals/corr/abs-2005-00660} is constructed from large corpora using BERT-based scoring as a filter and including contextual metadata as supporting information. Such kind of knowledge is more flexible and can help some downstream tasks such as question-answering.

\noindent\textbf{Commonsense Knowledge in Pre-trained Language Models}:
With the number of parameters in pre-trained language models~\cite{DBLP:conf/naacl/DevlinCLT19, DBLP:journals/corr/abs-1907-11692, radford2018improving, radford2019language, DBLP:conf/acl/LewisLGGMLSZ20, DBLP:journals/jmlr/RaffelSRLNMZLL20, DBLP:conf/nips/BrownMRSKDNSSAA20} increasing exponentially, researchers are exploring ways to mine commonsense knowledge directly from pre-trained language models, in view of their strong representation ability on large-scale corpora and compositional generalization ability. Such exploration includes both supervised approaches~\cite{DBLP:conf/acl/BosselutRSMCC19, DBLP:conf/aaai/HwangBBDSBC21} and unsupervised approaches~\cite{DBLP:conf/emnlp/PetroniRRLBWM19, davison2019commonsense, DBLP:journals/tacl/JiangXAN20, DBLP:conf/nips/BrownMRSKDNSSAA20}. 
For supervised learning based approaches, pre-trained language models such as BART~\cite{DBLP:conf/acl/LewisLGGMLSZ20} and GPT-2~\cite{radford2019language} are finetuned on large-scale commonsense knowledge bases on a conditional generation task, where the head and relation in the commonsense triple are given as input and the tail serves as the expected output. Those models finetuned on ConceptNet, ATOMIC, and ATOMIC$_{20}^{20}$ can generate commonsense tails with high precision, though may not be generalized enough to generate novel knowledge that is required for commonsense knowledge acquisition. 
For unsupervised approaches, prompts are designed to probe commonsense knowledge directly from large pre-trained models. LAMA~\cite{DBLP:conf/emnlp/PetroniRRLBWM19} and Davison et al.~\cite{davison2019commonsense} designed simple hand-written prompts to conduct factual probing in ConceptNet from BERT~\cite{DBLP:conf/naacl/DevlinCLT19}.
Automatically generated prompts such as best paraphrase-based prompts~\cite{DBLP:journals/tacl/JiangXAN20}, the best sequence of tokens maximizing the gold label likelihood~\cite{DBLP:conf/emnlp/ShinRLWS20}, directly optimized embeddings instead of prompts in the form of text~\cite{DBLP:conf/naacl/ZhongFC21} are used to feed into pre-trained language models to generate outputs, whereas the language model remains untrained.
ATOMIC$^{10x}$~\cite{DBLP:journals/corr/abs-2110-07178} leverages GPT-3, with 100x larger the scale than models such as GPT-2-XL, where some seeds from ATOMIC are used as prompts to acquire commonsense knowledge directly from GPT-3. Human evaluations demonstrate that such an automatically constructed commonsense knowledge base can outperform human annotation in terms of correctness and diversity.

\subsection{Conceptualization}

\noindent\textbf{Conceptualization in Cognitive Science}: People posit the importance of a specific element of human commonsense, conceptualization.
As observed by psychologists, ``concepts are the glue that holds our mental world together''~\cite{murphy2004big}.
Human beings are able to make reasonable inferences by utilizing the IsA relationship between real-world concepts and instances.
For example, without knowing what a ``floppy disk'' is, given that it is a ``memory device,'' people may infer that it may store data and be readable by a computer.
In K-lines theory~\cite{DBLP:journals/cogsci/Minsky80}, people conceptualize the world as a pyramid, and map a K-node (a mental state) to this pyramid, which has a lower-band limit and a higher band limit to ensure right common and non-conflicting properties.
When we want to remember something, we create a K-line for it; when later it is activated, the K-line induces a subset of those mental agencies resembling states that created the K-line.
A lower K-line could affect the instantiation of a more abstract higher level K-line so that K-nodes help us to make abstraction, logical, and procedural reasoning.
For example, we could create a K-line for Tesla by mapping and connecting ``company,'' ``big company,'' ``IT company,'' ``AI company,'' ``high-tech company,'' and ``automobile company.''
As properties are usual non-conflicting, combining the concrete accumulation of particular instances with the rejection of strongly dissonant properties automatically leads to a rather abstract unification.

\noindent\textbf{Conceptualization in Computer Science}: In the computer science community, researchers also explored how to leverage the conceptualization to help machines understand the world.
Probase~\cite{wu2011taxonomy} is a large-scale probabilistic taxonomy to store such ``IsA'' relations between instances and concepts, where 2.7 million concepts are automatically harnessed from 1.68 billion documents.
It has been found useful for several natural language understanding tasks~\cite{SongWWLC11,DBLP:conf/ijcai/WangZWMW15}.
Besides, pattern-based word co-occurrence statistics~\cite{DBLP:conf/acl/RollerKN18,DBLP:conf/acl/LeRPKN19} and distributed embedding models~\cite{DBLP:conf/emnlp/NguyenKWV17,DBLP:conf/naacl/ChangWVM18} can help detect the hypernymy relation to enrich the conceptualization knowledge base.
However, conceptualization needs to address the typicality and ambiguity.
Various computational approaches have been analyzed for deriving basic-level categorization as a trade-off~\cite{DBLP:conf/cikm/WangWWX15}.
To address this issue, contextualized conceptualization was proposed.
Previous works have explored how to leverage topic modeling~\cite{DBLP:conf/ijcai/KimWO13} and external knowledge~\cite{DBLP:conf/ijcai/WangZWMW15} to better conceptualize the concepts based on the local context.
A recent work also explored how to capture the connection between nouns and associated verbs for the better conceptualization~\cite{liu2020mining}.
Last but not least, some attempts on pre-trained models for context-dependent conceptualization also indicated the counter intuitiveness and conceptual inconsistency~\cite{DBLP:conf/naacl/PoradaSTC21}.
\section{Conclusions and Future Works}\label{sec:conclusion}

In this paper, we focus on the commonsense knowledge acquisition problem. Throughout the years, the community has devoted enormous efforts to acquiring commonsense knowledge with either human annotation or information extraction techniques. However, these works are either not scalable or can only handle a specific kind of pre-defined commonsense knowledge. 
To explore a more fundamental understanding of the commonsense knowledge about daily events and states, we follow previous research on the lower bound of semantic theory~\cite{katz1963structure}, partial information~\cite{Wilks1975IAU}, and K-lines theory~\cite{DBLP:journals/cogsci/Minsky80}, and propose to represent commonsense knowledge with higher-order selectional preference over eventualities. Specifically, we first leverage the distribution of daily eventualities and their relations in raw corpus to simulate the plausibility of different semantic combinations, and then leverage the conceptualization module to conceptualize the observed knowledge into an abstract level.
Following this methodology, we develop a large-scale eventuality-centric commonsense knowledge graph ASER, which is a large-scale eventuality knowledge graph that contains 438 million eventualities and 648 million edges.
Considering the large scale of commonsense, we propose an unsupervised pipeline to extract rich commonsense knowledge about events from the raw corpus instead of human annotation.
To effectively represent humans' preference about daily events, we design ASER to be weighed, and larger weight indicates that the eventuality or edge is more likely to happen.
We conduct human evaluations, case studies, and extrinsic evaluations to evaluate the quality of ASER.
As one of the main extraction methodologies of our approach is that we prefer accuracy over recall because we can easily scan more data, even though our current extraction pipeline may sacrifice the recall, it guarantees the high quality of the extracted knowledge.
Further experiments also demonstrate that the knowledge in ASER can be effectively converted into human-crafted commonsense knowledge in other commonsense knowledge bases such as ConceptNet~\cite{liu2004conceptnet} and ATOMIC~\cite{Maarten2019Atomic} and then help downstream tasks such as reading comprehension~\cite{ostermann2018semeval} , dialogue generation~\cite{DBLP:conf/ijcnlp/LiSSLCN17}, story completion~\cite{DBLP:conf/naacl/MostafazadehCHP16}, temporal relation prediction~\cite{DBLP:conf/acl/RothWN18}, and causal relation prediction~\cite{DBLP:conf/semeval/GordonKR12}.

As a long-standing artificial intelligence problem, commonsense reasoning is still challenging for current natural language understanding models. 
In this work, even though we shed some light on how to represent the commonsense knowledge from the angle of partial information, there is still a long way to go to fully solve the commonsense reasoning problem. Specifically, our current research has the following limitations that need to be addressed in the future:
\begin{enumerate}
    \item \textbf{Evaluation}: The first challenge we are still facing is the lack of a good evaluation system. Unlike other tasks, most current commonsense reasoning tasks (e.g., Winograd Schema Challenge~\cite{levesque2011winograd}) are not directly evaluating models' commonsense reasoning abilities. Instead, they are a kind of approximation. Take WSC as an example, many research has discovered that current models can bypass the essential commonsense reasoning and solve the questions with other information~\cite{DBLP:conf/emnlp/ElazarZGR21}. 
    It is quite often that we are just solving a ``dataset'' without solving the underlining ``task'' we truly want to solve.
    \item \textbf{Storage and Computation Efficiency}: As aforementioned, ASER has 438 million eventualities and 648 million edges. Such a large scale guarantees the coverage of ASER, but it also brings a huge burden for storage and computation. Our current hardware architecture and inference algorithms still cannot support fast inference and response. We can try to address this issue from two angles: (1) Better hardware architecture; (2) Better knowledge graph organization.
    \item \textbf{Contextualized Conceptualization}: Another critical challenge we are facing is how to correctly contextualize the observed concepts. As discussed by the K-lines theory~\cite{DBLP:journals/cogsci/Minsky80} and recent research on conceptualization~\cite{liu2020mining}, it is important to conceptualize the observed objects into the correct concept level based on the local context. However, to the best of our knowledge, there is still no reliable contextualized conceptualization model that can handle the open-world scenario. In this work, we use the distribution of concepts over a big corpus instead of the local context to partially remedy this issue. For example, after observing ``dogs can bark'' and the probability of an animal being a dog is 0.08, we will conclude that the plausibility of eventuality ``animal bark'' is 0.08, which indicates that an animal may not always be able to bark, but compared with other entities such as ``house,'' an animal is more likely to bark. A potential limitation of this method is the reporting bias issue, as studied in~\cite{zhang2019sp-10k}, the correlation between the natural distribution and human's commonsense knowledge is slightly less than 0.8. How to handle the reporting bias issue and effectively conceptualize the observed entities based on the local context is a problem worth exploring in the future.
\end{enumerate}

All codes, data, and APIs are published at the project page\footnote{https://github.com/HKUST-KnowComp/ASER} to encourage further research on commonsense and event understanding.

\section*{Acknowledgements}

This paper was supported by  the GRF (16211520) and the RIF (R6020-19 and R6021-20) from RGC of Hong Kong, the NSFC Fund (U20B2053) from the NSFC of China, the MHKJFS (MHP/001/19) from ITC of Hong Kong with special thanks to HKMAAC and CUSBLT, and  the Jiangsu Province Science and Technology Collaboration Fund (BZ2021065).
\section*{Contributions}


The contributions of all authors are as follows. 

\begin{itemize}
    \item \textbf{Hongming Zhang}: Proposing the idea of using higher-order selectional preference over eventualities to represent commonsense knowledge, designing the ASER structure, designing the eventuality and edge patterns, designing the eventuality extraction algorithm, selecting data, conducting intrinsic evaluation, conducting extrinsic evaluations (except dialogue system), and writing the paper.
    \item \textbf{Xin Liu}: Designing and implementing data pre-processing, constituency parsing, clause analyzing, relation extracting with discourse parsing systems, the ASER database schema, and the construction pipeline, providing scripts for extraction and conceptualization and APIs for knowledge databases, managing data and code,
    and drafting the major of Section~\ref{sec:aser-construction} and~\ref{sec:statistics}.
    \item \textbf{Haojie Pan}: Exploring and implementing the conceptualization with Probase, conducting the extrinsic evaluation on the dialogue system, designing the client-server model for the distributed ASER system, and preparing the online demo, and drafting sections relevant to conceptualization.
    \item \textbf{Haowen Ke}: Pre-processing raw data with CoreNLP to acquire lemmatized tokens, pos-tags, name entities, dependency tree, and constituency tree, analyzing the inference results in ASER, and drafting Section~\ref{sec:inference}.
    \item \textbf{Jiefu Ou}: Implementing the rule-based inference over ASER with the AMIE+ system. Assisting Xin Liu for discourse relation extraction and assisting Haowen Ke for analyzing the inference results in ASER.
    \item \textbf{Tianqing Fang}: Analyzing the relation between ASER and other commonsense knowledge bases, and drafting Section~\ref{sec:transferability}.
    \item \textbf{Yangqiu Song}: Proposing the ideas of building an eventuality centric knowledge graph, using conceptualization for abstraction and instantiation, managing the ASER project, and revising the paper.
\end{itemize}






\bibliographystyle{elsarticle-num}
\bibliography{ASER}







\end{document}